\pdfoutput=1
\documentclass[11pt]{article}

\usepackage[margin=1in]{geometry}
\usepackage[T1]{fontenc}
\usepackage{natbib}
\usepackage{amsmath,amssymb,amsthm}
\usepackage{graphicx}
\usepackage{booktabs}
\usepackage{longtable}
\usepackage{caption}
\usepackage{enumitem}
\usepackage{placeins}
\usepackage{xcolor}
\usepackage{xspace}
\usepackage{tikz}
\usetikzlibrary{arrows.meta,positioning,calc,fit,backgrounds}
\usepackage{hyperref}   

\definecolor{drsblue}{RGB}{31,90,145}
\definecolor{drsgrey}{RGB}{100,100,100}
\definecolor{drsfill}{RGB}{234,240,246}
\definecolor{drswarn}{RGB}{176,58,46}
\definecolor{drswfill}{RGB}{250,235,233}

\tikzset{
  block/.style     = {draw=drsblue, fill=drsfill, rounded corners=2pt,
                      align=center, inner sep=4pt, font=\small},
  plain/.style     = {draw=drsgrey, fill=white, rounded corners=2pt,
                      align=center, inner sep=3pt, font=\footnotesize},
  warn/.style      = {draw=drswarn, fill=drswfill, rounded corners=2pt,
                      align=center, inner sep=3pt, font=\footnotesize},
  sum/.style       = {draw=drsblue, fill=white, circle, inner sep=1pt,
                      minimum size=5mm, font=\small},
  flow/.style      = {-{Stealth[length=2mm]}, draw=drsgrey, thick},
  flowb/.style     = {-{Stealth[length=2mm]}, draw=drsblue, thick},
  cell/.style      = {draw=drsgrey!50, minimum width=13mm, minimum height=7mm,
                      font=\footnotesize, inner sep=1pt},
  axlabel/.style   = {font=\footnotesize\bfseries, drsblue},
  note/.style      = {font=\scriptsize, drsgrey, align=left}
}

\hypersetup{
  colorlinks = true,
  linkcolor  = blue,
  citecolor  = blue,
  urlcolor   = blue,
  pdftitle   = {A Unified Framework for Dynamic Reward Shaping in Reinforcement Learning},
  pdfsubject = {Reinforcement learning; reward shaping},
  pdfkeywords= {reinforcement learning, reward shaping, potential-based reward
                shaping, dynamic reward shaping, intrinsic motivation,
                reward design, sample efficiency}
}

\newtheorem{theorem}{Theorem}

\newtheorem{definition}{Definition}
\newtheorem{remark}{Remark}

\title{A Unified Framework for Dynamic Reward Shaping in Reinforcement Learning}

\author{
  \textit{Fouad Bahrpeyma} \\
  \textit{Industrial AI Group} \\
  \textit{Professorship for Information Management} \\
  \textit{Faculty of Informatics / Mathematics} \\
  \textit{Dresden University of Applied Sciences} \\
  \textit{Dresden, Germany} \\
  \texttt{Bahrpeyma@IEEE.org}
}
\date{}

\begin{document}
\maketitle

\begin{abstract}
Sparse, delayed, and weakly informative rewards remain central obstacles to efficient reinforcement learning. Reward shaping addresses these limitations by supplementing the task reward with an auxiliary signal that can accelerate learning while, in the classical setting, the original objective remains the evaluation criterion. Established theory guarantees safety for fixed shaping signals: potential-based reward shaping preserves optimal policies when the auxiliary term is the discounted difference of a time-invariant potential. In contemporary reinforcement learning systems, however, both the learner and the information available for guidance evolve during training: value estimates improve, novelty diminishes, feedback shifts, and predictive models are refined. Adaptive reward mechanisms occur across exploration, Bayesian inference, human-in-the-loop learning, automated reward design, and foundation-model-based approaches. This study introduces a unified analytical framework for comparing dynamic reward shaping and neighbouring adaptive reward mechanisms. The proposed framework distinguishes parametric revision from state-dependent variation, separates additive shaping from reward replacement and reward-adjacent guidance, and organises existing methods along temporal, informational, and theoretical dimensions. Using this framework, twelve method families are comparatively analysed. The framework further highlights the conditions under which optimality guarantees survive contemporary deep reinforcement learning pipelines, replay buffers, bootstrapped critics, and reward normalisation, while exposing the unresolved relationship between adaptation rate and learner stability.
\end{abstract}

\tableofcontents
\newpage

\section{Introduction}
\label{sec:intro}
Reinforcement learning (RL) has emerged as an influential paradigm for sequential decision making, enabling autonomous agents to learn complex behaviours through interaction with their environments rather than explicit supervision \citep{sutton2018reinforcement}. Unlike supervised learning, where desired outputs are provided for every training example, reinforcement learning allows agents to discover effective strategies by continually exploring, acting, and adapting according to feedback received from the environment. This capability has driven progress across a broad spectrum of research areas, including strategic game playing \citep{schrittwieser2020mastering}, robotic manipulation \citep{riedmiller2018learning}, autonomous driving, industrial process optimisation, recommendation systems, wireless communications, finance, healthcare, multi-agent coordination \citep{papoudakis2021benchmarking}, and, more recently, the post-training and alignment of large language models \citep{ouyang2022training}. Recent deep-RL advances have enabled successful applications in high-dimensional and complex decision-making settings. Together, these applications illustrate the use of reinforcement learning as a general framework for sequential decision-making under uncertainty.

Despite these achievements, however, the practical success of reinforcement learning remains fundamentally constrained by the quality of the learning signal available during optimisation. In the standard MDP idealisation, a reward function specifies the objective, and an optimal policy may be learned under suitable assumptions about the environment, exploration, and learning procedure. In many practical tasks, the available reward is sparse, delayed, coarse, or only an imperfect proxy for the intended objective. Rewards are frequently sparse, appearing only after long sequences of interaction, as in long-horizon robotic manipulation tasks where success is registered only after a full sequence of reaching, grasping, and placement decisions \citep{riedmiller2018learning}; delayed, making it difficult to associate successful outcomes with the actions responsible for them; coarse, providing only trajectory-level evaluations without identifying which intermediate decisions were beneficial, as reflected in the large differences in sample efficiency and outright learning failures reported across cooperative multi-agent benchmarks \citep{papoudakis2021benchmarking}; or even imperfect proxies of the true objective, as when a reward model learned from human preference comparisons provides only a scalar assessment of an entire response \citep{christiano2017deep,ouyang2022training}, introducing unintended incentives that may be exploited by the learner. These reward properties can complicate exploration, temporal credit assignment, and policy optimisation, contributing to slow learning, poor sample efficiency, or undesirable behaviour. Consequently, in many practical applications, reward design remains an important challenge alongside the capability of the learning algorithm itself.

To overcome these challenges, the reinforcement learning community has investigated numerous complementary research directions aimed at improving learning efficiency, robustness, and generalisation. Advances in exploration strategies have sought to improve the discovery of informative experiences \citep{bellemare2016unifying,pathak2017curiosity,burda2019exploration}, while representation learning, hierarchical reinforcement learning, model-based reinforcement learning, imitation learning, curriculum learning \citep{narvekar2020curriculum}, and offline reinforcement learning have each addressed different aspects of sample efficiency and policy optimisation. Collectively, these developments have expanded the range of problems addressed by contemporary RL methods. Nevertheless, many of these approaches focus on improving exploration, representation, planning, or optimisation while leaving the task reward or reference objective fixed. In other words, considerable effort has been devoted to improving \emph{how} an agent learns, whereas comparatively less attention has been directed towards improving \emph{what} information is presented to the learner during the learning process.

Among the various directions proposed to improve reinforcement learning, reward design occupies a distinct position because it directly influences the optimisation objective from which the agent learns \citep{ng1999policy}. Rather than modifying the learning algorithm itself, reward design attempts to improve the information content of the feedback received by the learner. The most established framework in this direction is reward shaping, where an auxiliary reward signal is introduced to supplement the original task reward with additional guidance during training. When an auxiliary signal is appropriately designed, reward shaping can provide intermediate feedback that may improve exploration or learning efficiency in long-horizon tasks \citep{ng1999policy, laud2003influence, gupta2022unpacking}. At the same time, however, modifying the reward signal also introduces a fundamental challenge: while the auxiliary reward should facilitate learning, it should not alter the original objective that the agent is ultimately expected to optimise. Balancing these two often competing requirements has consequently become one of the central themes of reward shaping research.


In this study, contemporary reinforcement-learning pipelines refer to systems in which function approximation is combined with training-time mechanisms such as online value estimation, exploration bonuses, learned predictive or reward models, replay, and iterative feedback. Despite the success of classical potential-based reward shaping, its standard guarantees are derived for a shaping function that remains fixed during learning. Many methods considered in this study instead update an auxiliary reward signal, or the information from which it is derived, as experience accumulates. This computational formulation traces a lineage from operant conditioning \citep{skinner1938behavior} and behaviour engineering for physical robots \citep{dorigo1998robot} to a setting in which an auxiliary term $F$ is added to the environment reward $R$, so that the agent is trained using $R+F$ while performance is still evaluated with respect to $R$ alone. When the auxiliary signal is appropriately designed, additive shaping can provide intermediate feedback while the original task reward remains the evaluation criterion. The significance of designing $F$ correctly is illustrated by the canonical bicycle example of \citet{randlov1998learning}, in which progress-based shaping feedback induced the agent to circle repeatedly and collect shaping reward rather than reach the goal. This failure mode motivated the policy-invariance result of \citet{ng1999policy}. Adding $F$ to $R$ is only one way to change the signal a learner sees. A mechanism may instead \emph{replace} $R$ with a learned or generated substitute, \emph{redistribute} or \emph{relabel} the task signal it already has, or leave $R$ untouched and act on the policy, the gradient, or the task distribution. Only the additive case is reward shaping in the sense used here and is the direct application domain of the standard shaping-invariance results; the other three mechanisms are reviewed alongside it, may establish analogous preservation properties only through separate constructions, and are distinguished formally in Section~\ref{sec:unified-framework}. Theoretical developments, most notably potential-based reward shaping (PBRS),
\begin{equation}
F(s,a,s') = \gamma \Phi(s') - \Phi(s),
\label{eq:pbrs}
\end{equation}
further established the conditions under which such auxiliary rewards preserve the optimal policy of the underlying task, namely that the potential $\Phi$ is fixed and the standard assumptions of the result hold \citep{ng1999policy}, providing an important foundation for the practical adoption of reward shaping. Potential-based shaping has become a standard theoretical reference point for reward-shaping research because it provides a policy-invariance result under specified assumptions \citep{ng1999policy}.

Classical potential-based shaping is typically formulated with a shaping function that is fixed throughout learning \citep{ng1999policy}. This assumption need not hold in methods that update value estimates, exploration bonuses, predictive models, or feedback models during training. As learning progresses, the information available to guide the agent may change. Value estimates may be updated as experience accumulates \citep{grzes2010online}, while uncertainty-aware methods update posterior or confidence estimates from observed data \citep{marom2018belief,ma2025highly}; in novelty-based exploration methods, the bonus assigned to familiar states typically decreases as visitation or prediction improves \citep{bellemare2016unifying,burda2019exploration}; and predictive models may also be revised throughout training. In some interactive, preference-based, world-model-based, and foundation-model-based systems, the guidance source is also updated as the learner's behaviour changes \citep{macglashan2017interactive, ouyang2022training,ma2024eureka,xie2024text2reward}. Guidance that is useful early in training may become less informative, or may need to be revised, as the learner's policy and estimates change.

A fixed shaping signal may be restrictive in settings where the auxiliary information available to the learner changes during training. Relaxing it, however, raises questions that classical shaping theory does not fully answer: if the shaping signal itself changes during learning, which guarantees established for fixed shaping survive, particularly the preservation of the optimal policy, and under what conditions? The reviewed theory provides limited direct guidance on how evolving shaping signals interact with function approximation, bootstrapping, and replay, relative to the established fixed-potential results.

Training-time adaptation of reward-related signals and neighbouring guidance mechanisms appears in several RL research areas, including exploration, Bayesian learning, interactive learning, automated reward design, formal task specification, and reward-model-based methods. Exploration-oriented methods gradually reduce novelty bonuses as states become familiar \citep{bellemare2016unifying,pathak2017curiosity,burda2019exploration}; Bayesian approaches modify the influence of prior knowledge as evidence accumulates \citep{marom2018belief,ma2025highly}; learned value functions refine additive shaping terms during optimisation \citep{grzes2010online,adamczyk2025bootstrapped}; interactive learning frameworks update feedback or policy guidance \citep{knox2009interactively,macglashan2017interactive}; automated reward-design methods optimise shaping weights or replacement rewards according to measured task performance \citep{sorg2010reward,zheng2018learning,hu2020learning,gupta2023behavior}; and formal task specifications and curricula change the state, subgoal, or task presented to the learner \citep{camacho2019ltl,icarte2022reward,narvekar2020curriculum}. Foundation-model-based approaches propose or refine reward code from training outcomes \citep{ma2024eureka,xie2024text2reward,bhambri2024extracting,sun2025card}, while vision-language models can be used to update shaping potentials from trajectory preferences \citep{muller2026automating}. These mechanisms share training-time adaptation, but they do not all modify reward and are therefore separated terminologically below.

The representative methods reviewed here originate in several research areas; this study compares them using training-time adaptation as a shared organising principle.


The reviews identified in this study organise reward models, reward engineering, and human advice around different dimensions, rather than using training-time revision of reward-related guidance as their primary organising principle \citep{yu2025reward,ibrahim2024comprehensive,najar2021reinforcement}.
A degree of theoretical unification nonetheless exists: policy invariance has been extended to time-varying potentials \citep{devlin2012dynamic}, broader optimality-preserving classes have been identified \citep{forbes2024pbim,forbes2024generalized,forbes2025action}, and a Bayes-adaptive formulation unifies intrinsic motivation and reward shaping within a common framework \citep{lidayan2025bamdp}. This study relates these results to one another and to the applied mechanisms in the candidate set. These lines of work use different terminology, assumptions, theoretical guarantees, and application settings. The framework proposed here is intended to make their relationships easier to compare. This study adopts training-time revision of reward-related guidance as its organising principle. Within that broad perspective, \emph{dynamic reward shaping} is reserved for additive shaping whose rule is revised after learning has begun; reward replacement, redistribution, and reward-adjacent guidance are treated as neighbouring mechanisms rather than as shaping proper. This distinction also separates parametric revision from a fixed shaping rule evaluated on a changing argument. Potentials indexed by the state of a reward machine, by the current step of a plan, or by an evolving belief may appear dynamic, but are fixed functions over an augmented state representation when their parameters are not revised.

The study develops this idea in three steps. It first formalises dynamic reward shaping as an information-conditioned process in which a shaping rule is updated as evidence accumulates, and consolidates the theory that governs when such revision preserves the optimal policy. It then organises the literature along three dimensions: how the signal changes over training, what information drives the change, and what guarantee the method carries. Finally, it compares the resulting method families in terms of computational cost, tuning burden, and compatibility with contemporary deep RL pipelines, replay buffers, bootstrapped critics, and reward normalisation, where theoretical guarantees and implementation practice frequently diverge.

\subsection{Scope and review approach}
\label{sec:review-approach}
This review centres on methods that revise an additive shaping term during policy optimisation. It also examines three neighbouring mechanism classes: dynamic reward replacement, reward redistribution or relabelling, and reward-adjacent adaptive guidance. These neighbouring classes are included to clarify conceptual boundaries and shared failure modes, but claims about reward-shaping policy invariance are restricted to additive shaping. A method need not expose a separate original task reward in every setting; where replacement or learned-reward methods are considered, the relevant reference is the intended task criterion, held-out evaluator, human judgement, or other external measure against which the adaptive signal is assessed.

The literature was assembled through an iterative, theory-led review rather than a systematic database search. A seed set of foundational results \citep{ng1999policy,wiewiora2003potential,devlin2012dynamic} was expanded by forward and backward citation tracing (publisher pages, Semantic Scholar, and Google Scholar) across the method families connected to training-time reward adaptation, with searches conducted between 2025 and August 2026. Priority was given to works that establish a formal property, introduce a distinct adaptation mechanism, correct an earlier claim, or provide a widely used baseline. Representative neighbouring methods were included when they sharpen the boundary between additive shaping and other forms of adaptive guidance. A work was excluded from Table~\ref{tab:classification} when it adapts a learning signal without touching the reward and does not sharpen a class boundary in Section~\ref{sec:unified-framework}, or when a later, more general result from the same research group superseded it.

This protocol does not satisfy the reporting elements of a systematic review: it does not specify database queries, exact search strings, a full screening log, or reasons for excluding individual papers. Consequently, the entries in Table~\ref{tab:classification} should be read as a representative candidate pool assembled for conceptual and theoretical comparison, not as an exhaustive or statistically representative account of all reward-design research, and counts derived from it (Section~\ref{sec:taxonomy-observations}) describe this reviewed set rather than the field as a whole. Appendix~\ref{app:candidates} lists every C1--C4 entry of Table~\ref{tab:classification} together with its venue and its peer-review status, so that a reader can distinguish claims resting on published, reviewed results from claims resting on preprints.

\paragraph{Organisation and contributions.}
Building upon the unifying perspective introduced above, this review makes four principal contributions. First, it consolidates the theoretical foundations of dynamic reward shaping by bringing together results that have largely developed in isolation, including classical policy invariance, time-indexed potential-based shaping, broader optimality-preserving constructions, Bayesian formulations, and extensions beyond the standard episodic Markov decision process. Second, it introduces a unified taxonomy that organises existing methods according to how an adaptive mechanism changes, the source of information driving that change, and the theoretical guarantees accompanying the resulting learning process. Third, it reviews the major families of adaptive reward and neighbouring guidance mechanisms, analysing their underlying principles, theoretical properties, computational characteristics, and practical trade-offs within a common conceptual framework. Finally, it identifies common implementation challenges, examines current evaluation practice, and outlines open theoretical and practical research directions that may guide the future development of dynamic reward shaping.

The remainder of the paper is organised as follows. Section~\ref{sec:foundations} reviews the theoretical foundations of dynamic reward shaping. Section~\ref{sec:taxonomy} introduces the proposed taxonomy and positions existing methods within it. Section~\ref{sec:families} presents the main literature review of dynamic reward shaping methods, while Section~\ref{sec:analysis} provides a cross-cutting comparison of their theoretical guarantees, computational characteristics, and implementation considerations. Sections~\ref{sec:applications}, \ref{sec:evaluation}, and \ref{sec:agenda} discuss representative application domains, evaluation practices, and open research challenges, respectively.

\section{Formal Foundations}
\label{sec:foundations}
The introduction established two observations that this section formalises. First, the guarantees of classical reward shaping were developed for shaping signals that remain fixed during learning, whereas many methods included in this review adapt rewards or neighbouring guidance during training. Second, these methods share training-time revision but differ in whether the revised object is an additive shaping term, a replacement reward, or a non-reward learning signal. This section defines the setting in which shaping guarantees are stated (Section~\ref{sec:setting}), formalises dynamic reward shaping as an information-conditioned update process and separates genuine revision of the shaping rule from state-dependent variation of a fixed rule (Section~\ref{sec:unified-framework}), and then reviews what the theory establishes: policy invariance for static potentials (Section~\ref{sec:static}), its time-indexed extension (Section~\ref{sec:dynamic-theory}), optimality-preserving constructions beyond the potential-based form (Section~\ref{sec:beyond-pbrs}), and the conditions required outside the standard episodic MDP (Section~\ref{sec:outside}). Section~\ref{sec:theory-practice} closes by identifying what the theory does not deliver, which motivates the empirical and implementation analyses of the later sections.
\subsection{Setting}
\label{sec:setting}

Consider a Markov decision process $M = \langle S,A,T,R,\gamma\rangle$, where $S$ is the state set, $A$ is the action set, $T(s'\mid s,a)$ is the transition kernel, $R:S\times A\times S\to\mathbb{R}$ is a bounded reward function, and $\gamma\in[0,1)$ is the discount factor \citep{sutton2018reinforcement}. Reward shaping replaces $M$ with $M' = \langle S,A,T,R+F,\gamma\rangle$. The agent is trained on $M'$ but evaluated on the original process $M$. The theory is therefore organised around two questions. First, under what conditions does a solution to $M'$ remain a solution to $M$? Second, under what conditions is $M'$ easier to solve than $M$? The first question has established answers under specified constructions, reviewed below. The second has received more limited and construction-specific theoretical treatment, as discussed in Section~\ref{sec:theory-practice}.


\subsection{A unified information-conditioned formulation}
\label{sec:unified-framework}

Dynamic reward shaping is not defined solely by explicit dependence on a time index. Its defining feature is that the shaping rule is revised as information accumulates. Let $\mathcal{I}_k$ denote the information available at the $k$th update. A general dynamic shaping mechanism can then be expressed as
\begin{equation}
R'_k(s,a,s') = R(s,a,s') + F_{\theta_k}(s,a,s'),
\qquad
\theta_{k+1}=U(\theta_k,\mathcal{I}_{k+1}),
\label{eq:general-drs}
\end{equation}
where $U$ denotes an update rule and $\theta_k$ parameterises the shaping function. The information state may include
\begin{equation}
\mathcal{I}_k = \{k,\,\pi_k,\,\mathcal{D}_k,\,\mathcal{M}_k,\,\mathcal{H}_k,\,\mathcal{L}_k,\,\mathcal{A}^{-i}_k\},
\label{eq:information-state}
\end{equation}
where the elements represent, respectively, training progress, the current policy, accumulated experience, learned predictive or reward models, human or AI feedback, foundation-model outputs, and information about other agents. This formulation separates two issues that are often conflated: the information that causes the signal to change and the structural restrictions imposed on the resulting signal.

\begin{definition}[Dynamic reward shaping]
\label{def:drs-general}
A reward-shaping mechanism is \emph{dynamic} if an additive shaping function used during policy optimisation is revised in response to information acquired by the learner, designer, trainer, or an auxiliary model after optimisation has begun. A shaping function that is fixed in advance from prior knowledge alone is static, however much knowledge it encodes. In the classical setting, the task reward remains present and supplies the reference objective. Adaptive reward replacement, redistribution, and reward-adjacent guidance may instead rely on an intended task criterion or an external evaluator rather than a separately available original reward; these mechanisms are neighbouring classes covered by the review, but they are not dynamic reward shaping proper.
\end{definition}

The mechanism classes introduced below are analytical constructs proposed for the purpose of this review. They organise existing methods by their relationship to the task reward and should not be read as terminology originating from the cited literature.

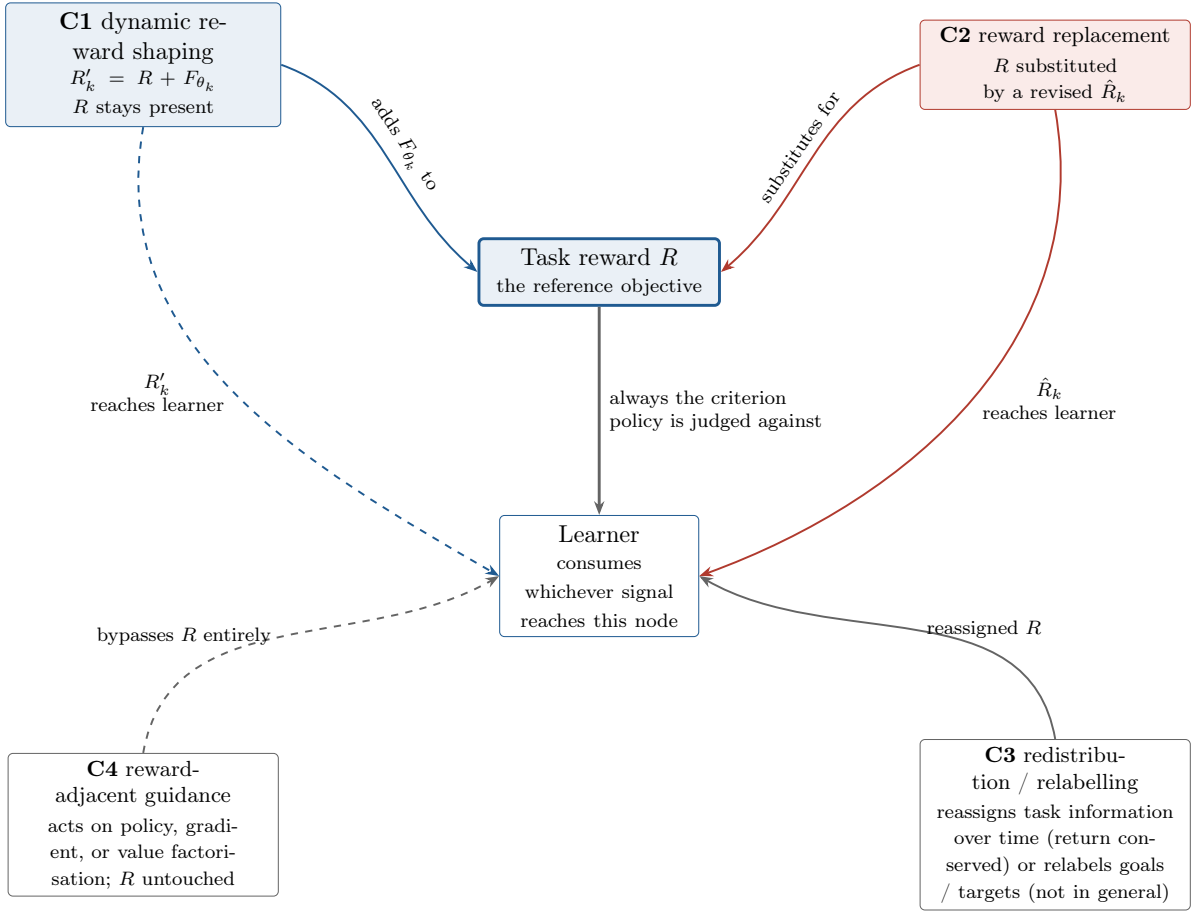
\begin{figure}[!htbp]
\centering
\resizebox{0.95\textwidth}{!}{%
\begin{tikzpicture}[node distance=8mm]
  \node[block, very thick, text width=32mm] (R) at (0,1.6) {Task reward $R$\\[-1pt]\scriptsize the reference objective};
  \node[block, fill=white, text width=26mm] (learner) at (0,-2.8) {Learner\\[-1pt]\scriptsize consumes whichever signal reaches this node};
  \draw[flow, very thick] (R) -- node[right, font=\scriptsize, align=left]{always the criterion\\policy is judged against} (learner);

  \node[block, text width=37mm] (c1) at (-6.6,4.6) {\textbf{C1} dynamic reward shaping\\[1pt]\scriptsize $R'_k = R + F_{\theta_k}$\\$R$ stays present};
  \draw[flowb, thick] (c1.east) to[out=-20,in=140] node[above, sloped, font=\scriptsize]{adds $F_{\theta_k}$ to} (R.west);
  \draw[flowb, thick, dashed] (c1.south) to[out=-100,in=150] node[left, font=\scriptsize, align=center]{$R'_k$\\reaches learner} (learner.west);

  \node[warn, text width=37mm] (c2) at (6.6,4.6) {\textbf{C2} reward replacement\\[1pt]\scriptsize $R$ substituted by a revised $\hat R_k$};
  \draw[-{Stealth[length=2mm]}, draw=drswarn, thick] (c2.west) to[out=-160,in=40] node[above, sloped, font=\scriptsize]{substitutes for} (R.east);
  \draw[-{Stealth[length=2mm]}, draw=drswarn, thick] (c2.south) to[out=-80,in=20] node[right, font=\scriptsize, align=center]{$\hat R_k$\\reaches learner} (learner.east);

  \node[plain, text width=37mm] (c3) at (6.6,-6.4) {\textbf{C3} redistribution / relabelling\\[1pt]\scriptsize reassigns task information over time (return conserved) or relabels goals / targets (not in general)};
  \draw[flow, thick] (c3.north) to[out=100,in=-30] node[right, font=\scriptsize]{reassigned $R$} (learner.east);

  \node[plain, text width=37mm] (c4) at (-6.6,-6.4) {\textbf{C4} reward-adjacent guidance\\[1pt]\scriptsize acts on policy, gradient, or value factorisation; $R$ untouched};
  \draw[draw=drsgrey, dashed, thick, -{Stealth[length=2mm]}] (c4.north) to[out=80,in=-150] node[left, font=\scriptsize]{bypasses $R$ entirely} (learner.west);

  \node[note, align=center, text width=90mm] at (0,-9.0) {Blue (C1): the reward-shaping invariance results of \S\ref{sec:static}--\S\ref{sec:beyond-pbrs} apply directly. Red (C2) and grey (C3, C4): no policy-invariance claim attaches to the mechanism itself; any guarantee reported for these classes concerns its own criterion and is starred ($^{\ast}$) in Table~\ref{tab:classification}. A C2 or C3 construction may still preserve the optimum via a separate equivalence proof of its own.};
\end{tikzpicture}}
\caption{The C1--C4 mechanism classes, organised around a single question: how does the mechanism relate to the task reward $R$? C1 keeps $R$ present and adds to it; C2 substitutes a revised object for $R$; C3 redistributes task information across time, conserving the trajectory return, or relabels the goal or target under which a transition is scored, which does not in general conserve it; C4 leaves $R$ untouched and acts elsewhere in the learning system. Guarantee inheritance follows this structure: C1 is the direct application domain of the potential-based invariance theorems in this taxonomy, while a C2 or C3 construction can establish an analogous preservation result only through a separate explicit equivalence or proof of its own (as for \emph{Dense Reward for Free} in Table~\ref{tab:classification}).}
\label{fig:hierarchy}
\end{figure}

\paragraph{Mechanism classes C1--C4.} To relate additive shaping to nearby approaches without conflating them, this review introduces four mechanism classes and restricts every statement about reward-shaping invariance to the first. These classes are definitions adopted by this review, not standard terminology of the cited literature. They classify a mechanism by its relationship to the task reward $R$ rather than ranking the classes by quality or importance, although the ordering does have consequences for which guarantees can be inherited: the guarantee results of Sections~\ref{sec:static}--\ref{sec:beyond-pbrs} apply directly only to C1.
\begin{itemize}[leftmargin=*]
  \item \textbf{C1, dynamic reward shaping proper.} An auxiliary term is added to the task reward: $R'_k=R+F_{\theta_k}$. The task reward remains present in the signal the learner consumes. Only this class supports the invariance results of Sections~\ref{sec:static}--\ref{sec:beyond-pbrs}.
  \item \textbf{C2, dynamic reward replacement.} The effective reward model substitutes for the task reward during policy optimisation and is itself revised, as in RLHF with a retrained reward model, or generated reward programs. Invariance with respect to the original task is not defined unless the substitute is related to it by an explicit construction.
  \item \textbf{C3, reward redistribution or relabelling.} The task signal is reassigned rather than augmented, and the class covers two subtypes that differ in what is conserved. \emph{C3a, redistribution}, reallocates a fixed trajectory return across time steps, so that the episode return, rather than the per-step reward, is the conserved quantity: token-level redistribution of a terminal sequence score \citep{chan2024dense} is of this kind. \emph{C3b, relabelling}, changes the goal, task interpretation, or target label under which a transition is scored, and recomputes the reward accordingly; hindsight relabelling \citep{andrychowicz2017hindsight} treats an achieved outcome as though it had been the intended one. Return conservation is a property of the redistribution constructions only; relabelling does not in general preserve the original task's episode return. Neither subtype introduces an additive term, so neither is shaping proper, but both densify the learning signal and inherit the corresponding failure mode when the reassignment is a poor proxy for the intended objective.
  \item \textbf{C4, reward-adjacent adaptive guidance.} The reward consumed by the learner is unchanged; adaptation occurs in action selection, in the gradient estimator, in value factorisation, or in the task distribution. These methods are reviewed because they solve the same problem by other means, and because their failure modes recur, but they cannot be assigned reward-shaping guarantee classes.
\end{itemize}
Class membership is recorded for every entry in Table~\ref{tab:classification}. For C2--C4, any reported property concerns that mechanism's own objective or soundness criterion and not reward-shaping policy invariance; the table marks such cells accordingly. Dynamic shaping alone does not imply policy invariance. Invariance that holds for every transition kernel is recovered by restricting Eq.~\ref{eq:general-drs} to the time-consistent potential form
\begin{equation}
F_{\theta_k}(s,a,s')=\gamma\Phi_{\theta_{k+1}}(s')-\Phi_{\theta_k}(s),
\label{eq:information-pbrs}
\end{equation}
whose invariance properties are reviewed in Section~\ref{sec:dynamic-theory}. This distinction underpins the proposed framework. RLHF, learned reward models, and generated reward code are adaptive reward mechanisms, but they are not additive dynamic reward shaping unless they enter as an auxiliary term alongside the task reward. Latent-model bonuses may belong to C1 when they are added to that reward. The classification establishes consistent terminology while maintaining a clear boundary between additive shaping proper and neighbouring mechanisms. Figure~\ref{fig:hierarchy} illustrates the resulting classification.

\paragraph{Temporal vocabulary.} Because the subject of this review is time variation, three temporal notions are kept distinct throughout. A \emph{transition index} $t$ counts environment interaction: the learner departs $s_t$ and arrives at $s_{t+1}$. An \emph{update index} $k$ counts revisions of the shaping parameters $\theta$ in Eq.~\ref{eq:general-drs}, so $\theta_k$ is the $k$th version of the shaping rule. An \emph{epoch} is an interval during which the shaping rule is held fixed while data are collected or the critic is optimised. Writing $k(t)$ for the version active when transition $t$ is generated, the two clocks coincide, $k(t)=t$, when the rule is revised at every environment step, and $k(t)$ is constant within an epoch when it is revised at epoch boundaries only. Section~\ref{sec:dynamic-theory} states the potential-based construction with the two indices identified, which is the convention of the cited theorems; Section~\ref{sec:implementation} is where the distinction becomes operative, since replay, target networks, and epoch-boundary updates decouple the clocks.

Table~\ref{tab:terminology} fixes the vocabulary used to distinguish reward objects from non-reward learning signals throughout the remainder of the review, and records the preliminary C1--C4 placement of each term.

\begin{table}[!htbp]
\centering
\small
\caption{Terminology and C1--C4 mechanism placement used throughout the review. The final column states whether the object is dynamic reward shaping proper or a neighbouring mechanism, using the mechanism classes defined immediately above.}
\label{tab:terminology}
\begin{tabular}{@{}p{0.20\linewidth}p{0.53\linewidth}p{0.20\linewidth}@{}}
\toprule
\textbf{Term} & \textbf{Meaning in this review} & \textbf{Status} \\
\midrule
Task reward $R$ & Environment reward that defines the reference objective in additive shaping. & Reference objective for C1 \\
\addlinespace
Reference criterion & Independently stated measure used to assess an adaptive mechanism. It may be $R$, held-out human judgement, or another external evaluator. & Evaluation concept, not necessarily a reward \\
\addlinespace
Shaping term $F$ & Auxiliary reward added to $R$ during policy optimisation. & Additive shaping, C1 \\
\addlinespace
Dynamic reward shaping & Revision of an additive shaping term $F_{\theta_k}$ after learning begins. & Dynamic shaping proper, C1 \\
\addlinespace
Reward replacement & Learned or generated reward used instead of a separately available task reward. & Neighbouring mechanism, C2 \\
\addlinespace
Reward redistribution or relabelling & Reassignment of task information across time steps (redistribution, which conserves the trajectory return) or across goals and target labels (relabelling, which does not in general conserve it). & Neighbouring mechanism, C3 \\
\addlinespace
Pseudo-reward & Source-dependent literature term for an internally supplied reward-like quantity. It is C1 only when added as $F$; in BAMDP analysis it denotes the broader class defined by that framework. & Context dependent \\
\addlinespace
Advantage or gradient signal & Quantity used in action selection or gradient estimation without changing the reward consumed by the learner. & Not a reward; C4 \\
\addlinespace
Curriculum or structure state & Change in task distribution, subgoal, plan index, belief, or reward-machine state. It is not dynamic shaping unless it revises an additive $F$. & State change or C4 \\
\addlinespace
Adaptive guidance & Umbrella phrase used only when discussing C1--C4 collectively; it does not assert that every included object is a reward. & Cross-class descriptive term \\
\bottomrule
\end{tabular}
\end{table}
\FloatBarrier

\begin{figure}[!htbp]
\centering
\resizebox{\textwidth}{!}{%
\begin{tikzpicture}[node distance=6mm]
  \node[block] (env)  at (0,0)      {Environment $M$\\[-1pt]\scriptsize$\langle S,A,T,R,\gamma\rangle$};
  \node[sum]   (sum)  at (3.6,0)    {$+$};
  \node[block] (agent) at (7.2,0)   {Learner\\[-1pt]\scriptsize policy $\pi_k$};

  \draw[flow] (env) -- node[above,font=\scriptsize] {$R(s,a,s')$} (sum);
  \draw[flow] (sum) -- node[above,font=\scriptsize] {$R'_k$} (agent);
  \draw[flow] (agent) -- ++(0,1.15) -| (env);
  \node[font=\scriptsize, text=drsgrey] at (3.6,1.35) {action $a$};

  \node[block] (shape)  at (0,-2.5)   {Shaping\\[-1pt]$F_{\theta_k}$};
  \node[block] (update) at (3.6,-2.5) {Update rule\\[-1pt]$\theta_{k+1}=U(\theta_k,\mathcal{I}_{k+1})$};
  \node[block] (info)   at (7.2,-2.5) {Information state\\[-1pt]$\mathcal{I}_k$};

  \draw[flowb] (info)   -- (update);
  \draw[flowb] (update) -- node[above,font=\scriptsize] {$\theta_{k+1}$} (shape);
  \draw[flowb] (shape)  -- (sum);
  \draw[flowb] (agent)  -- (info);

  \node[plain, align=left] (src) at (7.2,-4.5)
    {I1 designer \; I4 advice \; I7 foundation models\\
     I2 symbolic \; I5 human/AI feedback \; I8 reward models\\
     I3 agent estimates \; I6 other agents \; I9 world models};
  \draw[flowb] (src) -- (info);

  \node[note, align=center] at (3.6,-6.1)
    {For additive shaping, $R$ remains the reference objective.};
\end{tikzpicture}}
\caption{The information-conditioned view of dynamic reward shaping (Eq.~\ref{eq:general-drs}--\ref{eq:information-state}).
The upper loop is ordinary reinforcement learning on the shaped reward $R'_k=R+F_{\theta_k}$.
The lower loop is what makes the scheme \emph{dynamic}: information accumulated during learning updates
the shaping parameters through $U$. Method families differ in which sources populate $\mathcal{I}_k$
and in what structure is imposed on $F_{\theta_k}$. The potential form of Eq.~\ref{eq:information-pbrs}
is the standard restriction that preserves optimality \emph{without any assumption on the transition
dynamics}; broader optimality-preserving classes exist under additional structural assumptions or over
an enlarged state representation (Section~\ref{sec:beyond-pbrs}).}
\label{fig:framework}
\end{figure}
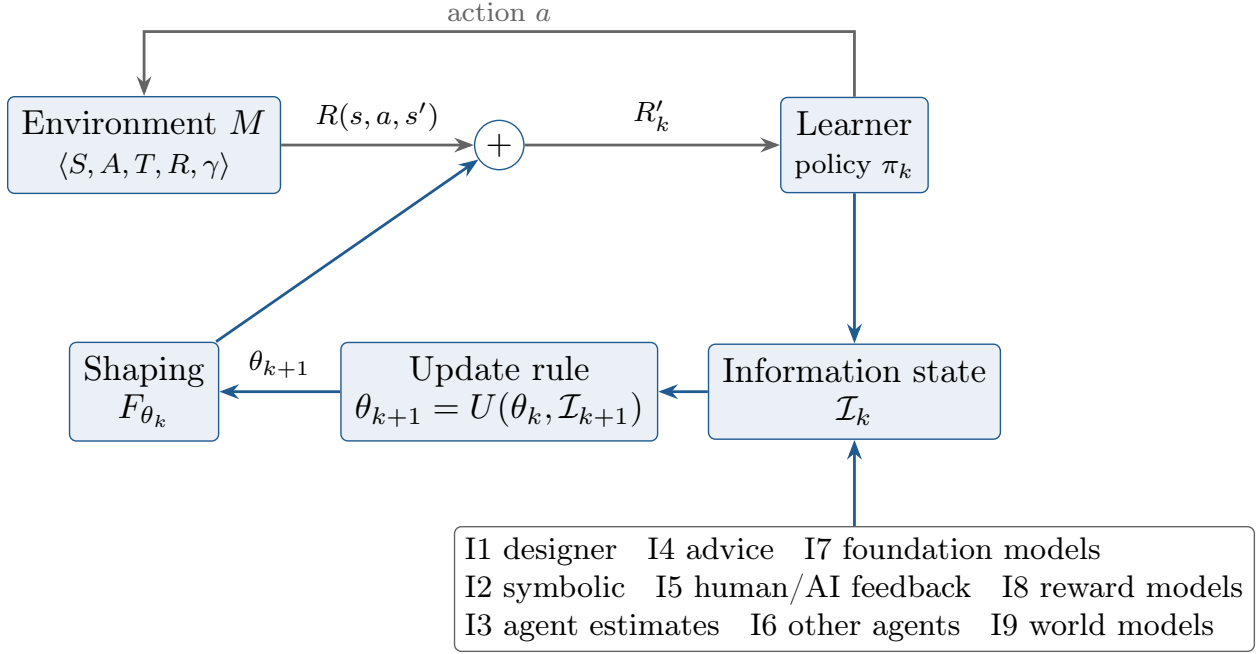

Definition~\ref{def:drs-general} applies directly to scheduled annealing, changing curiosity bonuses, and online potentials when they are added to the task reward. Online reward models, preference-driven refinement, generated reward programs, redistribution, and agent-specific credit signals are retained as neighbouring mechanism classes. Figure~\ref{fig:framework} shows the two loops for additive dynamic shaping: policy optimisation on the shaped reward and revision of the shaping rule itself.

\paragraph{Parametric revision against state dependence.} Definition~\ref{def:drs-general} concerns the parameter trajectory $\theta_0,\theta_1,\dots$, not the values the shaping function happens to take. The distinction between parameter revision and state-dependent evaluation is fundamental to the proposed classification. Two mechanisms produce a shaping term that varies during learning:
\begin{enumerate}[leftmargin=*,label=(\roman*)]
  \item $F_{\theta_k}$ changes because $\theta_k$ is revised by $U$; and
  \item $F_{\theta}(x_k)$ changes because its argument $x_k$ changes, while $\theta$ remains fixed.
\end{enumerate}
Only (i) is dynamic in the sense used here. Mechanism (ii) is a fixed shaping rule evaluated on a varying input, and includes any state-dependent potential, since the state changes at every step by construction. Several constructions that appear dynamic belong to (ii): a fixed potential over belief states, a potential indexed by the current step of a plan, and a potential defined over the states of a reward machine all hold $\theta$ constant while the argument advances. Such methods are more accurately described as \emph{state-dependent static shaping over an augmented representation}. They are recorded separately in Table~\ref{tab:classification} and are excluded from the counts in Figure~\ref{fig:landscape}, because their guarantees follow from the static theory applied to an enlarged state space rather than from any result about revision during learning. Borderline cases exist: an online planner that recomputes a potential from a belief it is simultaneously refining exhibits both mechanisms, and is recorded as such.

\paragraph{A neighbouring paradigm: successor features.} A distinct line of work changes what the reward function \emph{is} across tasks while holding the environment's transition dynamics fixed, rather than revising the reward within a single task as training proceeds. Successor features \citep{barreto2017successor} decompose the action-value function into a dynamics-dependent representation, the successor feature, and a task-dependent weight vector, so that a new task whose reward is linear in the same features admits fast policy evaluation and transfer via generalised policy improvement. This is not dynamic reward shaping under Definition~\ref{def:drs-general}: the reward changes because the \emph{task} changes, not because information accumulated during training revises an additive term within one task, and the framework carries no guarantee about revising a single task's reward while learning proceeds. It is nonetheless close enough in spirit, constraining what is allowed to change so that a computable relationship survives between the old and new solutions, to be a natural point of contact for the ``learn the potential, not the reward'' pattern identified in Section~\ref{sec:taxonomy-observations}. This review does not assign successor-feature methods to C1--C4; they are mentioned here only to mark the boundary of the framework.

\subsection{Static potential-based shaping}
\label{sec:static}

\begin{theorem}[Policy invariance under reward transformations; \citealp{ng1999policy}]
\label{thm:pbrs}
For a discounted MDP, adding a transition-based shaping reward of the potential form of Eq.~\ref{eq:pbrs} preserves the set of optimal policies. Conversely, among transition-based reward transformations required to preserve optimal policies without further assumptions on the transition or reward structure, the potential-based form is necessary, up to the standard positive affine transformation of reward.
\end{theorem}

The theorem above states the classical continuing discounted result and its qualified necessity claim. Additional analytical conditions are needed when the telescoping argument is applied to particular settings. For the infinite-horizon identity below, $\Phi$ is assumed bounded. In episodic tasks, the remaining boundary term must be action-independent; a standard sufficient convention is $\Phi(s)=0$ at every absorbing terminal state, with an analogous condition at a finite horizon \citep{grzes2017reward}. These later qualifications are not part of the theorem attribution to \citet{ng1999policy} and are discussed further in Section~\ref{sec:outside}.

The result follows from a telescoping identity. Along any trajectory,
\begin{equation}
\sum_{k=0}^{T-1}\gamma^{k}\big(\gamma\Phi(s_{k+1}) - \Phi(s_k)\big) \;=\; \gamma^{T}\Phi(s_T) - \Phi(s_0) \;\xrightarrow[T\to\infty]{}\; -\Phi(s_0),
\label{eq:telescope}
\end{equation}
The limiting term $\gamma^{T}\Phi(s_T)$ vanishes because $\gamma<1$ and $\Phi$ is bounded; in the episodic and finite-horizon cases it is instead eliminated by the boundary condition described above. Thus, from a given initial state, shaping contributes the same constant to the return of every policy, and $Q^{*}_{M'}(s,a) = Q^{*}_{M}(s,a) - \Phi(s)$.

In addition, no cycle of states is \emph{profitable relative to the original objective}. This is weaker than the statement, sometimes made, that a cycle generates no shaping reward: individual transitions around a cycle may carry non-zero and individually positive shaping rewards. What the telescoping establishes is that their discounted sum reduces to the same boundary term for every policy, so no policy can raise its shaped return by looping. That property is what prevents the failure observed by \citet{randlov1998learning}.

The precise learning-theoretic interpretation was established by \citet{wiewiora2003potential} for tabular Q-learning. Consider a learner receiving $F(s,a,s')=\gamma\Phi(s')-\Phi(s)$ with initial values $Q_0(s,a)=0$, and an otherwise identical unshaped learner initialised with $Q'_0(s,a)=\Phi(s)$. After corresponding updates, their tables satisfy $Q'_t(s,a)=Q_t(s,a)+\Phi(s)$. Thus, the two tables are not numerically identical; they differ by the state-dependent offset $\Phi(s)$, while their temporal-difference updates are equivalent under this transformation. Their behaviour is identical only for action-selection rules that are invariant to adding the same constant to every action value in a state, including greedy and standard $\epsilon$-greedy selection. \citet{wiewiora2003potential} notes that the update argument extends to Sarsa and related temporal-difference methods, but the equivalence should not be asserted without qualification for arbitrary learning rules, nonlinear function approximation, or policies that depend on absolute value levels. Static PBRS can therefore be interpreted as value initialisation within these conditions. The framework was extended by \citet{wiewiora2003principled} to action-dependent look-ahead and look-back advice. To remain sound, such advice must modify action selection rather than the reward.

\subsection{Time-indexed potentials}
\label{sec:dynamic-theory}

\begin{definition}[Dynamic potential-based shaping]
\label{def:dpbrs}
Let $\Phi_k$ denote the potential used at training step $k$. Here the potential-update index and the transition index are identified, $k(t)=t$ in the notation of Section~\ref{sec:unified-framework}: the construction requires the potential version current at the transition's departure and the version current at its arrival, and the cited theorem is stated for a potential revised once per step. Where the potential is instead revised at epoch boundaries, $\Phi_k$ is to be read as the version active when the transition was generated, which is what makes the replay analysis of Section~\ref{sec:implementation} non-trivial. The dynamic potential-based shaping term applied to a transition between steps $k$ and $k+1$ is
\begin{equation}
F_k(s,a,s') \;=\; \gamma\,\Phi_{k+1}(s') \;-\; \Phi_{k}(s).
\label{eq:dpbrs}
\end{equation}
\end{definition}

\begin{theorem}[Dynamic potential-based shaping; \citealp{devlin2012dynamic}]
\label{thm:dpbrs}
Consider a discounted MDP with $0\leq\gamma<1$ whose transition law and task reward are unchanged by shaping. Let $\{\Phi_k\}_{k\geq 0}$ be a sequence of real-valued potentials known at the corresponding decision times, and add to each transition only the term in Eq.~\ref{eq:dpbrs}, using $\Phi_k$ at departure and $\Phi_{k+1}$ at arrival. Assume either a continuing task with potentials uniformly bounded over states and time, or a finite episodic task whose terminal potential is fixed independently of the preceding action, conventionally at zero. Then the discounted shaping return telescopes to $-\Phi_0(s_0)$ plus a vanishing or action-independent boundary term. Consequently, for a fixed initial state, the shaped and unshaped problems induce the same ordering of policies and the same set of optimal policies.
\end{theorem}

The theorem is a statement about the time-augmented decision process on which the indexed potentials are defined. It does not by itself cover a potential that depends on hidden learner state, retrospectively recomputed replay rewards, or information unavailable when the transition is generated. In the multi-agent extension, the analogous consistently paired potential differences preserve the consistent Nash equilibria of the underlying stochastic game under the construction's stated assumptions \citep{devlin2011theoretical,devlin2012dynamic}.

\paragraph{Extension and interpretation used in this review.} The cited result establishes the dynamic potential construction. The following augmentation analysis is a synthesis used here to delimit where its telescoping argument can be applied; it should not be read as a theorem quoted from \citet{devlin2012dynamic}. If the potential depends only on time, the process can be represented over $(s,k)$. If it depends on additional information, a policy-invariance statement can be made only over a process whose state or history contains enough information to evaluate that dependence. This distinction motivates the following cases.

\begin{remark}[Requirements for augmented state representations]
\label{rem:augmentation}
The augmentation required by the interpretation above depends on what $\Phi_k$ is a function of. Five cases should be kept apart. These cases are stated to delimit where the augmentation argument is even meaningful to invoke, not as a menu of routes for recovering a guarantee; only case (i) and, under an explicit sufficient-statistic assumption, case (ii) typically yield a claim that is checkable at the level of the original environment MDP. Case (iii) is mathematically well defined but is not, by itself, operationally useful: the resulting state space is enormous, evolving, and learner-specific, and establishing that it defines a stationary process in the sense standard RL results require is a further, non-trivial step that this review does not take. A policy-invariance result stated over such an augmented learner-environment process is a different claim from, and does not imply, preservation of optimal policies in the original environment MDP over $S$; the two should not be conflated even when both are informally described as ``invariance.''
\begin{enumerate}[leftmargin=*,label=(\roman*)]
  \item \emph{Exogenous schedule.} $\Phi_k$ depends only on $k$. Augmentation by time suffices and the shaped process is Markov on $(s,k)$.
  \item \emph{Markov information state.} $\Phi_k$ is a function of a statistic $\iota_k$ that evolves as a Markov chain, such as a belief or a summary of visitation counts. The argument goes through over $(s,\iota_k)$.
  \item \emph{Learner-internal quantities.} $\Phi_k$ depends on value-function parameters, replay contents, or the current policy. The shaped process is Markov only over a state space that includes those quantities. This is precisely the enlargement performed by the Bayes-adaptive formulation of Section~\ref{sec:beyond-pbrs}. Invariance in the \emph{original stationary MDP over $S$} does not follow and should not be claimed.
  \item \emph{History dependence.} If $\Phi_k$ depends on the full interaction history with no sufficient statistic, the shaped problem is not an MDP over any finite augmentation, and the result must be stated over histories.
  \item \emph{Anticipative information.} If $\Phi_{k+1}$ is reconstructed using information unavailable at the transition's original arrival, most commonly a potential subsequently updated from data collected in a later training epoch, while the paired $\Phi_k$ at departure is left at the value it had when the transition was generated, the resulting pair no longer matches the chronological construction of Eq.~\ref{eq:dpbrs}, and the argument does not apply to that historical trajectory. This is distinct from consistently relabelling every term of a replayed transition under one current potential, which is a different construction with its own status; both cases are treated in full in Remark~\ref{rem:misreadings} and Section~\ref{sec:implementation}.
\end{enumerate}
Episodic and finite-horizon tasks additionally require an action-independent terminal boundary term, commonly enforced by zero terminal potentials, at each episode boundary. The cited result concerns the shaped decision problem and is silent on whether data are collected on-policy or replayed; that distinction is an optimisation question treated in Section~\ref{sec:implementation}.
\end{remark}

Eq.~\ref{eq:dpbrs} is the information-free statement of Eq.~\ref{eq:information-pbrs}: the index $k$ records \emph{that} the potential has changed, whereas Eq.~\ref{eq:information-pbrs} records \emph{why}. The proof again relies on telescoping. The arrival term at step $k$ cancels the departure term at step $k+1$ when both are evaluated at the time the corresponding state is occupied. The result is consequently permissive about the \emph{source} of the update: the potential may be revised from the agent's value estimates, replay statistics, human feedback, or an outer optimisation loop. It is not permissive about the space over which the resulting guarantee holds, which must be augmented to include whatever $\Phi_k$ depends on, as set out in Remark~\ref{rem:augmentation}. This result provides the theoretical basis for many of the methods reviewed in Section~\ref{sec:families}. A multi-agent counterpart was provided by \citet{devlin2011theoretical}. In that setting, potential-based shaping leaves the Nash equilibria of the underlying stochastic game unchanged, and the dynamic extension preserves this property.

\begin{remark}[Common interpretations and limitations]
\label{rem:misreadings}
(i) Theorem~\ref{thm:dpbrs} constrains the \emph{set of optimal policies}, but it does not characterise the learner's finite-time trajectory. A dynamic potential can preserve optimal-policy invariance while still changing finite-time learning behaviour; the direction and magnitude of that effect are not determined by Theorem~\ref{thm:dpbrs}.  (ii) The theorem requires the index pairing in Eq.~\ref{eq:dpbrs}, and two replay-time patterns that are often run together must be kept apart. \emph{Mismatched-pairing recomputation}, in which a stored transition's departure term $\Phi_k(s)$ is left as originally recorded but its arrival term is replaced by a later $\Phi_{k+1}$ reconstructed from a subsequent training epoch, breaks the telescoping identity for that trajectory and invalidates the result; no version of the dynamic argument applies to it. \emph{Consistent current-potential relabelling}, in which every term of a replayed transition, departure and arrival alike, is recomputed under one shared current potential $\Phi_m$, so that the recomputed reward is $\gamma\Phi_m(s')-\Phi_m(s)$, is not the same failure: this is an instance of the \emph{static} construction of Theorem~\ref{thm:pbrs} evaluated at $m$, and therefore does carry static-PBRS structure at that instant. It does not reconstruct the chronological dynamic process the transition was originally part of, and because $\Phi_m$ is itself revised at the next recomputation, the learner is solving a sequence of distinct static-PBRS problems rather than one stationary objective, which is a different, and separately non-trivial, optimisation question rather than a proof failure. (iii) The theorem concerns the shaped MDP rather than a particular learning algorithm. Under function approximation, bootstrapping, and replay, a time-varying reward creates a moving regression target. The resulting optimisation behaviour is not addressed by the theorem. These three issues are examined in Section~\ref{sec:implementation}.
\end{remark}

\subsection{Optimality-preserving shaping beyond potential-based form}
\label{sec:beyond-pbrs}

The necessity clause of Theorem~\ref{thm:pbrs} is sometimes interpreted as requiring all useful shaping signals to be potential-based. The actual statement is weaker: no \emph{other} form is guaranteed to be safe for \emph{all} transition dynamics. When additional structure is imposed on the shaping signal, broader optimality-preserving classes can be obtained. Their characterisation has become an active area of research, motivated largely by intrinsic-motivation bonuses. Such bonuses are not naturally potential-based and have commonly been used despite their known capacity to change the optimal policy.

Potential-Based Intrinsic Motivation (PBIM) \citep{forbes2024pbim} converts an intrinsic-motivation reward into potential-based form by defining the potential as the expected discounted future intrinsic return under the current policy, $U_t^\pi$:
\begin{equation}
\label{eq:pbim}
\Phi_t =
\begin{cases}
-U_0^\pi/\gamma^N & t = N,\\
-U_t^\pi & \text{otherwise},
\end{cases}
\end{equation}
The construction in Eq.~\ref{eq:pbim} makes the intrinsic reward the difference of successive potential values, exactly as required by Theorem~\ref{thm:pbrs}. The guarantee is conditional: optimality is preserved provided the shaping reward at each step does not depend on actions taken after that step (the \emph{future-agnostic} assumption of \citet{forbes2024pbim}), which in turn requires the cumulative discounted intrinsic return to be action-independent. Under this condition the optimal policy set of the original episodic MDP is unchanged. Both the dependence and the condition are structural: they restrict the \emph{form} of the shaping signal rather than the learner.

Generalized Reward Matching (GRM) \citep{forbes2024generalized} broadens the admissible class from potential differences to history-dependent corrections. The corrected shaping reward is
\begin{equation}
\label{eq:grm}
F_t^{\prime\,\mathrm{GRM}} = F_t - \sum_{i=0}^{t} \gamma^{\,i-t} F_i\, m_{t,i},
\end{equation}
In Eq.~\ref{eq:grm} the matching function $m_{t,t'} \in [0,1]$ specifies which fraction of the intrinsic reward received at step $i$ is subtracted back at step $t$. Optimality is preserved under two conditions on $m$: it must be \emph{fully matching}, so that every unit of intrinsic reward is fully subtracted back by the end of the episode ($\sum_{j=t'}^{N-1} m_{j,t'} = 1$ for all $t'$), and \emph{future-agnostic}, so that $m_{t,t'} = 0$ for $t' > t$. The admissible dependence is thereby extended to the full history of past intrinsic rewards, which is why the class encompasses all optimality-preserving potential-based shaping functions while remaining restricted to episodic tasks. As with PBIM, the guarantee applies to the optimal policy set of the original episodic MDP, and the conditions were validated in sparse gridworld tasks, where uncorrected intrinsic motivation was shown to induce convergence to suboptimal policies \citep{forbes2024pbim,forbes2024generalized}.

Action-Dependent Optimality-Preserving Shaping (ADOPS) \citep{forbes2025action} replaces the structural restriction with an online correction. The shaping reward is $F' = F + F_2$, where the correction
\begin{equation}
\label{eq:adops}
F_2 =
\begin{cases}
\min\!\big(0,\; \hat V_E^\pi - \hat Q_E^\pi + \hat V_I^\pi - \gamma \hat V_I^\pi(s') - F - \epsilon\big) & \text{if } \hat Q_E^\pi < \hat V_E^\pi,\\[2pt]
\max\!\big(0,\; \hat V_E^\pi - \hat Q_E^\pi + \hat V_I^\pi - \gamma \hat V_I^\pi(s') - F\big) & \text{otherwise},
\end{cases}
\end{equation}
in Eq.~\ref{eq:adops} is computed from the learner's own estimates of the extrinsic and intrinsic value functions ($\hat V_E^\pi, \hat Q_E^\pi, \hat V_I^\pi$). The correction enforces two conditions at every step: extrinsically optimal actions remain tied under the augmented value function, and extrinsically suboptimal actions remain strictly dominated. Because the condition is enforced behaviourally through the critic rather than imposed on the form of the potential, the cumulative intrinsic return is permitted to depend on the agent's actions, and neither the episodic assumption nor the future-agnostic assumption is required. The guarantee holds under the assumption that the training algorithm executes only stable policies upon convergence, and it again concerns the optimal policy set of the original MDP. This relaxation is important in long-horizon, exploration-intensive domains: on Montezuma's Revenge, ADOPS retains the benefit of intrinsic motivation where the earlier PBRS-style corrections fail to outperform unshaped exploration \citep{forbes2025action}.

A broader formulation was developed by \citet{lidayan2025bamdp}, who represent all pseudo-rewards, including intrinsic motivation and reward shaping, as shaping within a Bayes-Adaptive MDP (BAMDP) defined over the agent's knowledge state. Under this view, the value of a BAMDP state decomposes into the value of information that can be acquired and the prior value of the physical state, and a pseudo-reward is beneficial when it promotes behaviour that increases these components. The key construction is a potential over the BAMDP state, or equivalently over the interaction history:
\begin{equation}
\label{eq:bampf}
F(h_t) = \gamma\,\phi(h_t) - \phi(h_{t-1}),
\end{equation}
where $h_{t-1}$ is the length-$(t-1)$ prefix of the history $h_t$. Two guarantees are established. In the meta-RL setting, being potential-based on the BAMDP state is \emph{necessary and sufficient} for the optimal algorithm of the shaped BAMDP to remain Bayes-optimal for the underlying RL problem. This is a stronger characterisation than the sufficient conditions of Theorem~\ref{thm:pbrs}. In the ordinary RL setting, pseudo-rewards expressible in the form of Eq.~\ref{eq:bampf} with a potential that is bounded and monotone increasing over training time eventually preserve \emph{approximate} optimality: for every $\epsilon > 0$ there is a training step after which the shaped policy's unshaped return is within $\epsilon$ of optimal. The mechanism is the telescoping argument of Section~\ref{sec:static} applied to the episode-level boundary term: a bounded monotone potential can no longer confer an advantage exceeding $\epsilon$ once its increments have fallen below that threshold. The two settings therefore receive different classifications in Table~\ref{tab:classification}: G1 for exact Bayes-optimal preservation in meta-RL and G3 for a formal eventual-approximation result in ordinary RL. This formulation provides a broad unifying account, and its key step of enlarging the state space to include the agent's epistemic state formalises the information change central to dynamic guidance.

\subsection{Conditions outside the standard setting}
\label{sec:outside}

\paragraph{Episodic tasks and terminal states.} The telescoping identity in Eq.~\ref{eq:telescope} depends on the final term. In episodic problems, invariance requires an appropriate treatment of absorbing states. The standard condition is $\Phi(s_{\mathrm{terminal}}) = 0$. As shown by \citet{grzes2017reward}, violating this condition changes the solution to which the shaped learner converges. When the potential is \emph{learned}, the condition must be imposed explicitly because it is not guaranteed to arise from the learning process.

\paragraph{Partial observability.} Most reward-shaping theory assumes that the potential is a function of the Markov state. Potential-based shaping was extended to finite-horizon online POMDP planning by \citet{eck2016potential}. Belief-state information is incorporated into the potential to provide guidance about rewards beyond the planning horizon. A Bayesian approach to the RL setting was proposed by \citet{marom2018belief}. The reward distribution is augmented with prior beliefs whose influence decreases with experience, and consistency with the optimal policy of the original MDP is established under suitable conditions for Q-learning. Both approaches adapt as beliefs evolve, but they differ under Definition~\ref{def:drs-general}: \citet{eck2016potential} holds the potential fixed over a belief-augmented state and is therefore state-dependent static shaping in the sense of Section~\ref{sec:unified-framework}, whereas \citet{marom2018belief} revises the effective reward model as evidence accumulates and qualifies as dynamic. Their guarantees correspondingly derive from different results.

\paragraph{Multiple agents.} When several agents learn simultaneously, equilibrium consistency replaces single-agent policy invariance as the relevant guarantee \citep{devlin2011theoretical}. The shaping problem also acquires a second dimension: a shared reward must be attributed to individual contributors. This credit-assignment problem is addressed by difference rewards \citep{tumer2007distributed} and by their combination with potential-based shaping \citep{devlin2014potential}.

\subsection{Limitations of existing theoretical guarantees}
\label{sec:theory-practice}

The preceding results concern \emph{safety}: they identify shaping signals that leave the task solution unchanged. They do not establish that shaping is beneficial. Within the works reviewed here, analyses of shaping benefit take at least three partial forms. A conceptual account was given by \citet{laud2003influence}, who interpret shaping as a reduction in the \emph{reward horizon}, defined as the number of decisions between an action and an informative reward. When this horizon is short, the learning time is shown to be polynomial in the size of a critical region rather than in the size of the complete MDP. A sample-complexity analysis was provided by \citet{gupta2022unpacking}. By incorporating shaped rewards into a novelty-based exploration method, they showed that specific shaping choices can provably improve sample efficiency. The reviewed convergence results for individual methods include tabular analyses \citep{adamczyk2025bootstrapped,marom2018belief}.

The distinction between a structural preservation result and observed learning behaviour is also visible in deep-RL experiments. \citet{muller2025improving} observed that an additive shift of the potential function measurably improves the effectiveness of potential-based shaping in deep RL. In the idealised setting, this operation is theoretically irrelevant because it changes the shaped return only by a constant. Limitations of continuous potential functions were also identified that do not arise in the tabular theory. This example illustrates that a structural policy-invariance result does not by itself characterise empirical learning behaviour under deep-RL implementation choices.

\subsection{Theorem-assumption summary}
\label{sec:theorem-summary}

The results reviewed in this section are stated under different combinations of setting, timing, and representation assumptions, and the prose above states each one individually. Table~\ref{tab:theorems} collects them in one place so that the assumptions under which a G1--G3 label in Table~\ref{tab:classification} was earned can be checked directly, without re-deriving them from the surrounding paragraphs. Two columns matter most for the audit this table supports: \emph{information timing} states what must be known and when, which is where the replay-recomputation hazard of Section~\ref{sec:implementation} originates; and \emph{tabular / function approximation} states whether the cited result itself addresses deep-RL implementation, which none of them does beyond the tabular case except where the table says so explicitly.

{\small
\begin{longtable}{@{}p{0.15\linewidth}p{0.17\linewidth}p{0.15\linewidth}p{0.19\linewidth}p{0.13\linewidth}p{0.19\linewidth}@{}}
\caption{Assumptions and exact conclusions of the formal results in Section~\ref{sec:foundations}, stated so that a G1--G3 assignment in Table~\ref{tab:classification} can be checked against its source. ``Information timing'' states the causality condition required by the dynamic construction. Mismatched replay-time recomputation, in which a stored transition's departure term is left stale while its arrival term is reconstructed from a later potential, can violate that condition (Remark~\ref{rem:augmentation}, case v). Consistently recomputing both terms of a transition under a single current potential instead defines a different, static-PBRS construction, whose non-stationary optimisation behaviour remains separately unresolved (Remark~\ref{rem:misreadings}, case ii; Section~\ref{sec:implementation}). ``Tabular / FA'' states whether the cited result itself covers function approximation, bootstrapping, or replay, as distinct from being merely compatible with them in practice.}
\label{tab:theorems}\\
\toprule
\textbf{Result} & \textbf{Decision process} & \textbf{Episodic / continuing; terminal condition} & \textbf{Information timing} & \textbf{Tabular / FA} & \textbf{Exact conclusion} \\
\midrule
\endfirsthead
\multicolumn{6}{c}{\tablename\ \thetable\ (continued)}\\
\toprule
\textbf{Result} & \textbf{Decision process} & \textbf{Episodic / continuing; terminal condition} & \textbf{Information timing} & \textbf{Tabular / FA} & \textbf{Exact conclusion} \\
\midrule
\endhead
\midrule
\multicolumn{6}{r}{Continued on next page}\\
\endfoot
\bottomrule
\endlastfoot
Theorem~\ref{thm:pbrs}, static PBRS \citep{ng1999policy} & Original MDP state $S$, unaugmented & Continuing discounted (primary); episodic requires $\Phi(s_{\mathrm{terminal}})=0$, added by \citet{grzes2017reward}, not part of the cited theorem & $\Phi$ fixed before learning begins; no timing condition to state & Representation-agnostic as a claim about the transformed MDP; the tabular value-initialisation equivalence is \citet{wiewiora2003potential}'s separate, tabular-specific result & Optimal-policy set of $M'$ equals that of $M$; $Q^{*}_{M'}(s,a)=Q^{*}_{M}(s,a)-\Phi(s)$ \\
\addlinespace
Theorem~\ref{thm:dpbrs}, dynamic PBRS \citep{devlin2012dynamic} & Original $S$ only if $\Phi_k$ depends solely on $k$ or a Markov sufficient statistic (Remark~\ref{rem:augmentation}, cases i--ii); otherwise an augmented, learner-specific process not established to be stationary & Both; continuing requires potentials uniformly bounded over states and time, episodic requires an action-independent terminal potential, conventionally zero & $\Phi_k$ known at departure, $\Phi_{k+1}$ at arrival; reconstructing $\Phi_{k+1}$ from a later training epoch while leaving $\Phi_k$ at its original value mismatches the pairing and invalidates the result for that trajectory, whereas consistently recomputing both terms under one current potential is a different, static construction (Remark~\ref{rem:misreadings}, case ii) & Theorem is stated at the level of the decision process and is silent on function approximation, bootstrapping, or replay (Remark~\ref{rem:misreadings}, case iii); these are treated as a separate optimisation question in Section~\ref{sec:implementation}, not covered by the theorem & Same ordering and same optimal-policy set for shaped and unshaped problems, for a fixed initial state, over the (possibly augmented) decision process \\
\addlinespace
PBIM \citep{forbes2024pbim} & Original episodic MDP state & Episodic only, fixed horizon $N$; terminal potential fixed at $-U_0^\pi/\gamma^N$ & Future-agnostic: the shaping reward at each step must not depend on actions taken after that step & Not restricted in principle by the construction; this review does not independently confirm deep-RL robustness beyond what the cited paper reports & Optimal-policy set of the original episodic MDP unchanged, conditional on the future-agnostic and action-independent cumulative-intrinsic-return conditions \\
\addlinespace
GRM \citep{forbes2024generalized}, preprint & Original episodic MDP state & Episodic only & Future-agnostic matching function ($m_{t,t'}=0$ for $t'>t$) and fully matching (all intrinsic reward subtracted back by episode end) & Not restricted in principle; validated in sparse gridworld tasks in the cited preprint, which this review has not independently reproduced & Optimal-policy set of the original episodic MDP unchanged under the matching conditions; the admissible class subsumes all optimality-preserving potential-based functions \\
\addlinespace
ADOPS \citep{forbes2025action} & Original MDP state; no episodic restriction & Not restricted to episodic tasks, unlike PBIM and GRM & Correction computed online from the learner's own current value estimates at each step; no future information required & Uses learned estimates $\hat V_E,\hat Q_E,\hat V_I$ in practice; the formal guarantee additionally assumes the training algorithm executes only stable policies upon convergence, which is itself an implementation-level assumption & Optimal-policy set of the original MDP unchanged, conditional on convergence to stable policies and the tie/dominance conditions holding at every step \\
\addlinespace
BAMDP, meta-RL setting \citep{lidayan2025bamdp} & BAMDP state (history-based, belief-augmented) over the underlying RL problem & Meta-RL, across episodes or tasks; no separate terminal condition stated & Standard dynamic-PBRS pairing over the history, $\phi(h_t)$ and $\phi(h_{t-1})$ & Not restricted in principle; not independently confirmed for deep-RL implementation by this review & Potential-based on the BAMDP state is necessary and sufficient for the shaped BAMDP's optimal algorithm to remain Bayes-optimal for the underlying RL problem (G1) \\
\addlinespace
BAMDP, ordinary-RL setting \citep{lidayan2025bamdp} & Same BAMDP-based construction, evaluated against the ordinary RL objective & Ordinary RL; argument uses an episode-level boundary term & Same pairing, with the additional requirement that the potential is bounded and monotone increasing over training time & Not restricted in principle; not independently confirmed for deep-RL implementation by this review & Eventual approximate optimality only: for every $\epsilon>0$ a training step exists after which the shaped policy's unshaped return is within $\epsilon$ of optimal (G3, not exact recovery) \\
\end{longtable}
}

\section{Taxonomy}
\label{sec:taxonomy}

\begin{figure}[!htbp]
\centering
\resizebox{\textwidth}{!}{%
\begin{tikzpicture}
  \tikzset{
    tbox/.style  = {cell, minimum width=26mm, text width=23mm, minimum height=10mm,
                    align=center, fill=white},
    ibox/.style  = {cell, minimum width=15mm, text width=13mm, minimum height=10mm,
                    align=center, font=\scriptsize, fill=white},
    gbox/.style  = {cell, minimum width=26mm, text width=23mm, minimum height=10mm,
                    align=center, fill=white},
    hi/.style    = {fill=drsfill, draw=drsblue, very thick}
  }
  \node[axlabel, anchor=east] at (-0.4,0) {Temporal};
  \node[tbox]      (T1) at (1.4,0)  {T1 schedule};
  \node[tbox,hi]   (T2) at (4.2,0)  {T2 experience};
  \node[tbox]      (T3) at (7.0,0)  {T3 performance};
  \node[tbox]      (T4) at (9.8,0)  {T4 structure};
  \node[tbox]      (T5) at (12.6,0) {T5 interaction};

  \node[axlabel, anchor=east] at (-0.4,-1.7) {Source};
  \node[ibox]    (I1) at (0.85,-1.7)  {I1 designer};
  \node[ibox]    (I2) at (2.50,-1.7)  {I2 symbolic};
  \node[ibox,hi] (I3) at (4.15,-1.7)  {I3 own est.};
  \node[ibox]    (I4) at (5.80,-1.7)  {I4 advice};
  \node[ibox]    (I5) at (7.45,-1.7)  {I5 feedback};
  \node[ibox]    (I6) at (9.10,-1.7)  {I6 agents};
  \node[ibox]    (I7) at (10.75,-1.7) {I7 FMs};
  \node[ibox]    (I8) at (12.40,-1.7) {I8 rew.\ models};
  \node[ibox]    (I9) at (14.05,-1.7) {I9 world models};

  \node[axlabel, anchor=east] at (-0.4,-3.4) {Guarantee};
  \node[gbox,hi] (G1) at (1.4,-3.4)  {G1 exact};
  \node[gbox]    (G2) at (4.2,-3.4)  {G2 exact in the limit};
  \node[gbox]    (G3) at (7.0,-3.4)  {G3 other formal};
  \node[gbox]    (G4) at (9.8,-3.4)  {G4 objective-aware};
  \node[gbox]    (G5) at (12.6,-3.4) {G5 no guarantee};

  \node[plain, anchor=west] (ex) at (0.4,-4.9)
    {Worked example: bootstrapped shaping $=$ (T2,\;I3,\;G1)};
  \draw[flowb, dashed] (ex.north) -- (G1.south);
  \draw[flowb, dashed] (G1.north) -- (I3.south);
  \draw[flowb, dashed] (I3.north) -- (T2.south);
\end{tikzpicture}}
\caption{The three classifying dimensions of Section~\ref{sec:taxonomy}. A method is located by one value
(or a small set of values) on each axis; the shaded path shows a worked example, bootstrapped shaping
\citep{adamczyk2025bootstrapped}. The dimensions are designed to be largely independent, so their combinations define a design
space; sparsely populated regions of this space are as informative as the dense ones
(Section~\ref{sec:taxonomy-observations}, Figure~\ref{fig:landscape}).}
\label{fig:taxonomy}
\end{figure}
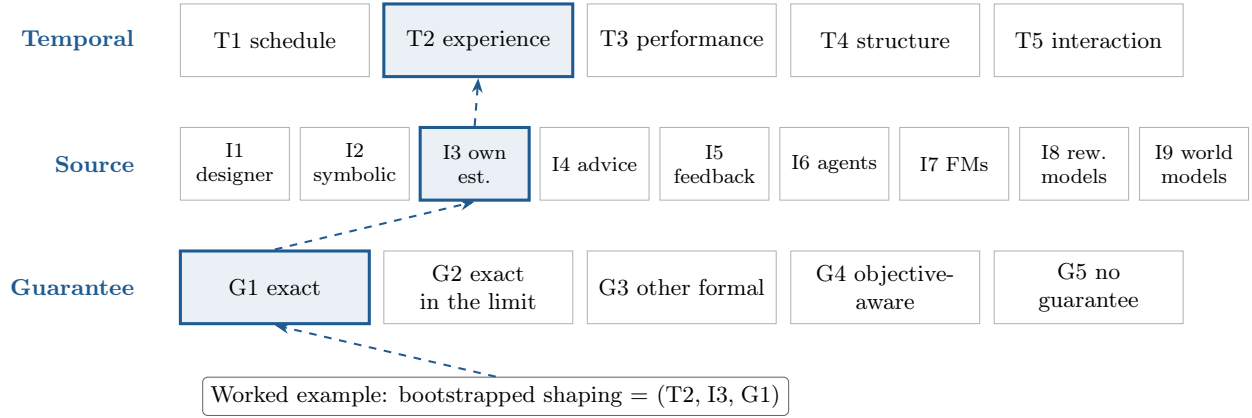

The taxonomy that follows, the temporal-signature, information-source, and guarantee-class dimensions, together with the C1--C4 mechanism classes of Section~\ref{sec:unified-framework}, is this review's own organising apparatus, not settled terminology from the reviewed field. Terms such as \emph{dynamic reward shaping proper}, \emph{reward replacement}, and \emph{reward-adjacent guidance} are definitions this review adopts for the classification that follows; a cited author's own use of ``shaping,'' ``intrinsic reward,'' or ``dynamic'' is not assumed to match the sense given here, and where the two senses could be confused the text says so explicitly (as in Section~\ref{sec:unified-framework}'s distinction between parametric revision and state dependence). Under this taxonomy, the reviewed methods are organised along three largely independent dimensions, summarised in Figure~\ref{fig:taxonomy}: how the shaping signal changes over training, what information drives the change, and what guarantee survives the adaptation. A method is located by one value, or a small set of values, on each dimension. The dimensions are designed to be largely independent; their combinations provide a design space for comparing dense and sparse regions of the reviewed candidate set.

\subsection{Dimension 1: temporal signature}
\label{sec:dim-temporal}

The temporal signature describes how the shaping signal changes and which event causes the change.

\begin{itemize}[leftmargin=*]
  \item \textbf{T1: Schedule-driven.} The signal follows a predetermined function of training progress, such as linear or exponential annealing or staged switching at fixed budgets. This open-loop adaptation does not depend on observed performance.
  \item \textbf{T2: Experience-driven.} The signal is determined by the agent's accumulated experience, including visitation counts, prediction errors, empirical success rates, or replay statistics. Adaptation occurs automatically without an explicit controller.
  \item \textbf{T3: Performance-driven.} The signal is adjusted according to measured task performance, typically through an outer optimisation loop that uses the task reward or another stated external reference criterion as its objective.
  \item \textbf{T4: Structure-driven.} The adaptive mechanism changes when an external structure advances, for example after an automaton transition, completion of a subgoal, progression to a new curriculum stage, or selection of the next planning step. The changed object may be an additive shaping term, a task state, or a task distribution; T4 does not by itself imply reward modification.
  \item \textbf{T5: Interaction-driven.} The signal changes in response to feedback from an external agent, usually a human trainer. The content of this feedback depends on the learner's current policy.
  \item \textbf{S: State-dependent, not parametrically revised.} The shaping function is fixed; only its argument varies, because that argument forms part of an augmented state such as a belief, a plan index, or an automaton configuration. Such methods are recorded for completeness, since they are frequently described as dynamic, but they are static shaping over an enlarged representation in the sense of Section~\ref{sec:unified-framework} and are excluded from the counts in Figure~\ref{fig:landscape}.
\end{itemize}

\subsection{Dimension 2: information source}
\label{sec:dim-source}

The information-source dimension identifies the origin of the content used to construct the shaping signal. Nine sources are distinguished: \textbf{I1}, designer specification, including manually defined potentials, weights, and schedules; \textbf{I2}, symbolic artefacts, such as plans, automata, temporal-logic specifications, and declared subgoals; \textbf{I3}, the agent's estimates, including value functions, prediction errors, and visitation statistics; \textbf{I4}, demonstrations, rankings, or advice; \textbf{I5}, live human or AI feedback; \textbf{I6}, other learning agents; \textbf{I7}, foundation models, including language and vision-language models; \textbf{I8}, learned reward or preference models; and \textbf{I9}, learned world models and latent dynamics. These sources instantiate the information state $\mathcal{I}_k$ in Eq.~\ref{eq:information-state}.

\subsection{Dimension 3: guarantee class}
\label{sec:dim-guarantee}

The guarantee class records the strongest property established for the adaptive mechanism in the stated setting. The five classes are mutually exclusive and separate formal preservation results from objective-aware optimisation heuristics. \textbf{G1} denotes exact structural preservation of the relevant optimum or equilibrium under the construction's stated assumptions, as in potential-based or another explicitly optimality-preserving form \citep{ng1999policy,devlin2012dynamic,forbes2024generalized,forbes2025action}. \textbf{G2} denotes asymptotic recovery of the \emph{exact} reference solution: the shaping or advice influence vanishes, or the adaptive process converges, so that the limiting policy or decision rule agrees with the unshaped reference problem \citep{behboudian2022policy,marom2018belief}. \textbf{G3} denotes another formal result that does not imply either G1 or G2, including an approximate, bounded, local, or special-case performance or convergence statement. Thus an eventual $\epsilon$-optimality result is G3 unless the theorem also proves convergence to the exact reference solution. \textbf{G4} denotes objective-aware adaptation without a formal result of the G1--G3 kinds. These methods use a reference return, held-out evaluator, or related performance signal to select or suppress shaping, but this optimisation choice does not itself prove soundness \citep{hu2020learning,gupta2023behavior}. \textbf{G5} denotes methods for which neither such a formal result nor objective-aware selection is established and whose justification is primarily empirical. Classification proceeds in that order, so no method is assigned both G2 and G3. For C2--C4 methods, a starred class records the corresponding method's own criterion and must not be read as a reward-shaping invariance result. A guarantee taxonomy of this kind must report the strongest guarantee that currently survives scrutiny, not the strongest guarantee a paper originally claimed for itself: where a subsequent work has shown an invariance argument to be incorrect, the entry is reclassified at the level that remains defensible and marked with a dagger, and it is counted at that corrected level wherever the review reports counts over the taxonomy, including Figure~\ref{fig:landscape}.

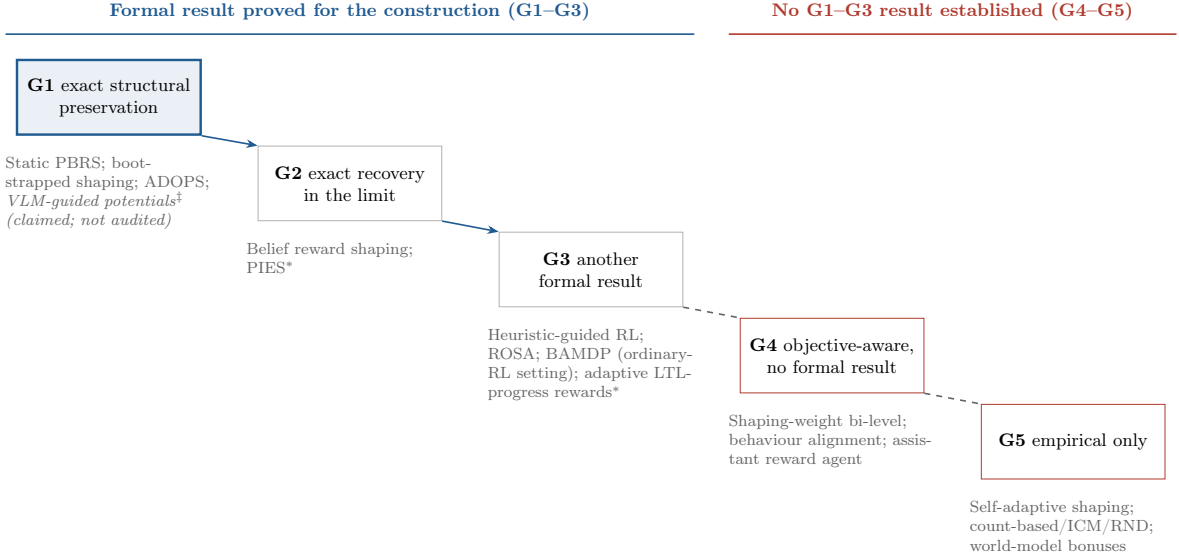
\begin{figure}[!htbp]
\centering
\resizebox{0.95\textwidth}{!}{%
\begin{tikzpicture}
  \tikzset{
    gstep/.style  = {cell, minimum width=32mm, text width=29mm, minimum height=13mm,
                    align=center, fill=white, font=\footnotesize},
    exbox/.style = {font=\scriptsize, drsgrey, align=left, text width=36mm, anchor=north}
  }
  \node[gstep, fill=drsfill, draw=drsblue, very thick] (g1) at (0,3.4)      {\textbf{G1} exact structural preservation};
  \node[gstep]                                          (g2) at (4.2,1.9)  {\textbf{G2} exact recovery in the limit};
  \node[gstep]                                          (g3) at (8.4,0.4)  {\textbf{G3} another formal result};
  \node[gstep, draw=drswarn]                             (g4) at (12.6,-1.1) {\textbf{G4} objective-aware, no formal result};
  \node[gstep, draw=drswarn]                             (g5) at (16.8,-2.6) {\textbf{G5} empirical only};

  \draw[flowb, thick] (g1.south east) -- (g2.north west);
  \draw[flowb, thick] (g2.south east) -- (g3.north west);
  \draw[draw=drsgrey, thick, dashed] (g3.south east) -- (g4.north west);
  \draw[draw=drsgrey, thick, dashed] (g4.south east) -- (g5.north west);

  \node[exbox] at (0,2.5)     {Static PBRS; bootstrapped shaping; ADOPS; \textcolor{drsgrey}{\textit{VLM-guided potentials$^{\ddagger}$ (claimed; not audited)}}};
  \node[exbox] at (4.2,1.0)   {Belief reward shaping; PIES$^{\ast}$};
  \node[exbox] at (8.4,-0.5)  {Heuristic-guided RL; ROSA; BAMDP (ordinary-RL setting); adaptive LTL-progress rewards$^{\ast}$};
  \node[exbox] at (12.6,-2.0) {Shaping-weight bi-level; behaviour alignment; assistant reward agent};
  \node[exbox] at (16.8,-3.5) {Self-adaptive shaping; count-based/ICM/RND; world-model bonuses};

  \draw[draw=drsblue, thick] (-1.8,4.5) -- (10.2,4.5);
  \node[font=\footnotesize\bfseries, drsblue] at (4.2,4.9) {Formal result proved for the construction (G1--G3)};
  \draw[draw=drswarn, thick] (10.8,4.5) -- (18.6,4.5);
  \node[font=\footnotesize\bfseries, drswarn] at (14.7,4.9) {No G1--G3 result established (G4--G5)};
\end{tikzpicture}}
\caption{The guarantee-class ladder of Section~\ref{sec:dim-guarantee}. Horizontal position is not a second independent variable; it encodes the same descending strength as vertical position, staggered so that the classes read as a staircase rather than a stacked list. The blue/red divide is the one that matters operationally: G1--G3 rest on a proof about the construction itself, while G4 and G5 rest on, respectively, an objective-aware selection procedure and no formal check at all. Method examples are illustrative, drawn from Table~\ref{tab:classification}; $^{\ast}$ marks a C2--C4 method whose guarantee is its own criterion rather than a reward-shaping invariance result, and $^{\ddagger}$, set in grey italics, marks a guarantee reported as claimed by its source rather than independently verified by this review, so that a claimed placement is visually distinct from an audited one (Table~\ref{tab:classification}).}
\label{fig:guarantee-ladder}
\end{figure}

\subsection{Classification of the reviewed methods}
\label{sec:classification}

Table~\ref{tab:classification} classifies the reviewed methods along the three dimensions, together with the mechanism class of Section~\ref{sec:unified-framework}, which determines whether the guarantee column expresses a reward-shaping invariance claim at all. The methods are grouped according to the families introduced in Section~\ref{sec:families}.

\paragraph{What a G1 label in this table does and does not certify.} A G1 entry means that a fixed or consistently time-indexed potential-based transformation, in the sense of Theorem~\ref{thm:pbrs} or Theorem~\ref{thm:dpbrs}, is what the cited paper proves preserves the optimal-policy set. It does not by itself mean that a particular online update rule for estimating that potential has been shown to converge safely, and it does not mean that a deep, off-policy implementation of the method with a replay buffer, target networks, and function approximation has been shown to preserve the theorem's conditions; those are separate claims, addressed where the evidence supports them in Sections~\ref{sec:fam-value} and \ref{sec:implementation}, and not established merely by invoking the structural theorem. A reader using the decision guide of Section~\ref{sec:decision-guide} to select a family should treat G1 there in this narrower, structural sense, and should still apply the implementation safeguards of Section~\ref{sec:implementation} before treating a deployed system as inheriting the guarantee.

{\small
\begin{longtable}{@{}p{0.31\linewidth}cccc p{0.31\linewidth}@{}}
\caption{Methods included in the review, by mechanism class (C1--C4, Section~\ref{sec:unified-framework}), temporal signature (T1--T5 or S, Section~\ref{sec:dim-temporal}), information source (I1--I9, Section~\ref{sec:dim-source}), and guarantee class (G1--G5, Section~\ref{sec:dim-guarantee}). Static potential-based shaping is included as a reference point. Entries marked S are state-dependent shaping over an augmented representation rather than parametrically revised shaping. A guarantee marked $^{\ast}$ belongs to a C2--C4 method: it records that method's own soundness criterion and is \emph{not} a reward-shaping policy-invariance claim. A guarantee marked $^{\dagger}$ was originally claimed at a stronger class by the cited paper but subsequently shown incorrect by later work; the entry is classified at the guarantee that currently survives scrutiny, not the one originally claimed, and is counted accordingly in Figure~\ref{fig:landscape}. A guarantee marked $^{\ddagger}$ is reported as claimed by the source without independent verification by this review, either because the work is an unreviewed preprint whose claim has not been through peer review, or because the source's own description raises a question, under this review's own criteria in Section~\ref{sec:implementation}, that its stated proof does not itself resolve; the specific reason is given where the entry is discussed. $^{\ddagger}$-marked entries are excluded from the counts in Figure~\ref{fig:landscape}, alongside the S entries and C2--C4 entries, because that figure reports the pattern of guarantees this review can independently stand behind. Only C1 entries are counted in Figure~\ref{fig:landscape}. Assignments follow the criteria defined in
Sections~\ref{sec:dim-temporal}--\ref{sec:dim-guarantee}.}
\label{tab:classification}\\
\toprule
\textbf{Method} & \textbf{Cls} & \textbf{Temp.} & \textbf{Src} & \textbf{Grt} & \textbf{Distinguishing feature} \\
\midrule
\endfirsthead
\multicolumn{6}{c}{\tablename\ \thetable\ (continued)}\\
\toprule
\textbf{Method} & \textbf{Cls} & \textbf{Temp.} & \textbf{Src} & \textbf{Grt} & \textbf{Distinguishing feature} \\
\midrule
\endhead
\midrule
\multicolumn{6}{r}{Continued on next page}\\
\endfoot
\bottomrule
\endlastfoot
\multicolumn{6}{@{}l}{\textit{Reference point}}\\
Static PBRS \citep{ng1999policy} & C1 & N/A & I1 & G1 & Fixed potential; equivalent to value initialisation \\
\midrule
\multicolumn{6}{@{}l}{\textit{\S\ref{sec:fam-schedule} Schedule-based}}\\
Scheduled auxiliary control \citep{riedmiller2018learning} & C4 & T1/T3 & I1/I3 & G4$^{\ast}$/G5$^{\ast}$ & Return-driven scheduler is objective-aware; random selection has no such check \\
Heuristic-guided RL \citep{cheng2021heuristic} & C1 & T1 & I4 & G3 & Horizon-controlled mixing with a heuristic \\
Reward Training Wheels \citep{wang2025reward} & C1 & T3 & I3 & G4 & Objective-aware teacher adapts auxiliary weights to student capability \\
\midrule
\multicolumn{6}{@{}l}{\textit{\S\ref{sec:fam-value} Value-derived potentials}}\\
Online shaping-reward learning \citep{grzes2010online} & C1 & T2 & I3 & G1 & Potential from an abstract-state value function \\
Bootstrapped shaping \citep{adamczyk2025bootstrapped} & C1 & T2 & I3 & G1 & Potential set to the current $V$ estimate \\
Exploration-guided shaping \citep{devidze2022exploration} & C1 & T2 & I3 & G3 & Shaping designed jointly with exploration \\
\midrule
\multicolumn{6}{@{}l}{\textit{\S\ref{sec:fam-bayes} Uncertainty-aware}}\\
Belief reward shaping \citep{marom2018belief} & C1 & T2 & I1/I3 & G2 & Reward priors that decay with experience \\
Self-adaptive shaping \citep{ma2025highly} & C1 & T2 & I3 & G5 & Beta-posterior success rates; empirical stability rather than a preservation theorem \\
POMDP potential shaping \citep{eck2016potential} & C1 & S & I3 & G1 & Belief-dependent potentials for online planning \\
\midrule
\multicolumn{6}{@{}l}{\textit{\S\ref{sec:fam-advice} Advice- and demonstration-derived}}\\
Plan-based shaping \citep{grzes2008plan} & C1 & S & I2 & G1 & Potential indexed by plan step \\
Dynamic potential-based advice \citep{harutyunyan2015expressing} & C1 & T2 & I4 & G5$^{\dagger}$ & Auxiliary value function as potential; originally claimed G1, subsequently refuted $^{\dagger}$ \\
PIES \citep{behboudian2020useful,behboudian2022policy} & C4 & T1 & I4 & G2$^{\ast}$ & Explicit advice with decaying influence \\
Subgoal-based shaping \citep{okudo2021subgoal} & C1 & S & I2 & G1 & Human-specified intermediate states \\
Self-supervised online shaping \citep{memarian2021self} & C2 & T2 & I3/I4 & G3$^{\ast}$ & Dense reward fitted to sparse-reward rankings \\
\midrule
\multicolumn{6}{@{}l}{\textit{\S\ref{sec:fam-human} Interactive human shaping}}\\
TAMER \citep{knox2009interactively} & C2 & T5 & I5 & G5$^{\ast}$ & Model of human evaluative reinforcement \\
Policy shaping \citep{griffith2013policy} & C4 & T5 & I5 & G5$^{\ast}$ & Feedback as policy labels and Bayesian fusion; evaluated empirically \\
COACH \citep{macglashan2017interactive} & C4 & T5 & I5 & G3$^{\ast}$ & Feedback treated as policy-dependent advantage \\
Deep TAMER \citep{warnell2018deep} & C2 & T5 & I5 & G5$^{\ast}$ & TAMER with deep function approximation \\
\midrule
\multicolumn{6}{@{}l}{\textit{\S\ref{sec:fam-im} Intrinsic motivation}}\\
Count-based / ICM / RND \citep{bellemare2016unifying,pathak2017curiosity,burda2019exploration} & C1 & T2 & I3 & G5 & Novelty bonuses decaying with familiarity \\
PBIM \citep{forbes2024pbim} & C1 & T2 & I3 & G1 & Conversion of bonuses to optimality-preserving form \\
GRM \citep{forbes2024generalized} & C1 & T2 & I3 & G1$^{\ddagger}$ & Broadens PBIM to history-dependent corrections; preprint, excluded from Figure~\ref{fig:landscape} pending peer review \\
ADOPS \citep{forbes2025action} & C1 & T2 & I3 & G1 & Action-dependent intrinsic returns permitted \\
BAMDP shaping \citep{lidayan2025bamdp} & C1 & T2 & I3 & \shortstack{G1 meta-RL\\G3 ordinary RL} & Exact Bayes-optimal preservation in meta-RL; formal eventual approximate preservation for bounded monotone potentials in ordinary RL \\
\midrule
\multicolumn{6}{@{}l}{\textit{\S\ref{sec:fam-bilevel} Bi-level and meta-optimised}}\\
Online reward design \citep{sorg2010reward} & C2 & T3 & I3 & G3$^{\ast}$ & Gradient ascent on reward parameters \\
Learned intrinsic rewards \citep{zheng2018learning} & C1 & T3 & I3 & G4 & Intrinsic reward trained for extrinsic return \\
Shaping-weight bi-level \citep{hu2020learning} & C1 & T3 & I1/I3 & G4 & Upper level optimises weights for true reward; no preservation theorem \\
Behaviour alignment \citep{gupta2023behavior} & C1 & T3 & I1/I3 & G4 & Objective-aware bi-level optimisation over the reward function \\
ROSA \citep{mguni2023learning} & C1 & T3 & I3 & G3 & Shaping as a two-player game \\
Assistant reward agent \citep{ma2024assistant} & C1 & T3 & I3 & G4 & Auxiliary reward agent co-trained with the policy agent to supply exploration-to-exploitation guidance \\
\midrule
\multicolumn{6}{@{}l}{\textit{\S\ref{sec:fam-structure} Structure-driven}}\\
Reward machines \citep{icarte2022reward} & C4 & S & I2 & n/a$^{\ast}$ & Automaton state exposed to the learner \\
LTL-based shaping \citep{camacho2019ltl,jiang2021temporal} & C1 & S & I2 & G1 & Potentials from temporal-logic specifications \\
Adaptive LTL-progress rewards \citep{kwon2025adaptive} & C2 & T3 & I2/I3 & G3$^{\ast}$ & Automaton-progress reward functions revised when measured success rate falls below a threshold; task-progression optimality proved for the construction \\
Curriculum-coupled shaping \citep{narvekar2020curriculum} & C4 & T1/T4 & I1 & G5$^{\ast}$ & Shaping advances with task difficulty \\
\midrule
\multicolumn{6}{@{}l}{\textit{\S\ref{sec:fam-preference} Preference-based and RLHF}}\\
Iterative RLHF \citep{ouyang2022training} & C2 & T3/T5 & I5/I8 & G5$^{\ast}$ & Reward model updated from policy-dependent preferences \\
Constitutional AI \citep{bai2022constitutional} & C2 & T3/T5 & I7/I8 & G5$^{\ast}$ & AI feedback generated from explicit principles \\
DPO \citep{rafailov2023direct} & C4 & T3 & I4/I8 & n/a$^{\ast}$ & Near neighbour: direct objective, no exposed additive reward \\
\midrule
\multicolumn{6}{@{}l}{\textit{\S\ref{sec:fam-world} World-model and latent-prediction}}\\
Predictive uncertainty / disagreement \citep{pathak2019disagreement,sekar2020planning} & C1 & T2 & I9 & G5 & Model uncertainty used as exploration or progress bonus \\
Latent novelty / imagined reachability \citep{fu2023gobi} & C1 & T2 & I9 & G5 & Shaping derived from evolving latent dynamics \\
\midrule
\multicolumn{6}{@{}l}{\textit{\S\ref{sec:fam-foundation} Foundation-model-driven}}\\
Eureka \citep{ma2024eureka} & C2 & T3 & I7 & G4$^{\ast}$ & Candidate rewards selected by measured task performance without a preservation theorem \\
Text2Reward \citep{xie2024text2reward} & C2 & T3 & I7 & G4$^{\ast}$ & Dense reward programs refined with evaluation feedback \\
LLM heuristics \citep{bhambri2024extracting} & C1 & T4 & I7 & G5 & LLM plans converted to shaping signals \\
VLM semantic rewards \citep{rocamonde2024vision,baumli2024vision} & C2 & T2/T4 & I7 & G5$^{\ast}$ & Visual-language similarity used as dense reward \\
VLM-guided potentials \citep{muller2026automating} & C1 & T2 & I7 & G1$^{\ddagger}$ & Potential learned from VLM preference labels; periodic updates recompute buffered rewards from the latest potential \\
Dense reward for free \citep{chan2024dense} & C3 & T2 & I4 & G1$^{\ast}$ & Redistribution is proved equivalent to potential-based shaping, preserving the optimum \\
\midrule
\multicolumn{6}{@{}l}{\textit{\S\ref{sec:fam-marl} Multi-agent}}\\
Potential-based difference rewards \citep{devlin2014potential} & C1 & T2 & I6 & G1 & Credit assignment plus equilibrium consistency \\
Counterfactual advantages \citep{foerster2018counterfactual} & C4 & T2 & I3/I6 & n/a$^{\ast}$ & Policy-dependent agent contribution signal \\
Value decomposition \citep{rashid2018qmix,wang2021qplex} & C4 & T2 & I3/I6 & n/a$^{\ast}$ & Learned individual utilities from team return \\
\end{longtable}
}

\paragraph{Basis of the audited G3 and G4 assignments.} Each G3 entry is tied to a formal result in the cited work: heuristic-guided RL proves solution of the original task under its horizon and heuristic conditions \citep{cheng2021heuristic}; exploration-guided shaping gives a theoretical result for a specified family of MDPs \citep{devidze2022exploration}; BAMDP shaping gives eventual $\epsilon$-optimality in ordinary RL under bounded monotone potentials \citep{lidayan2025bamdp}; ROSA proves convergence under its game and approximation assumptions \citep{mguni2023learning}; and the starred C2--C4 entries record, respectively, conditional optimal-policy agreement, convergence for the online reward-design objective, local convergence of the policy-gradient interpretation, and, for the adaptive LTL-progress construction, a proof that an optimal policy of the adaptively updated product-MDP reward eventually attains the best achievable task-progression value \citep{memarian2021self,sorg2010reward,macglashan2017interactive,kwon2025adaptive}. In each starred case the result concerns that construction's own criterion, task progression or policy agreement under the replacement reward, and is not a claim that the original task reward's optimal policy set is preserved. Each G4 entry instead uses measured task performance or an external evaluation criterion to revise or select the learned signal, but the cited work does not establish a G1--G3 preservation or recovery result \citep{riedmiller2018learning,wang2025reward,zheng2018learning,hu2020learning,gupta2023behavior,ma2024eureka,xie2024text2reward,ma2024assistant}. The assistant-reward-agent method trains a second agent to supply auxiliary reward to the policy agent and reports empirical gains in sample efficiency and stability, but no invariance or convergence theorem for the joint two-agent system \citep{ma2024assistant}. The audit therefore moves self-adaptive shaping and policy shaping to G5, distinguishes the random and return-driven scheduled-auxiliary variants as G5$^{\ast}$ and G4$^{\ast}$, and moves the exact return-preservation result of dense reward redistribution to G1$^{\ast}$ \citep{ma2025highly,griffith2013policy,chan2024dense}.

\subsection{Observations from the taxonomy}
\label{sec:taxonomy-observations}

The taxonomy reveals three patterns that are difficult to identify when methods are considered individually. The first is shown in Figure~\ref{fig:landscape}.

\begin{figure}[!htbp]
\centering
\resizebox{\textwidth}{!}{%
\begin{tikzpicture}
  \tikzset{
    c/.style     = {cell, minimum width=22mm, minimum height=10mm},
    chdr/.style  = {font=\footnotesize\bfseries, text=black, align=center,
                    text width=24mm},
    rhdr/.style  = {font=\footnotesize\bfseries, text=black, align=right,
                    text width=26mm, anchor=east}
  }
  \node[chdr] at (0,1.15)    {G1 exact};
  \node[chdr] at (2.6,1.15)  {G2 exact in the limit};
  \node[chdr] at (5.2,1.15)  {G3 other formal};
  \node[chdr] at (7.8,1.15)  {G4 objective-aware};
  \node[chdr] at (10.4,1.15) {G5 none};
  \node[rhdr] at (-1.35,0)    {T1 schedule};
  \node[rhdr] at (-1.35,-1.2) {T2 experience};
  \node[rhdr] at (-1.35,-2.4) {T3 performance};
  \node[rhdr] at (-1.35,-3.6) {T4 structure};
  \node[rhdr] at (-1.35,-4.8) {T5 interaction};
  \node[c, fill=white]        at (0,0)    {0};
  \node[c, fill=white]        at (2.6,0)  {0};
  \node[c, fill=drsblue!8]    at (5.2,0)  {1};
  \node[c, fill=white]        at (7.8,0)  {0};
  \node[c, fill=white]        at (10.4,0) {0};
  \node[c, fill=drsblue!30]   at (0,-1.2)   {6};
  \node[c, fill=drsblue!8]    at (2.6,-1.2) {1};
  \node[c, fill=drsblue!8]    at (5.2,-1.2) {2};
  \node[c, fill=white]        at (7.8,-1.2) {0};
  \node[c, fill=drsblue!24]   at (10.4,-1.2) {5$^{\dagger}$};
  \node[c, fill=drswfill, draw=drswarn, very thick] (t3g1) at (0,-2.4) {0};
  \node[c, fill=white]        at (2.6,-2.4) {0};
  \node[c, fill=drsblue!8]    at (5.2,-2.4) {1};
  \node[c, fill=drsblue!24]   at (7.8,-2.4) {5};
  \node[c, fill=white]        at (10.4,-2.4) {0};
  \node[c, fill=white]        at (0,-3.6)   {0};
  \node[c, fill=white]        at (2.6,-3.6) {0};
  \node[c, fill=white]        at (5.2,-3.6) {0};
  \node[c, fill=white]        at (7.8,-3.6) {0};
  \node[c, fill=drsblue!8]    at (10.4,-3.6) {1};
  \node[c, fill=white]        at (0,-4.8)   {0};
  \node[c, fill=white]        at (2.6,-4.8) {0};
  \node[c, fill=white]        at (5.2,-4.8) {0};
  \node[c, fill=white]        at (7.8,-4.8) {0};
  \node[c, fill=white]        at (10.4,-4.8) {0};
  \node[warn, text=black, anchor=west, text width=44mm, align=left] (ann) at (12.0,-2.4)
    {The highlighted cell is empty: no included method both selects its shaping signal by measured return and preserves exact optimality.};
\end{tikzpicture}}
\caption{Distribution over temporal signature and guarantee class of the \textbf{C1 entries} of
Table~\ref{tab:classification}, that is, of dynamic reward shaping proper. C2--C4 methods are excluded because
their guarantee classes are not reward-shaping invariance claims, and the four S entries are excluded
because their shaping functions are not parametrically revised (Section~\ref{sec:unified-framework});
the static reference row is also excluded. Two further C1, T2, G1 entries, GRM and VLM-guided
potentials, are excluded because this review has not independently verified their guarantee claim
(marked $^{\ddagger}$ in Table~\ref{tab:classification}); the T2, G1 cell reports the six entries this review
can stand behind. Twenty-one entries remain. A method carrying a combined
temporal or setting-dependent guarantee label is counted in each cell it spans. Three features are visible. Exact optimality
preservation occurs only under experience-driven adaptation. Performance-driven adaptation, the
category that explicitly selects shaping against a measured reference-performance criterion, never
coincides with it in this set. And
interaction-driven adaptation does not appear as additive shaping among the included methods: it is
realised as reward replacement or as policy-level guidance, both outside C1.
These are properties of the reviewed set and not proofs of impossibility; they may reflect the scope
and composition of the literature considered here.
$^{\dagger}$One T2 entry, dynamic potential-based advice, originally claimed G1 but its invariance
argument was subsequently shown incorrect (Section~\ref{sec:fam-advice}); it is counted here under G5,
the guarantee that currently survives scrutiny, not under the class originally claimed for it.}
\label{fig:landscape}
\end{figure}

\paragraph{In the C1 methods included in Table \ref{tab:classification}, no performance-driven method is assigned G1.} Among dynamic reward shaping proper, every G1 entry adapts through experience-driven read-outs (T2). The structure-driven methods that appear at first to belong here include plan-based, subgoal-based, and automaton-derived potentials. On inspection, these methods are state-dependent rather than parametrically revised and are therefore recorded as S. Performance-driven methods (T3) select or revise the shaping signal against a declared reference performance measure, and so can register that it is harmful, but this objective-aware selection is classified as G4 rather than as a preservation guarantee; no included T3 method attains G1. The empty T1 column is a similar artefact of selection rather than a theoretical barrier: Section~\ref{sec:fam-schedule} shows that a time-varying weight is G1-compatible when it is incorporated into a time-indexed potential and paired consistently as $\gamma w_{k+1}\Phi(s')-w_k\Phi(s)$, so schedule-driven adaptation and G1 are not theoretically incompatible. No schedule-driven method selected for Table~\ref{tab:classification} uses that consistently paired construction; the auxiliary objectives reviewed there multiply a static term by a schedule rather than time-indexing the potential itself, which is why they fall to G4 or G5 instead.

This is an observation about the reviewed literature, not a theorem. The two properties are not formally incompatible, and the absence may reflect the boundaries of this review as much as the state of the field; establishing it as a field-wide property would require the broader compliance audit identified in Section~\ref{sec:agenda}. What the included methods do show is a plausible mechanism for the gap: an outer loop that selects a shaping signal according to its measured effect on learning is not thereby constrained to produce a signal of invariant form, and no method reviewed here imposes both constraints simultaneously.

A distinct route to the same objective is at least claimed in Table~\ref{tab:classification}. VLM-guided potential learning \citep{muller2026automating}, a preprint, reports drawing on an expressive and semantically rich information source while remaining in G1, because the object that is learned is the \emph{potential} rather than the reward. This method is experience-driven rather than performance-driven, so if the claim holds it is not an exception to the pattern described above. The guarantee cell for this entry is marked $^{\ddagger}$ in Table~\ref{tab:classification}, however: it is reported as claimed rather than independently verified, for reasons discussed in Section~\ref{sec:fam-foundation}. With that qualification, it illustrates the design pattern most likely to close the gap if the claim is upheld: constrain the learned object to an invariant form, and allow any information source to supply its content.

\paragraph{Within Table~\ref{tab:classification}, I3, the agent's own estimates, appears across more included method families than any other single information-source category.} The inclusion of sources I8 and I9 reveals parallel feedback loops in reward-model and world-model methods. Value-derived potentials, intrinsic motivation, posterior success rates, and meta-learned intrinsic rewards are all derived from the learner's internal state; they differ primarily in the statistic used. This is an observation about the methods reviewed here, not a claim that I3 dominates the entire field. It highlights a recurring pattern in the reviewed methods: an adaptive signal is derived from estimates maintained by the learner that subsequently consumes that signal. Within the reviewed works, the guarantees are construction-specific; no general stability characterisation for this feedback pattern is identified.

\paragraph{Interaction-driven guidance is distinct within the reviewed set.} Among the included families, T5 is the case in which non-stationarity is supplied by an external trainer rather than generated solely within the learning system. \citet{macglashan2017interactive} showed that human evaluative feedback is policy-dependent. Trainers tend to reward improvement relative to the learner's current competence rather than absolute performance. Such feedback is mathematically distinct from an MDP reward and is more appropriately represented as an advantage. The same distinction may apply to other signals defined relative to current behaviour, but the present review does not treat that possibility as an established property of the broader literature.

\paragraph{Status of the framework.} The taxonomy and the C1--C4 mechanism classification introduced in this review constitute an analytical synthesis of the reviewed literature rather than terminology adopted from individual source papers. They are intended to provide a consistent conceptual framework for relating methods that have previously been studied in isolation.

\section{Method Families}
\label{sec:families}

The families below group methods by their dominant source or mode of guidance rather than by a single mathematical mechanism. A family may therefore contain additive shaping (C1), reward replacement (C2), reward redistribution (C3), and reward-adjacent guidance (C4) when those mechanisms use the same information source or serve the same functional role. The C1--C4 labels in Table~\ref{tab:classification} remain authoritative: family membership supports comparison across related approaches, but it does not imply that all members inherit the same policy-invariance result. In particular, guarantees attached to C2--C4 methods concern their own objective or soundness criterion and must not be read as guarantees for additive reward shaping.

\subsection{Schedule-based shaping}
\label{sec:fam-schedule}

The simplest dynamic approach retains a fixed shaping term while varying its weight according to a predetermined schedule. Scheduled auxiliary objectives have been used in robotic manipulation, while adaptive variants modify auxiliary reward weights as the learner's capability changes \citep{riedmiller2018learning,wang2025reward}. Such designs allow auxiliary guidance to dominate early training and place greater emphasis on the task objective as competence develops. The boundary between open-loop and closed-loop scheduling is not sharp. Scheduled auxiliary control, for example, may select among auxiliary intentions either at random or through a scheduler learned from main-task returns; the first variant is open-loop (T1) whereas the second is performance-driven (T3) under the definitions in Section~\ref{sec:dim-temporal} \citep{riedmiller2018learning}. Systems of this kind are therefore recorded as T1/T3 in Table~\ref{tab:classification}. A time-varying weight does not preserve invariance merely by multiplying a static PBRS term. To inherit the dynamic result, the weight must be incorporated into a time-indexed potential and paired consistently, yielding $\gamma w_{k+1}\Phi(s')-w_k\Phi(s)$ rather than $w_k[\gamma\Phi(s')-\Phi(s)]$. The auxiliary objectives used in these representative systems are not constructed to satisfy this condition; return-driven selection is therefore G4, whereas variants without an objective-aware check are G5.

Two developments extend this family beyond manual tuning. \citet{cheng2021heuristic} formalise the combination of a task objective and a heuristic value function through a horizon-controlled trade-off. Their analysis clarifies the role of the schedule and provides bounds that relate the quality of the heuristic to the resulting performance. \citet{wang2025reward} replace the fixed schedule with a teacher-student loop. The teacher adjusts auxiliary reward weights according to the student's evolving capability, and improvements over manually designed rewards are reported in simulated navigation and physical off-road driving. This method moves from T1 to T3 because open-loop scheduling is replaced by closed-loop control of the shaping signal. Under the taxonomy used here, it extends schedule-based shaping by replacing open-loop scheduling with performance-driven adjustment.

\subsection{Value-derived potentials}
\label{sec:fam-value}

A potential assigns a scalar shaping value to each state. A natural construction is therefore to derive it from a value function that is updated during training. An early example was provided by \citet{marthi2007automatic}, who automatically constructed shaping rewards by decomposing the reward function and solving an abstract version of the task. This work established that a potential could be computed rather than specified manually. Dynamic methods extend this idea by recomputing the potential throughout training. \citet{grzes2010online} learn a value function over an abstract state space online and use it as a potential over the original state space. The resulting signal captures structure discovered during learning rather than knowledge supplied in advance. \citet{adamczyk2025bootstrapped} use the agent's current state-value estimate directly as the potential. They provide convergence results for the tabular setting, analyse the resulting deep-RL dynamics, and report faster training across the Atari suite. From an exploration perspective, \citet{devidze2022exploration} jointly design the shaping signal and exploration mechanism for sparse-reward tasks.

A potential advantage of this family is that it can derive guidance from quantities learned by the agent, rather than requiring a manually specified potential. The signal can provide denser feedback derived from information already acquired by the learner, rather than introducing external information. These methods create a feedback loop because the shaping signal is derived from an estimate that is itself updated using shaped experience. The cited tabular analyses do not, by themselves, establish how these methods behave when combined with nonlinear function approximation, target networks, and off-policy replay. As practical safeguards, one may compute the potential from a target network and limit potential-update frequency relative to critic updates; these strategies are design suggestions rather than established solutions.

\subsection{Uncertainty-aware and Bayesian shaping}
\label{sec:fam-bayes}

When a shaping signal is estimated, its uncertainty provides information about how strongly it should influence learning. Methods in this family use the confidence of the estimate to control that influence. \citet{marom2018belief} augment the reward distribution with prior beliefs whose contribution decreases with experience. Under suitable conditions for Q-learning, the resulting augmented MDP is shown to be consistent with the optimal policy of the original MDP. The reduction in influence is not imposed through a schedule. Instead, the posterior becomes concentrated as observations accumulate, causing the contribution of the prior to vanish. The method is therefore G2 by construction. \citet{ma2025highly} apply the same principle to the magnitude of shaping rewards. Success rates are sampled from Beta distributions that evolve from uncertain to reliable as data are collected. Consequently, shaping is stochastic and exploratory during early training and becomes more decisive later. Kernel density estimation with random Fourier features makes the construction tractable in high-dimensional continuous spaces without requiring a learned model. In partially observable planning, \citet{eck2016potential} incorporate belief-state information into the potential. This last case sits at the boundary drawn in Section~\ref{sec:unified-framework}: a fixed function $\Phi(b)$ over beliefs varies at every step because the belief does, which is state dependence rather than parametric revision, and the method is recorded as S in Table~\ref{tab:classification}. Where a planner additionally re-estimates the potential from the belief it is refining, both mechanisms operate at once, and the classification notes the ambiguity.

A distinguishing feature of this family is that the influence of shaping is adjusted through a statistical update rule rather than a manually specified schedule. Its limitation is shared with Bayesian RL more generally. The prior remains a modelling choice; posterior adaptation alone does not ensure that an initially misspecified prior becomes unimportant quickly enough for a particular learning problem.

\subsection{Advice- and demonstration-derived shaping}
\label{sec:fam-advice}

This family converts an external artefact into a shaping signal during learning. The artefact may be a plan, a set of subgoals, expert advice, or a ranking over trajectories.

\citet{grzes2008plan} derive potentials from a symbolic plan and index the potential by the agent's current position in that plan. The shaping signal therefore advances with plan execution. However, as with the automaton constructions of Section~\ref{sec:fam-structure}, what advances is the argument of a fixed rule: the plan index is part of an augmented state, the potential over that augmented state does not change, and the method is recorded as S rather than as parametrically dynamic. \citet{okudo2021subgoal} further reduce the specification burden by requesting intermediate \emph{states} from human users instead of abstract potentials. Such states are easier to elicit and can be converted directly into a potential; the same classification applies.

An important development in this family concerns arbitrary advice. \citet{harutyunyan2015expressing} proposed dynamic potential-based advice (DPBA). An auxiliary value function is learned from an arbitrary advice reward and is then used as the potential, with a claimed policy-invariance guarantee. Subsequently, \citet{behboudian2020useful,behboudian2022policy} showed both theoretically and empirically that DPBA can alter the optimal policy. They reported that the correction required to restore invariance reduced much of DPBA's empirical benefit in their evaluated settings \citep{behboudian2020useful,behboudian2022policy}. Their alternative, Policy Invariant Explicit Shaping (PIES), does not modify the reward. Advice is applied directly during action selection, and its influence is gradually reduced to zero. Under the method's stated conditions, the agent converges to an optimal policy even when the advice is misleading, while informative advice can still accelerate learning \citep{behboudian2022policy}. This case illustrates the need to examine invariance claims carefully. It also exemplifies a recurring trade-off between representational convenience and unconditional guarantees.

An alternative construction was proposed by \citet{memarian2021self}, who remove the need for an external artefact. The sparse task reward acts as a self-supervisory signal by ranking observed trajectories according to their returns. A dense reward is fitted to these rankings and alternately updated with the policy. The resulting dense signal is derived entirely from sparse evidence, and conditions are provided under which it does not change the optimal policy.

\subsection{Interactive human shaping}
\label{sec:fam-human}

Evaluative feedback supplied by a human observer constitutes shaping in its original behavioural sense. This family provides computational methods for incorporating such feedback. \citet{knox2009interactively} introduced TAMER, in which the agent models human evaluative reinforcement and selects actions that maximise the predicted human reward. \citet{warnell2018deep} extended the method to high-dimensional state spaces using deep function approximation, enabling pixel-based tasks to be trained from short human-feedback sessions.

The central theoretical contribution of this family is the recognition that human feedback is not an MDP reward. \citet{griffith2013policy} argued that evaluative feedback is more appropriately interpreted as information about the \emph{policy}. Their Bayesian method, Advise, treats feedback as direct policy labels and combines it with the agent's own learning process. \citet{macglashan2017interactive} further showed that feedback is policy-\emph{dependent}. Trainers assess improvement relative to the learner's current behaviour, and the same action may therefore receive different feedback at different levels of competence. COACH consequently interprets feedback as an advantage rather than a reward and converges under this interpretation. The broader literature on human advice is reviewed by \citet{najar2021reinforcement}.

In the mechanism classification of Section~\ref{sec:unified-framework} this family is heterogeneous, and the table records it as such: TAMER and Deep TAMER are reward replacement (C2), since the agent maximises predicted human reinforcement in place of a task reward, while policy shaping and COACH are reward-adjacent guidance (C4), acting on the policy and the gradient respectively. None of the four is additive shaping, and none contributes to the counts of Figure~\ref{fig:landscape}.

This family provides the clearest example of interaction-driven non-stationarity. Its central insight also applies beyond human feedback. Any shaping signal defined relative to the learner's current competence, including most value-derived and intrinsic-motivation signals, is closer to an advantage than to a stationary reward. Whether treating competence-relative signals as rewards, rather than advantages, contributes to instability in other adaptive-guidance methods remains an open question.

\subsection{Intrinsic motivation as dynamic shaping}
\label{sec:fam-im}

Exploration bonuses satisfy Definition~\ref{def:drs-general} even when they are not described in those terms: a visitation count or a predictor network is a parameter $\theta_k$ revised by an explicit update rule as experience accumulates, which is what the definition requires, regardless of whether the original authors intended their method to be read as shaping. This review therefore includes them under its broad C1 convention, while flagging that the inclusion is a consequence of the definition rather than a claim that count-based, curiosity, or novelty-distillation bonuses were constructed as a deliberately revised shaping rule $F_{\theta_k}$ in the same operational sense as an online potential update; a reader who reserves ``dynamic shaping'' for the latter, narrower sense should read this family as adaptive intrinsic-reward mechanisms captured by, rather than exemplifying, the review's convention. Count-based exploration \citep{bellemare2016unifying}, prediction-error curiosity \citep{pathak2017curiosity}, and random network distillation \citep{burda2019exploration} assign intrinsic rewards from novelty-related quantities, including visitation estimates, prediction error, and distillation error. In each case, the bonus typically decreases as the relevant state or observation becomes more familiar to the agent. These methods are not potential-based and can change the optimal policy. An uncorrected novelty bonus can favour novelty over task completion, thereby altering the optimal policy.

Recent work has sought to reconcile intrinsic motivation with optimality preservation, as reviewed in Section~\ref{sec:beyond-pbrs}. PBIM converts intrinsic rewards into potential-based form \citep{forbes2024pbim}. GRM enlarges the preserved function class and subsumes potential-based shaping \citep{forbes2024generalized}. ADOPS relaxes the action-independence requirement, which allows corrections to remain practical in long-episode, exploration-intensive domains such as Montezuma's Revenge \citep{forbes2025action}. The BAMDP formulation of \citet{lidayan2025bamdp} represents all pseudo-rewards as shaping over knowledge states. It identifies conditions under which a class of potential-based pseudo-rewards preserves the relevant Bayes-optimal objective and under which adverse effects are bounded. These constructions provide formal preservation results for selected forms of intrinsic reward, extending beyond the usual empirical treatment of novelty and curiosity bonuses.

\subsection{Bi-level and meta-optimised shaping}
\label{sec:fam-bilevel}

When the utility of a shaping signal is uncertain, its effect can be measured and the signal can be adjusted accordingly. This idea is formulated as a bi-level problem. The lower level optimises the policy using the shaped reward, whereas the upper level optimises the shaping signal according to the true task return.

The general template was established by \citet{sorg2010reward} using online gradient ascent over reward parameters with convergence guarantees. Their optimal-rewards formulation treats the designer's reward as a free variable selected to improve the performance of a bounded agent. In a similar setting, \citet{zheng2018learning} derived an algorithm that learns intrinsic rewards for policy-gradient agents. \citet{hu2020learning} address reward shaping directly. They note that human knowledge may be translated imperfectly into a numerical reward because of cognitive bias and other factors, so fully applying a supplied shaping function can be harmful. They derive the gradient of expected true reward with respect to the parameters of a shaping-weight function and propose three algorithms under different assumptions. Their results indicate that useful shaping can be exploited, harmful shaping can be ignored, and some initially unhelpful signals can be transformed into useful ones. \citet{gupta2023behavior} generalise the optimisation from shaping weights to the reward function itself. Their behaviour-alignment reward functions combine designer heuristics with the primary reward. They also identify settings in which potential-based shaping substantially impairs performance, showing that policy invariance does not imply harmless learning dynamics. A related effect is obtained by \citet{mguni2023learning} through a two-player formulation involving a shaping player and a policy player. \citet{ma2024assistant} extend the two-agent idea into a fully learned assistant reward agent. A policy agent and a reward agent interact with the environment simultaneously: the reward agent observes the policy agent's trajectory and produces a dense auxiliary reward that is added to the sparse task reward, and is itself updated by an independent reinforcement-learning process trained to improve the policy agent's environment return. The resulting signal is therefore performance-driven (T3) and sourced from the learner's own trajectory (I3), matching the template of \citet{sorg2010reward} and \citet{zheng2018learning} but replacing the single upper-level optimisation with a second interacting agent. Reported gains in sample efficiency and training stability across sparse-reward continuous-control tasks are empirical; the paper does not establish a preservation or convergence guarantee for the joint two-agent system, and the method is recorded as G4 in Table~\ref{tab:classification}.

This family is relevant when supplied prior knowledge may be unreliable, because its outer objective can evaluate the effect of the shaping signal against a reference criterion. Its principal limitation is computational cost. Differentiating through the learning process, or approximating the corresponding gradient, introduces an additional optimisation level \citep{hu2020learning,gupta2023behavior}. Moreover, when the upper-level signal is estimated from performance differences between training runs, random-seed variation can complicate its interpretation. This concern follows the broader reproducibility challenges documented for deep RL \citep{henderson2018deep}.

\subsection{Structure-driven shaping}
\label{sec:fam-structure}

When the temporal structure of a task can be specified formally, the shaping signal can be derived from that structure and updated automatically. Reward machines \citep{icarte2022reward} represent reward-function structure as a finite-state machine, allowing its decomposition to be used directly by the learner. \citet{camacho2019ltl} connect temporal-logic specifications to reward functions and derive shaping signals from the resulting automata. Temporal-logic-based shaping for continuing tasks was developed by \citet{jiang2021temporal}. Two distinctions matter for the classification of this family. First, the reward-machine framework is a representation of reward structure, not in itself a shaping method: exposing the automaton state as part of the learner's input supports at least three distinct uses, of which shaping is only one. \emph{Task decomposition} follows because each automaton state can be treated as defining a separate sub-task with its own local reward, letting the learner be organised hierarchically around the machine's structure. \emph{Counterfactual experience generation} follows because a single environment transition can be relabelled against every automaton state it would have produced a transition from, not only the one actually visited, so off-policy algorithms can synthesise additional training signal from each real transition without further environment interaction \citep{icarte2022reward}. \emph{Automated reward shaping} is the third use, and the one this review is concerned with: a potential can be defined directly over the automaton state and combined with the environment state as described above. These three uses share the same underlying representation but are logically independent, and a system that only decomposes or replays counterfactually need not shape the reward at all. The optimality-preservation claim attaches to the specific automaton-derived potential construction \citep{camacho2019ltl}, not to the framework as a whole, and the framework itself is recorded as reward-adjacent in Table~\ref{tab:classification}. Second, once the automaton state is included in the state representation, the potential is a fixed function of that augmented state. What varies during an episode is the argument, not the shaping rule. By the distinction drawn in Section~\ref{sec:unified-framework} these are therefore examples of \emph{static} shaping over an augmented representation, and they inherit their guarantee from Theorem~\ref{thm:pbrs} applied to that representation rather than from any result about revision during learning. We record them as S accordingly. The family can nonetheless combine structural preservation results with interpretable task-progress representations: the active shaping objective can be inspected and compared with the formal specification. Curriculum-coupled shaping \citep{narvekar2020curriculum} provides a less formal counterpart in which shaping advances with task difficulty.

A distinct development in this family moves beyond a fixed potential over automaton state. \citet{kwon2025adaptive} define, for a task specified by a Linear Temporal Logic formula, a family of reward functions over the product MDP of environment state and automaton progress, and revise the parameters of that reward function online: after every fixed block of episodes, if the agent's measured success rate falls below a threshold, the reward function is updated to better reflect achievable task progression. Because the resulting signal is a self-contained reward over the augmented state rather than an additive term alongside a separately available task reward, it is classified as C2, adaptive reward replacement, rather than as dynamic shaping proper; unlike the automaton-indexed potentials above, the revision is genuinely parametric rather than merely state-dependent. The paper proves a task-progression optimality result for the construction: an optimal policy under the reward function reached after finitely many revisions attains the best task progression achievable under the specification. This result concerns the constructed replacement reward's own criterion and is recorded as G3$^{\ast}$ in Table~\ref{tab:classification}; it is not a claim that the optimal policy of a separately specified task reward is preserved.

The principal constraint is the availability of a suitable formal specification. When no temporal-logic formula, reward machine, or comparable task representation is available, one must be constructed before structure-driven shaping can be applied. This knowledge-engineering requirement motivates learned-shaping methods that infer guidance from data or feedback.

\subsection{Preference-based reward learning and RLHF}
\label{sec:fam-preference}

Preference-based reinforcement learning replaces a manually specified scalar reward with a model learned from comparisons, rankings, critiques, or principles \citep{christiano2017deep,wirth2017survey}. In the standard RLHF pipeline, demonstrations are used to initialise a policy, pairwise judgements are used to train a reward model, and the policy is subsequently optimised against the learned reward \citep{ouyang2022training}. The signal becomes dynamic when preferences are collected iteratively, the reward model is retrained as the policy distribution changes, or the policy and evaluator evolve jointly. Constitutional AI makes the source of adaptation explicit. Critiques and preferences are generated from a written set of principles and are then used for reinforcement learning from AI feedback \citep{bai2022constitutional}. Natural-language feedback and iterative refinement provide richer update channels than scalar comparisons alone \citep{scheurer2022training}.

Direct preference optimisation (DPO) removes the separately deployed reward model and instead optimises a closed-form preference objective \citep{rafailov2023direct}. It is therefore not reward shaping in the narrow implementation sense because no additive reward is presented to the learner. Nevertheless, DPO is an important boundary case. It implicitly represents a reward relative to a reference policy and addresses the same problem of translating evolving preference information into policy improvement. Equating RLHF with DPO would obscure this implementation distinction, whereas treating them as unrelated would overlook their shared information sources and failure modes.

The primary risk is over-optimisation of the reward model. A model that is accurate on the comparison distribution may be exploited by a policy that moves outside that distribution, resulting in a high proxy reward but lower human utility \citep{gao2023scaling,eisenstein2024helping,miao2024inform}. Iterative reward refinement can reduce this distribution shift, but it also creates a non-stationary objective and a coupled evaluator-policy feedback loop. In the taxonomy, this family is class C2, reward replacement: the learned model substitutes for the task reward, so its G5 marking records the absence of a preservation or recovery guarantee \emph{for that substitution} and is not a shaping-invariance claim, which is why it carries the $^{\ast}$ qualifier in Table~\ref{tab:classification}. Explicit RLHF with an updated reward model is T3/T5 over sources I5/I8, with constitutional and AI-feedback variants additionally using I7. DPO is class C4: it changes the optimisation objective directly, no reward of any kind is presented to a learner, and it is retained in the table only as a boundary case whose guarantee cell is n/a and which is excluded from all counts. This placement connects alignment practice to dynamic shaping while keeping the unit of analysis stable.

\subsection{World-model and latent-prediction shaping}
\label{sec:fam-world}

World models learn latent dynamics and use predicted futures to improve control. Their primary purpose is planning or policy learning from imagined experience rather than reward shaping, so a clear scope boundary is required. MuZero learns a latent model that predicts policy, value, and reward for planning without reconstructing observations \citep{schrittwieser2020mastering}. EfficientZero improves data efficiency through self-supervised consistency and value-prefix prediction \citep{ye2021mastering}. DreamerV3 learns behaviours from imagined latent trajectories across a range of domains \citep{hafner2025mastering}. The presence of a learned reward predictor alone does not make these methods instances of dynamic reward shaping.

These methods enter the present taxonomy when model-derived quantities are added to the task reward, redistribute it, or cause it to adapt. Model disagreement has been used as an intrinsic exploration reward \citep{pathak2019disagreement}; world models have been used to plan toward expected novelty \citep{sekar2020planning}; and imagined reachability expansion has been converted into intrinsic reward \citep{fu2023gobi}. Such terms can be represented in the general form
\begin{equation}
F_k(s,a,s') = g\!\left(\mathcal{M}_k, z_k, a, z_{k+1}, \sigma_k\right),
\label{eq:world-shaping}
\end{equation}
In Eq.~\ref{eq:world-shaping}, $\mathcal{M}_k$ denotes the current world model, $z_k$ is a latent state, and $\sigma_k$ measures uncertainty or disagreement. Because all three quantities change as data are collected, the resulting signal is classified as T2 with I9. Model-predictive relabelling may instead be T3 when an outer loop selects shaping parameters according to a stated reference performance measure.

A recent development within this family reframes what is being adapted. Rather than producing a single scalar bonus, \citet{li2026slope} construct a full potential landscape from optimistic distributional regression over the world model's predictions, so that the signal supplied to the learner has a shape, not only a magnitude, that is tuned to the current training regime; the stated motivation is to supply usable gradient in regions where a scalar bonus would be flat and uninformative. This is a preprint at the time of writing and this review has not verified whether the construction is stated or proved to be potential-based in the sense of Theorem~\ref{thm:pbrs}, so it is discussed here as an illustration of the family's direction, adapting the geometry of the shaping signal rather than only its scale, and is not assigned a row in Table~\ref{tab:classification}.

Latent-model quantities such as predicted novelty and reachability can supply an auxiliary signal before external task reward is observed \citep{sekar2020planning,fu2023gobi}. This capability also creates identifiable risks. A bonus derived from an inaccurate or evolving model can represent model error rather than task progress, and recomputing the bonus can assign different shaped rewards to the same stored transition. These consequences follow from using the learned model as the source of $F_k$ and are treated here as failure modes requiring explicit evaluation. Exact policy invariance is unavailable unless the model-derived quantity is converted into a valid time-indexed potential. World-model shaping should therefore be understood as an information source and adaptation mechanism. It does not imply that model-based RL as a whole is a form of reward shaping.

\subsection{Foundation-model-driven shaping}
\label{sec:fam-foundation}

A recent family of methods delegates aspects of reward construction, interpretation, or evaluation to pretrained foundation models. Code-generating language models can synthesise executable reward programs. Eureka alternates among reward-code generation, large-scale policy evaluation, and language-model reflection \citep{ma2024eureka}. Text2Reward grounds code generation in environment descriptions and supports refinement based on feedback \citep{xie2024text2reward}. Other systems convert language-model plans or heuristics into intermediate guidance \citep{bhambri2024extracting}. The dynamic component varies across these systems. In one-shot methods it is limited to the generation of an initial reward, whereas in iterative methods generation, evaluation, critique, and regeneration form an outer loop whose output changes in response to observed policy performance.

This family extends beyond text-only generation. Vision-language models can assess whether observations satisfy natural-language goals and thereby provide dense visual rewards without task-specific classifiers \citep{rocamonde2024vision,baumli2024vision}. Multimodal evaluators can combine images, language, demonstrations, and proprioceptive traces. Program-synthesis methods can compile task descriptions into verifiable predicates or executable reward code. Code-generating agents can also revise coefficients, conditions, and subgoals after unsuccessful training runs. These mechanisms occupy T3/T4 with I7 and may additionally use I5 when human users critique the generated rewards.

The ability to generate or revise reward code does not, by itself, establish policy preservation, optimisation stability, or robustness to reward exploitation. Generated dense rewards belong to G4 when a reference performance measure is used to select among candidates without a preservation theorem, and to G5 when no such objective-aware check is present. Generated dense rewards can introduce exploitable terms, inappropriate scaling, discontinuities, unintended terminal incentives, or profitable cycles; these possibilities require explicit verification against the task specification. Selection based on short policy-training runs can be sensitive to random-seed variation and proxy over-optimisation \citep{henderson2018deep,gao2023scaling}. A foundation model may also reproduce assumptions from its examples that are absent from the task specification. A more principled approach constrains the type of object that is generated. For example, \citet{muller2026automating}, a preprint at the time of writing, use vision-language preferences to learn a potential function, retraining it periodically from freshly queried VLM preferences and recomputing the shaped rewards already stored in the replay buffer against the updated potential. The model supplies semantic information, and the authors invoke the dynamic-PBRS result of \citet{devlin2012dynamic} to argue that the invariance guarantee survives this periodic revision. This review's own analysis of replay recomputation (Remark~\ref{rem:misreadings}, case ii; Figure~\ref{fig:pairing}) distinguishes two ways such a recomputation can be performed, and the published description does not settle, for this review, which applies here. If the departure term of a stored transition is left at its original value while only the arrival term is refreshed, the pairing is mismatched, the telescoping identity breaks, and no version of Theorem~\ref{thm:dpbrs} applies to that trajectory. If instead both terms of every stored transition are consistently recomputed under the current potential, the construction is the \emph{static} form of Theorem~\ref{thm:pbrs} applied at that instant rather than the \emph{dynamic} form the authors invoke, and it does not by itself establish the dynamic result either, since the potential keeps changing at the next recomputation and the learner is thereby solving a sequence of distinct static problems rather than the one time-indexed process Theorem~\ref{thm:dpbrs} describes. Neither reading confirms the claim as stated without a further argument the published description does not supply. This review therefore reports the source's G1 claim without independently verifying it, marks it accordingly in Table~\ref{tab:classification}, and treats the case as an open verification question rather than a confirmed instance of the pattern. Potential research directions include automated structural checks of generated reward functions, counterexample search, held-out evaluation against the reference criterion, and constrained generation of potential-based or reward-matching forms.

A related application occurs in language-model post-training, where a terminal sequence-level score is redistributed into token-level or step-level feedback \citep{chan2024dense}. If the evaluator, rubric, or generated critique changes during training, the resulting signal is an adaptive reward mechanism. It is dynamic reward shaping proper only when it enters additively alongside a task reward; redistribution, replacement, and direct preference optimisation remain neighbouring mechanisms within the review's scope.

\subsection{Multi-agent dynamic shaping}
\label{sec:fam-marl}

Cooperative multi-agent reinforcement learning combines temporal reward sparsity with an agent-level credit-assignment problem. A shared team return does not indicate which agent's action produced the outcome. Difference rewards estimate marginal contributions by comparing system utility with a counterfactual in which one agent's contribution is removed \citep{agogino2004unifying,tumer2007distributed}. Potential-based shaping can provide denser feedback while preserving equilibrium structure \citep{devlin2011theoretical}. Combining potentials with difference rewards can accelerate learning without changing consistent Nash equilibria \citep{devlin2014potential}. The mechanism becomes dynamic when potentials, counterfactuals, or contribution estimates are revised as the other agents improve.

Contemporary credit-assignment methods provide further connections, but they are not reward shaping, and the distinction is recorded rather than blurred. COMA modifies the policy-gradient estimator: a counterfactual baseline estimates each agent's advantage under the current joint policy, and the reward consumed by the learners is unchanged \citep{foerster2018counterfactual}. QMIX and QPLEX modify value factorisation: a learned team value is decomposed into agent utilities under monotonicity or duplex-dueling constraints, again without touching the reward \citep{rashid2018qmix,wang2021qplex}. All three are class C4 in Table~\ref{tab:classification}, and their guarantee cells are marked n/a$^{\ast}$ because their soundness criteria concern estimator unbiasedness and individual-global consistency rather than reward-shaping invariance. They therefore contribute nothing to the counts of Figure~\ref{fig:landscape}. They are reviewed because they derive policy-dependent, time-varying learning signals from a shared return, and thus solve the same credit-assignment problem that difference rewards solve inside C1.

A useful general form is
\begin{equation}
r^i_k = r^{\mathrm{team}} + F^i_k(s,\mathbf{a},s',\pi_k^{-i},\mathcal{D}_k),
\label{eq:marl-shaping}
\end{equation}
For additive multi-agent shaping, the term $F^i_k$ in Eq.~\ref{eq:marl-shaping} may represent a learned contribution estimate, a difference reward, a role-conditioned potential, or a communication-dependent bonus. Counterfactual advantages and value-factorisation signals are neighbouring C4 mechanisms: they can be time-varying and policy-dependent, but they act in the gradient estimator or value function rather than as an additive reward term. The information source is I6, commonly combined with I3, and the temporal signature is T2 or T3. Carefully constructed potentials can preserve equilibria in potential and stochastic games. In contrast, arbitrary agent-specific shaping may alter strategic incentives, encourage free-riding, or stabilise undesirable conventions.

One possible research direction is to combine explicit reward shaping with contemporary credit-assignment methods rather than treating them as alternatives. A dynamic potential can reduce the temporal reward horizon, while a counterfactual or difference-reward construction separates the contributions of individual agents. Open problems include scalable counterfactual estimation, shaping under changing team composition, robustness to heterogeneous or partially aligned agents, and guarantees when learned communication or role assignments are incorporated into the potential. These questions require evaluation separate from single-agent reward sparsity. An apparent gain in sample efficiency may otherwise result from changing the underlying game rather than accelerating its solution.

\section{Cross-Cutting Analysis}
\label{sec:analysis}

\subsection{Historical evolution of adaptive reward mechanisms}
\label{sec:trajectory}

\begin{figure}[!htbp]
\centering
\resizebox{\textwidth}{!}{%
\begin{tikzpicture}
  \tikzset{
    tnode/.style = {font=\scriptsize, align=center, text width=20mm},
    dot/.style   = {circle, fill=drsblue, draw=none, minimum size=1.6mm, inner sep=0pt},
    dotw/.style  = {circle, fill=drswarn, draw=none, minimum size=1.6mm, inner sep=0pt}
  }
  \draw[drsgrey!50, thick] (0,3.4) -- (17,3.4);
  \draw[drsgrey!50, thick] (0,0)   -- (17,0);
  \draw[drsgrey!50, thick] (0,-3.4) -- (17,-3.4);
  \node[axlabel, anchor=east, text width=22mm, align=right] at (-0.3,3.4)  {Theory};
  \node[axlabel, anchor=east, text width=22mm, align=right] at (-0.3,0)    {Heuristic \& empirical mechanisms};
  \node[axlabel, anchor=east, text width=22mm, align=right] at (-0.3,-3.4) {Learned \& foundation-model signals};

  \node[dot] at (1,3.4) {};     \draw[drsgrey] (1,3.4) -- (1,4.1);
  \node[tnode] at (1,4.5) {\textbf{1999} Ng et al.\ PBRS};

  \node[dot] at (4,3.4) {};     \draw[drsgrey] (4,3.4) -- (4,5.0);
  \node[tnode] at (4,5.4) {\textbf{2003} Wiewiora equivalence};

  \node[dot] at (7.4,3.4) {};   \draw[drsgrey] (7.4,3.4) -- (7.4,4.1);
  \node[tnode] at (7.4,4.5) {\textbf{2012} Devlin \& Kudenko DPBRS};

  \node[dot] at (10,3.4) {};    \draw[drsgrey] (10,3.4) -- (10,5.0);
  \node[tnode] at (10,5.4) {\textbf{2017} Grze\'s terminal condition};

  \node[dot] at (13,3.4) {};    \draw[drsgrey] (13,3.4) -- (13,4.1);
  \node[tnode] at (13,4.5) {\textbf{2024} PBIM; GRM$^{\ddagger}$};

  \node[dot] at (15.8,3.4) {};  \draw[drsgrey] (15.8,3.4) -- (15.8,5.0);
  \node[tnode] at (15.8,5.4) {\textbf{2025} ADOPS / BAMDP shaping};

  \node[dot] at (0.4,0) {};   \draw[drsgrey] (0.4,0) -- (0.4,-0.7);
  \node[tnode] at (0.4,-1.1) {\textbf{1998} Randl\o v \& Alstr\o m bicycle failure};

  \node[dot] at (3.0,0) {};   \draw[drsgrey] (3.0,0) -- (3.0,0.7);
  \node[tnode] at (3.0,1.1) {\textbf{2009} TAMER};

  \node[dot] at (5.6,0) {};   \draw[drsgrey] (5.6,0) -- (5.6,-0.7);
  \node[tnode] at (5.6,-1.1) {\textbf{2010} Value-derived / bi-level reward design};

  \node[dot] at (8.4,0) {};   \draw[drsgrey] (8.4,0) -- (8.4,1.6);
  \node[tnode] at (8.4,2.0) {\textbf{2015--22} DPBA proposed as potential-based advice, later shown to alter the optimum; replaced by PIES};

  \node[dot] at (11.2,0) {};  \draw[drsgrey] (11.2,0) -- (11.2,-0.7);
  \node[tnode] at (11.2,-1.1) {\textbf{2016--19} Count-based / ICM / RND bonuses};

  \node[dot] at (13.8,0) {};  \draw[drsgrey] (13.8,0) -- (13.8,0.7);
  \node[tnode] at (13.8,1.1) {\textbf{2023} Behaviour alignment; ROSA};

  \node[dot] at (16.4,0) {};  \draw[drsgrey] (16.4,0) -- (16.4,-0.7);
  \node[tnode] at (16.4,-1.1) {\textbf{2025} Bootstrapped / self-adaptive shaping};

  \node[dotw] at (2.4,-3.4) {};   \draw[drsgrey] (2.4,-3.4) -- (2.4,-4.1);
  \node[tnode] at (2.4,-4.5) {\textbf{2017} Deep RL from human preferences};

  \node[dotw] at (7.0,-3.4) {};   \draw[drsgrey] (7.0,-3.4) -- (7.0,-5.0);
  \node[tnode] at (7.0,-5.4) {\textbf{2022} InstructGPT RLHF; Constitutional AI};

  \node[dotw] at (9.6,-3.4) {};   \draw[drsgrey] (9.6,-3.4) -- (9.6,-4.1);
  \node[tnode] at (9.6,-4.5) {\textbf{2023} DPO};

  \node[dotw] at (12.2,-3.4) {};  \draw[drsgrey] (12.2,-3.4) -- (12.2,-5.0);
  \node[tnode] at (12.2,-5.4) {\textbf{2024} Eureka; Text2Reward; assistant reward agent};

  \node[dotw] at (14.6,-3.4) {};  \draw[drsgrey] (14.6,-3.4) -- (14.6,-4.1);
  \node[tnode] at (14.6,-4.5) {\textbf{2025} Adaptive LTL-progress rewards};

  \node[dotw] at (16.8,-3.4) {};  \draw[drsgrey] (16.8,-3.4) -- (16.8,-5.0);
  \node[tnode] at (16.8,-5.4) {\textbf{2026} VLM-guided potentials};

  \draw[-{Stealth[length=2.5mm]}, draw=drsblue, very thick] (0,-6.6) -- (17,-6.6);
  \node[font=\footnotesize\bfseries, drsblue] at (8.5,-6.35) {Illustrative shift in information sources among reviewed methods};
  \node[font=\scriptsize, drsgrey] at (2.5,-7.0) {designer-specified (I1)};
  \node[font=\scriptsize, drsgrey] at (8.5,-7.0) {agent-internal, increasingly corrected (I3)};
  \node[font=\scriptsize, drsgrey] at (14.5,-7.0) {learned / generative, externally supplied (I7--I9)};
\end{tikzpicture}}
\caption{Illustrative timeline of dynamic reward shaping and neighbouring adaptive mechanisms, organised into three threads rather than a single chronology: theoretical guarantees (top), heuristic and self-referential mechanisms (middle), and learned or foundation-model-supplied signals (bottom). Horizontal position reflects relative chronology within each thread and is not drawn to scale. The bottom arrow summarises the trend discussed below: successive families draw their information increasingly from the agent itself and, more recently, from external learned or generative models, rather than from designer specification alone.}
\label{fig:timeline}
\end{figure}
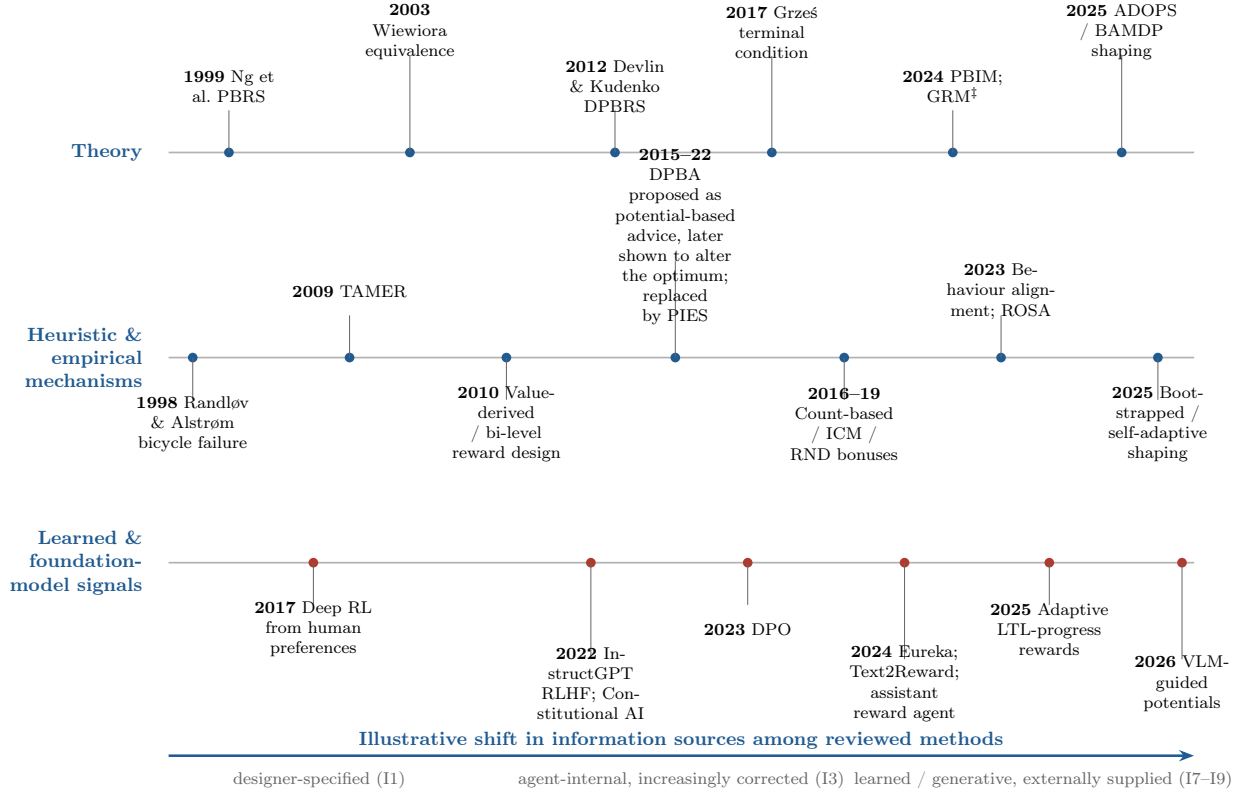

The family-by-family review of Section~\ref{sec:families} is organised by mechanism, which is useful for locating a method but leaves a question open: why did the literature move through these families in this order, which of them are still active research directions, and which have effectively been abandoned? Figure~\ref{fig:evolution} provides a conceptual map of prominent lines of development from hand-crafted engineering to autonomous, self-referential shaping. Figure~\ref{fig:timeline} then unpacks the same period into three parallel threads with the specific results that drove each transition, and the remainder of this subsection draws four cross-cutting conclusions from it that do not follow from reading the families in isolation.

\begin{figure}[!htbp]
\centering
\resizebox{\textwidth}{!}{%
\begin{tikzpicture}[node distance=6mm]
  \tikzset{
    era/.style   = {draw=drsblue, rounded corners=3pt, align=center,
                     minimum width=30mm, minimum height=14mm, font=\normalsize,
                     inner sep=4pt},
    sub/.style   = {font=\footnotesize, drsgrey, align=center, text width=38mm,
                     anchor=north}
  }
  \node[era, fill=drsblue!6]  (e0) at (0,0)     {Hand-crafted\\reward engineering};
  \node[era, fill=drsblue!13] (e1) at (4.3,0)   {Potential-based\\shaping};
  \node[era, fill=drsblue!20] (e2) at (8.6,0)   {Learned /\\dynamic shaping};
  \node[era, fill=drsblue!27] (e3) at (12.9,0)  {Meta-learned /\\bi-level shaping};
  \node[era, fill=drsblue!34] (e4) at (17.2,0)  {Human-in-the-\\loop shaping};
  \node[era, fill=drsblue!41] (e5) at (21.5,0)  {Preference learning\\/ RLHF};
  \node[era, fill=drsblue!48] (e6) at (25.8,0)  {Foundation-model\\shaping};
  \node[era, fill=drsblue!55] (e7) at (30.1,0)  {Autonomous,\\self-referential shaping};

  \node[sub] at (0,-1.0)    {before 1999\\\citep{skinner1938behavior,dorigo1998robot}};
  \node[sub] at (4.3,-1.0)  {1999\\\citep{ng1999policy}};
  \node[sub] at (8.6,-1.0)  {2010--25\\\citep{devlin2012dynamic,adamczyk2025bootstrapped}};
  \node[sub] at (12.9,-1.0) {2010--20\\\citep{sorg2010reward,hu2020learning}};
  \node[sub] at (17.2,-1.0) {2009--17\\\citep{knox2009interactively,macglashan2017interactive}};
  \node[sub] at (21.5,-1.0) {2017--22\\\citep{christiano2017deep,ouyang2022training}};
  \node[sub] at (25.8,-1.0) {2024--26\\\citep{ma2024eureka,muller2026automating}};
  \node[sub] at (30.1,-1.0) {2024--25\\\citep{ma2024assistant,forbes2025action}};

  \draw[-{Stealth[length=2.5mm]}, draw=drsblue, thick] (e0.east) -- (e1.west);
  \draw[-{Stealth[length=2.5mm]}, draw=drsblue, thick] (e1.east) -- (e2.west);
  \draw[-{Stealth[length=2.5mm]}, draw=drsblue, thick] (e2.east) -- (e3.west);
  \draw[-{Stealth[length=2.5mm]}, draw=drsblue, thick] (e3.east) -- (e4.west);
  \draw[-{Stealth[length=2.5mm]}, draw=drsblue, thick] (e4.east) -- (e5.west);
  \draw[-{Stealth[length=2.5mm]}, draw=drsblue, thick] (e5.east) -- (e6.west);
  \draw[-{Stealth[length=2.5mm]}, draw=drsblue, thick] (e6.east) -- (e7.west);

  \node[note, font=\footnotesize, align=center] at (15.05,-3.1) {Fill intensity tracks Figure~\ref{fig:timeline}'s information-source trend: designer-specified (left) to agent-internal to externally learned or generated (right).};
\end{tikzpicture}}
\caption{A conceptual map of prominent lines of development in dynamic reward shaping: from hand-crafted engineering, through the provably safe but static potential-based construction, to signals that are learned, meta-optimised, elicited from people, and finally generated or corrected by the agent and by external models. This is a simplified companion to Figure~\ref{fig:timeline}, which gives the three-thread account with the results that mark each transition; neither figure claims that earlier stages have stopped being used, only that each stage is where selected examples of newer development appear.}
\label{fig:evolution}
\end{figure}

\paragraph{Drivers of the historical evolution} Read against Dimension 2 of the taxonomy (Section~\ref{sec:dim-source}), Figure~\ref{fig:timeline} illustrates an apparent shift within the reviewed examples: from designer-specified potentials (I1), toward signals read off the agent's own internal state (I3) and, more recently, signals supplied by external learned or generative models (I7--I9). Several later methods can be interpreted as addressing limitations of earlier approaches. Hand-designed potentials can require substantial task knowledge and may transfer poorly when the structure of a new task differs from the one for which they were designed; a prominent response in the reviewed literature was to derive the potential from a quantity the agent already estimates, principally a value function, converting shaping from a specification problem into a bootstrapping problem (Section~\ref{sec:fam-value}). This, in turn, exposed a self-referential feedback loop between the shaping signal and the function that generates it, which the optimality-preserving corrections of Section~\ref{sec:beyond-pbrs} and the Bayesian treatments of Section~\ref{sec:fam-bayes} address by constraining the loop's structure rather than removing it. For tasks where manual specifications or agent-internal estimates are insufficient, recent work has also drawn on human preferences, learned reward models, and foundation models. The representative methods reviewed here draw shaping information from a wider range of sources than designer specification alone, including agent estimates, human feedback, reward models, world models, and foundation models.

\paragraph{Possible shifts within the reviewed literature.} The review's search protocol is theory-led and citation-traced rather than systematic (Section~\ref{sec:review-approach}), so the following is offered as a plausible reading of the reviewed set rather than a bibliometric finding; a claim of field-wide decline would need the kind of publication-count evidence this review does not collect. With that qualification, three patterns are visible in Figure~\ref{fig:timeline} and each suggests a distinct cause. Manually scheduled shaping (T1, Section~\ref{sec:fam-schedule}) appears, in the reviewed set, to be giving way to performance-driven alternatives, plausibly because a fixed schedule tuned on one task does not transfer to the next, and alternative mechanisms, such as outer optimisation loops or co-trained reward agents, are available in settings where a fixed schedule may be insufficient \citep{wang2025reward,ma2024assistant}. Uncorrected intrinsic-motivation bonuses (G5 in Figure~\ref{fig:guarantee-ladder}) show a similar pattern relative to the optimality-preserving corrections of Section~\ref{sec:beyond-pbrs}: as intrinsic motivation moved from small tabular demonstrations to long-horizon, higher-stakes exploration problems, the cost of altering the optimal policy may have motivated interest in optimality-preserving corrections for intrinsic rewards, though this review cannot rule out that both forms simply continue to coexist across different applications. Dynamic potential-based advice (DPBA) is a firmer case, since here the record is a documented correction rather than an inferred trend: \citet{harutyunyan2015expressing} proposed learning an auxiliary value function as a potential, \citet{behboudian2020useful,behboudian2022policy} subsequently showed both theoretically and empirically that the construction can alter the optimal policy, and PIES replaced it with a mechanism that abandons the reward-level claim entirely in favour of decaying policy-level advice. The interpretation of TAMER-style interactive shaping (Section~\ref{sec:fam-human}) is the most speculative of these observations: the reviewed set is consistent with the possibility that modelling a live human evaluator does not scale to the training budgets of large models and is amortised instead into a learned reward model trained once and queried repeatedly, which is the RLHF pipeline discussed next, but the review has not verified that TAMER-style methods stopped being used rather than simply being applied in settings this review's search did not surface.

\paragraph{Why RLHF differs from PBRS at the objective level.} Both potential-based reward shaping and RLHF are routinely described as reward shaping, and Table~\ref{tab:classification} classifies both within the C1--C4 mechanism classes, but the resemblance is mechanistic; the two paradigms address different objective-specification settings. PBRS presupposes that the task reward $R$ is already known and already the correct objective; the entire construction exists to make a known target easier to reach, and its guarantee is a closed-form proof that the target is unchanged (Theorem~\ref{thm:pbrs}). RLHF commonly addresses settings in which the intended objective is difficult to specify as a reliable manually designed scalar reward, so the learned reward model $\hat R_k$ serves as a proxy derived from preference data, accessible only indirectly through pairwise human comparisons. This is why RLHF sits in C2 rather than C1 in Figure~\ref{fig:hierarchy}: there is no separately available $R$ for the auxiliary signal to leave unchanged, so the question Theorem~\ref{thm:pbrs} answers cannot even be posed in RLHF's setting. The consequence is that ``safety'' means two different things in the two paradigms. For PBRS, safety is an object-level, checkable property: a telescoping identity either holds or it does not (Section~\ref{sec:static}). For RLHF and other C2 mechanisms, safety is a statistical and epistemic property: whether the proxy $\hat R_k$ remains calibrated to an unobserved target as the policy distribution drifts away from the data the proxy was trained on, which is exactly the reward-model over-optimisation failure of Section~\ref{sec:safety} and why every C2 entry in Table~\ref{tab:classification} carries a starred guarantee rather than a G1. Direct preference optimisation does not close this gap by removing the explicit reward model; it reparameterises the same preference-comparison epistemics, and remains C4 in the classification for the same reason. Treating PBRS and RLHF as points on a single spectrum therefore risks importing PBRS's checkable notion of safety into a setting where the underlying objective is, by construction, never directly observed.

\paragraph{Which directions are converging.} Three convergences cut across the families of Section~\ref{sec:families}. First, there is the pattern already identified in Section~\ref{sec:taxonomy-observations}: value-derived potentials, selected optimality-preserving intrinsic-motivation constructions, Bayes-adaptive shaping, and, if its reported claim is upheld, VLM-guided potential learning \citep{muller2026automating} share a related design principle: constrain the learned guidance signal to a form for which a preservation result can be stated, each subject to its own timing, boundary, and state-representation conditions (Table~\ref{tab:theorems}), while allowing increasingly rich information sources to determine its content. Figure~\ref{fig:timeline} shows this convergence as a trend rather than an isolated observation: the theory lane's most recent peer-reviewed entries and the foundation-model lane's most recent entry, still a preprint at the time of writing, apply the same design pattern to different information sources. Second, structure-driven and foundation-model-driven shaping are converging toward specification-grounded generation: adaptive LTL-progress rewards revise a reward function defined over formally specified task structure using measured performance, while VLM-guided potentials report using a foundation model's preferences to populate a potential whose form is fixed by construction; if that report holds up under the verification question raised in Section~\ref{sec:fam-foundation} and marked $^{\ddagger}$ in Table~\ref{tab:classification}, both pair an external, richly informative source with a structural constraint intended to make the result auditable, rather than treating formal specification and generative modelling as separate design philosophies. Third, bi-level and meta-optimised shaping connects to dual-agent construction: whereas \citet{sorg2010reward} and \citet{zheng2018learning} optimise an explicit upper-level objective, ROSA and the assistant reward agent of \citet{ma2024assistant} implement the outer shaping mechanism as a second learning agent. None of these convergences yet has a unifying theorem; identifying one is listed among the open directions in Section~\ref{sec:agenda}.

\subsection{Comparing the families}
\label{sec:comparison}

Table~\ref{tab:families} compares the method families along the principal dimensions that determine their practical selection. Representative references are included for each family; comprehensive citations are provided in Section~\ref{sec:families}.

\begin{table}[!htbp]
\centering
\small
\caption{Comparison of method families. ``Cls'' lists the mechanism classes (Section~\ref{sec:unified-framework}) occurring among the reviewed members of the family; the class varies by implementation, and the per-entry assignments in Table~\ref{tab:classification} remain authoritative; ``prior knowledge'' denotes the information supplied by the designer; ``overhead'' denotes the training cost relative to the unshaped baseline. A guarantee marked $^{\ast}$ belongs to a family that is not predominantly additive shaping, and records that family's own criterion rather than reward-shaping policy invariance. References are representative rather than exhaustive. Overhead categories are qualitative comparative assessments in this review and are not standardised measurements reported across the cited studies.}
\label{tab:families}
\begin{tabular}{@{}p{0.15\linewidth}p{0.13\linewidth}p{0.17\linewidth}p{0.10\linewidth}p{0.12\linewidth}p{0.20\linewidth}@{}}
\toprule
\textbf{Family} & \textbf{Cls} & \textbf{Prior knowledge} & \textbf{Overhead} & \textbf{Typical guarantee} & \textbf{Principal risk} \\
\midrule
Schedule-based \citep{wang2025reward} & C1/C4 & Shaping terms and a schedule & Negligible & G4 & Schedule is another hyperparameter set; over-shaping early \\
Value-derived \citep{adamczyk2025bootstrapped} & C1 & None & Low & G1 & Self-referential instability under approximation \\
Uncertainty-aware \citep{marom2018belief} & C1 & A prior & Low--mod. & G2 or G5 & The classification depends on whether exact limiting recovery is proved \\
Advice-derived \citep{behboudian2022policy} & C1/C4 & Plan, subgoals, or advice & Low & G1/G2 & Conversion step is where guarantees fail \\
Interactive human \citep{macglashan2017interactive} & C2/C4 & Live trainer time & Human cost & G3 or G5$^{\ast}$ & COACH has a local convergence result; the other included methods are empirical \\
Intrinsic motivation \citep{forbes2024pbim,forbes2024generalized} & C1 & None & Low--mod. & G1 (corrected) & Uncorrected forms change the optimum \\
Bi-level \citep{hu2020learning} & C1 & Optional heuristics & High & G4 & Objective-aware selection is not policy invariance \\
Structure-driven \citep{icarte2022reward} & C1/S & Formal specification & Low & G1 (static, augmented) & Specification may not exist \\
Preference/RLHF \citep{ouyang2022training} & C2 & Rankings or critiques & High & G5$^{\ast}$ & Reward-model shift and over-optimisation \\
World-model-derived bonuses \citep{sekar2020planning,fu2023gobi} & C1 when added to $R$; otherwise out of scope & Interaction data & High & G5 & Model error can become fictitious progress \\
Foundation-model \citep{ma2024eureka} & C1/C2 & Task description & Very high & G4$^{\ast}$ & Unverified generated reward code \\
Multi-agent \citep{devlin2014potential} & C1/C4 & Domain knowledge (optional) & Low & G1 (equilibria) & Co-adaptation; attribution remains hard \\
\bottomrule
\end{tabular}
\end{table}
\FloatBarrier

Table~\ref{tab:families} suggests two practical contrasts for method selection. Value-derived and intrinsic-motivation methods require no prior knowledge and introduce relatively little computational overhead. They are reasonable starting points when task-specific knowledge is unavailable, provided that their optimisation behaviour and any required preservation conditions are evaluated. Their principal risks concern optimisation stability rather than the semantic validity of supplied knowledge. In contrast, bi-level and foundation-model methods can incorporate richer information, but they introduce additional optimisation, evaluation, or model-inference costs. Structure-driven shaping lies between these cases. When a valid formal specification is available, structure-driven potential constructions can offer interpretable guidance and structural preservation results, subject to the assumptions of the specific construction. Its main limitation is the availability and construction cost of the specification.

\subsection{A practical decision guide}
\label{sec:decision-guide}

\begin{figure}[!htbp]
\centering
\resizebox{\textwidth}{!}{%
\begin{tikzpicture}
  \tikzset{
    stage/.style = {draw=black, thick, fill=white, rounded corners=2pt, align=left,
                     text width=54mm, inner sep=7pt, font=\small, anchor=north}
  }
  \node[stage] (s1) at (0,3) {
    \textbf{\Large 1}\quad\textbf{Characterise the task}\\[4pt]
    \textbullet\ Multi-agent credit assignment?\\
    \textbullet\ Trusted formal specification (LTL, automaton, plan)?\\
    \textbullet\ Is objective preservation safety-critical?\\
    \textbullet\ Is the objective itself hard to write down, with only comparisons available?\\
    \textbullet\ Live human feedback available and affordable?\\
    \textbullet\ Compute budget for an outer optimisation loop or a foundation model?
  };
  \node[stage] (s2) at (11.5,3) {
    \textbf{\Large 2}\quad\textbf{Select family and guarantee class}\\[4pt]
    \textbullet\ Map answers to a starting family via Table~\ref{tab:decision}\\
    \textbullet\ Read the guarantee class that route can reach off Figure~\ref{fig:guarantee-ladder}\\
    \textbullet\ Confirm the mechanism class (C1--C4) via Figure~\ref{fig:hierarchy}\\
    \textbullet\ If G1 is required but unavailable, consider the ``learn the potential, not the reward'' pattern of \S\ref{sec:taxonomy-observations}
  };
  \node[stage] (s3) at (23,3) {
    \textbf{\Large 3}\quad\textbf{Implement with safeguards}\\[4pt]
    \textbullet\ Implement the construction that earns the selected G1 guarantee: Eq.~\ref{eq:pbrs} for static PBRS, Eq.~\ref{eq:dpbrs} for time-indexed PBRS, or the method-specific conditions of Table~\ref{tab:theorems}\\
    \textbullet\ Choose a replay strategy from Figure~\ref{fig:pairing}\\
    \textbullet\ Zero terminal potentials; separate task and shaping normalisation; treat the potential offset as tunable (\S\ref{sec:implementation})
  };
  \node[stage] (s4) at (34.5,3) {
    \textbf{\Large 4}\quad\textbf{Evaluate}\\[4pt]
    \textbullet\ Unshaped, static-potential, and frozen-adaptation baselines (Table~\ref{tab:protocol})\\
    \textbullet\ Report against the reference criterion throughout training\\
    \textbullet\ Test a deliberately misspecified signal
  };

  \draw[-{Stealth[length=1.5mm]}, draw=black, thin] (s1.east) -- (s2.west);
  \draw[-{Stealth[length=1.5mm]}, draw=black, thin] (s2.east) -- (s3.west);
  \draw[-{Stealth[length=1.5mm]}, draw=black, thin] (s3.east) -- (s4.west);
\end{tikzpicture}}
\caption{A practical pipeline from task to deployment. Stage 1 characterises the task along the axes used in Table~\ref{tab:decision}; Stage 2 turns that characterisation into a family and an expected guarantee class using the figures introduced earlier in the review; Stage 3 applies the implementation safeguards of Section~\ref{sec:implementation}; Stage 4 applies the evaluation protocol of Section~\ref{sec:evaluation}. The pipeline does not remove judgement from the process, but it fixes the order in which questions should be asked.}
\label{fig:pipeline}
\end{figure}
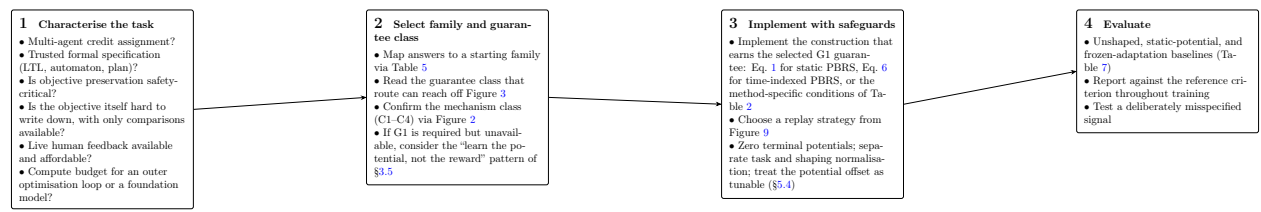

The comparisons above answer ``what does each family cost and guarantee?''. A researcher approaching a new task instead needs the inverse mapping: given the properties of \emph{this} task, which family should be tried first? Figure~\ref{fig:pipeline} lays out the resulting workflow as a pipeline, and Table~\ref{tab:decision} makes the first stage concrete by mapping task characteristics to a starting recommendation. Neither replaces the fuller analysis of Sections~\ref{sec:families} and \ref{sec:analysis}; both are meant to shorten the distance from ``I have this task'' to ``here is where the relevant analysis lives''.

\begingroup
\small
\setlength{\LTleft}{0pt}
\setlength{\LTright}{0pt}

\begin{longtable}{@{}p{0.24\linewidth}p{0.24\linewidth}p{0.12\linewidth}p{0.30\linewidth}@{}}

\caption{A starting point for family selection, keyed to task characteristics rather than to method names. ``Guarantee reachable'' states the strongest class realistically available on this route, read from Figure~\ref{fig:guarantee-ladder}; a starred class is a C2--C4 method's own criterion, not reward-shaping invariance. A G1 entry is a structural guarantee about the potential-based construction, not an implementation guarantee about a specific deep-RL system built on it; see the note preceding Table~\ref{tab:classification} in Section~\ref{sec:classification}. Characteristics are not mutually exclusive; where more than one applies, the earlier row in the table takes priority, since multi-agent structure and a trustworthy formal specification are the most constraining properties a task can have.}
\label{tab:decision}
\\

\toprule
\textbf{Task characteristic} &
\textbf{Starting family (\S)} &
\textbf{Guarantee reachable} &
\textbf{Why} \\
\midrule
\endfirsthead

\multicolumn{4}{@{}l}{\tablename~\thetable\ continued from the previous page} \\
\toprule
\textbf{Task characteristic} &
\textbf{Starting family (\S)} &
\textbf{Guarantee reachable} &
\textbf{Why} \\
\midrule
\endhead

\midrule
\multicolumn{4}{r@{}}{\textit{Continued on the next page}} \\
\endfoot

\bottomrule
\endlastfoot

Multi-agent credit assignment &
Multi-agent, potential-based difference rewards
(\S\ref{sec:fam-marl}) &
G1 (equilibria) &
Preserves consistent Nash equilibria while densifying individual credit
\citep{devlin2011theoretical,devlin2014potential} \\
\addlinespace

Trusted formal specification exists (LTL, automaton, plan) &
Structure-driven shaping (\S\ref{sec:fam-structure}) &
G1 (static, augmented) &
Interpretable and strongly guaranteed, but only as good as the specification; revise the specification itself, not just its state, to move beyond S in Table~\ref{tab:classification} \\
\addlinespace

Objective preservation is safety-critical; no formal specification &
Value-derived potentials or corrected intrinsic motivation
(\S\ref{sec:fam-value}, \S\ref{sec:fam-im}); a manually designed static potential
(\S\ref{sec:static}) remains available whenever enough task knowledge exists to hand-specify one &
G1 (structural; see note below) &
A practical route that does not require a manually specified potential and can retain a structural guarantee, provided the required construction and implementation conditions are met; not the only route, since a fixed hand-designed potential earns the same structural guarantee from Theorem~\ref{thm:pbrs} directly, at the cost of requiring that knowledge up front \\
\addlinespace

Objective hard to specify; only comparisons or rankings available &
Reward replacement / preference learning and RLHF
(\S\ref{sec:fam-preference}) &
G4$^{\ast}$ or G5$^{\ast}$ &
A PBRS-style invariance claim relative to an explicit task reward is unavailable when the reference criterion is itself unobserved (Section~\ref{sec:trajectory}); mitigate with held-out evaluation and ensembling \\
\addlinespace

Live human feedback available and affordable &
Interactive human shaping, framed as advice or advantage rather than reward
(\S\ref{sec:fam-human}) &
G3$^{\ast}$ or G5$^{\ast}$ &
Feedback is policy-dependent \citep{macglashan2017interactive}; prefer policy-level advice with decaying influence over direct reward modelling of a live trainer where both are available \\
\addlinespace

High compute budget; reward design itself is the bottleneck &
Bi-level / foundation-model-driven shaping
(\S\ref{sec:fam-bilevel}, \S\ref{sec:fam-foundation}) &
G4$^{\ast}$ or G4 &
Can use highly expressive information sources, but the reviewed methods do not generally provide a structural preservation guarantee; verification before deployment is an open problem (Section~\ref{sec:agenda}) \\
\addlinespace

None of the above is pressing (default, single-agent, sparse reward) &
Value-derived potentials or corrected intrinsic motivation
(\S\ref{sec:fam-value}, \S\ref{sec:fam-im}) &
G1 &
Requires no prior task knowledge, which a fixed hand-designed potential cannot say; among the reviewed approaches that do not require a manually specified potential, these methods typically have lower overhead than bi-level or foundation-model-driven alternatives. \\

\end{longtable}
\endgroup

This mapping is a starting point, not a substitute for the comparisons in Table~\ref{tab:families} and the safety analysis of Section~\ref{sec:safety}. Two qualifications matter in particular. First, several rows can apply simultaneously; a safety-critical multi-agent robotics task should follow the multi-agent row for credit assignment and the safety-critical row for the guarantee expected of whatever potential is used within it, rather than treating the rows as mutually exclusive labels. Second, the table records what is \emph{reachable}, not what is automatically obtained: reaching G1 on the safety-critical row still requires the implementation safeguards of Stage 3, and omitting them can cause a proved guarantee to no longer describe the deployed system (Section~\ref{sec:implementation}).

\subsection{Implementation compatibility}
\label{sec:implementation}

This review highlights implementation mismatches that can arise when a time-varying reward is combined with mechanisms not modelled in the corresponding preservation results. These failures need not reflect an incorrect shaping principle. They can instead result from a mismatch between the formal construction and the learning system in which it is deployed. Each failure below arises where a deployed system departs from the shaping term of Definition~\ref{def:dpbrs}, or from the conditions under which the guarantee attaching to it was proved; they are the concrete counterparts of the misreadings set out in Remark~\ref{rem:misreadings}. The mechanisms below should therefore be treated as an implementation audit for dynamic shaping rather than as properties guaranteed by the cited invariance results.

\begin{figure}[!htbp]
\centering
\resizebox{\textwidth}{!}{%
\begin{tikzpicture}
  \tikzset{
    hdr/.style  = {font=\footnotesize\bfseries, text=black, align=center,
                   text width=32mm},
    rhdr/.style = {font=\footnotesize\bfseries, text=black, align=right,
                   text width=46mm, anchor=east},
    m/.style    = {draw=drsgrey!50, minimum width=30mm, minimum height=12mm,
                   text width=27mm, font=\footnotesize, align=center,
                   inner sep=2pt}
  }
  \node[hdr] at (3.4,1.35)  {(A) store $\Phi_k(s),\Phi_{k+1}(s')$ with the transition};
  \node[hdr] at (7.0,1.35)  {(B) recompute from $\Phi_m$ at sampling time};
  \node[hdr] at (10.6,1.35) {(C) update $\Phi$ at epoch boundaries, flush};
  \node[hdr] at (14.2,1.35) {(D) rate-limit $\|\Phi_{k+1}-\Phi_k\|$};

  \draw[drsgrey!40] (1.75,0.72) -- (15.85,0.72);
  \draw[drsgrey!40] (1.65,0.72) -- (1.65,-3.85);

  \node[rhdr] at (1.4,0)    {P1 index pairing of Eq.~\ref{eq:dpbrs} holds};
  \node[rhdr] at (1.4,-1.6) {P2 the transition carries the reward it was given};
  \node[rhdr] at (1.4,-3.2) {P3 the critic's regression target is stationary};

  \node[m, fill=drsfill]    at (3.4,0)  {holds};
  \node[m, fill=drsfill!40] at (7.0,0)  {static at $m$, not dynamic pairing};
  \node[m, fill=drsfill]    at (10.6,0) {holds within epoch};
  \node[m, fill=drsfill!60] at (14.2,0) {approximate; pairing error bounded};

  \node[m, fill=drsfill]    at (3.4,-1.6)  {exact};
  \node[m, fill=drswfill]   at (7.0,-1.6)  {rewritten};
  \node[m, fill=drsfill]    at (10.6,-1.6) {exact};
  \node[m, fill=drsfill]    at (14.2,-1.6) {exact};

  \node[m, fill=drswfill]   at (3.4,-3.2)  {no: mixed epochs};
  \node[m, fill=drsfill!60] at (7.0,-3.2)  {within an epoch only};
  \node[m, fill=drsfill!60] at (10.6,-3.2) {conditional: within a frozen epoch, at a cost in sample reuse};
  \node[m, fill=drswfill]   at (14.2,-3.2) {no; drift bounded};
\end{tikzpicture}}
\caption{Replay, the index pairing of Eq.~\ref{eq:dpbrs}, and three properties that are frequently
conflated. P1 is a property of the shaped decision problem, P2 of the stored data, and P3 of the
optimisation. They are logically independent. Option (A) secures P1 and P2 but not P3: reconstructing
the shaping term exactly preserves the identity along the chronological trajectory, whereas a replay
batch is a set of transitions drawn from many potential epochs and does not form that trajectory, so
the critic still faces a moving target. Option (B) recomputes both the departure and arrival term of
a stored transition under the current $\Phi_m$; this is an instance of the \emph{static} construction
of Theorem~\ref{thm:pbrs} evaluated at $m$ rather than a violation of it, so P1 as defined for the dynamic pairing does not apply, but the original historical reward is lost (P2 fails) and $\Phi_m$ itself keeps changing, so the learner faces a sequence of distinct static problems rather than one stationary target (P3 fails). This differs from mismatching a stale departure term against a freshly recomputed arrival term, which does break the telescoping identity outright and is not a legitimate storage option at all (Remark~\ref{rem:misreadings}). Options (C) and (D) trade sample reuse, or an explicit approximation error, for a bounded version of all three. Rate limiting does not itself establish P1: if the paired potentials are stored exactly, P1 holds regardless of the rate, and if they are not, rate limiting only bounds the mismatch. No option in this set secures all three without cost, and the table is offered as a statement of the trade-off rather than as a recommendation. The reviewed literature did not identify a controlled study that isolates which trade-off dominates in practice.
On-policy learners do not face this replay-storage choice, although other potential-update choices still apply.}
\label{fig:pairing}
\end{figure}

\paragraph{Replay, and three properties that are not the same.} Off-policy algorithms store transitions together with their rewards, and a time-varying potential makes the stored reward an artefact of the moment of collection. Discussion of this problem tends to conflate three requirements that are in fact independent, and that no single storage policy secures simultaneously (Figure~\ref{fig:pairing}).

\emph{P1, invariance.} The index pairing of Eq.~\ref{eq:dpbrs} must hold for Theorem~\ref{thm:dpbrs}, the \emph{dynamic} result, to apply. Two ways of recomputing a stored transition's shaping term must be told apart. Recomputing only the arrival term from a later potential while leaving a stale departure term breaks the pairing outright and is the anticipative case (v) of Remark~\ref{rem:augmentation}; no version of the dynamic argument applies to that trajectory. Recomputing both terms consistently under one current potential does not break anything, but it also does not satisfy the dynamic pairing; it is instead an instance of the \emph{static} construction of Theorem~\ref{thm:pbrs} evaluated at that instant, which is a different, and separately non-stationary, construction discussed with Option (B) in Figure~\ref{fig:pairing}.

\emph{P2, faithful reconstruction.} The transition should carry the shaping reward it was actually given. Storing $\Phi_k(s)$ and $\Phi_{k+1}(s')$ alongside the transition achieves this exactly.

\emph{P3, target stationarity.} The critic should not be regressing toward a target that drifts for reasons unrelated to its own error. This is where the common recommendation is too strong. Storing the potentials preserves the shaping identity \emph{along the trajectory on which it was generated}; a replay batch is not that trajectory, but a sample drawn across many potential epochs. Faithful reconstruction therefore leaves the non-stationarity of the regression target untouched. P1 and P2 are properties of the decision problem and of the data respectively; P3 is a property of the optimisation, and it is not implied by either.

Practical responses trade among the three rather than resolving them: restricting potential updates to epoch boundaries and flushing or reweighting the buffer is proposed here as a mitigation that approximately restores all three \emph{within} an epoch, at a cost in sample reuse; bounding the rate of change of the potential similarly yields a quantified approximation to all three rather than a recovery. Both mitigations rest on further assumptions that are easy to leave implicit: that the potential is genuinely frozen across every critic update within the epoch and not only between buffer flushes, that target networks are refreshed on a schedule consistent with the epoch boundary rather than independently of it, and that multi-step or $\lambda$-returns are not silently mixing potentials from adjacent epochs (Remark below). None of this is established by the figure's classification alone; it would need to be verified for a specific algorithm rather than assumed from the storage policy. The reviewed literature did not identify a controlled study that isolates these replay-storage trade-offs; resolving them requires dedicated empirical comparison. On-policy algorithms avoid the replay-buffer mismatch described above because they do not sample transitions collected under multiple potential epochs from a replay buffer. This distinction should be stated when dynamic-shaping results are compared across algorithm classes. It is narrower than a claim that on-policy learners are unaffected by dynamic shaping: they remain exposed to a non-stationary reward target, to shifting advantage or return estimates, and to critic drift driven by a fast-changing potential, none of which is a replay artefact.

\paragraph{Bootstrapped critics.} When a potential is derived from a value function, the reward used to train that function depends on the function itself. This feedback loop has been analysed in tabular settings. In deep RL, however, it is combined with function approximation, bootstrapping, and off-policy sampling. The resulting dynamics are not covered by the theoretical results commonly cited for these methods.

\paragraph{Multi-step returns and traces.} Both $n$-step and $\lambda$-returns aggregate shaping terms across multiple transitions. To represent the chronological dynamic construction of Eq.~\ref{eq:dpbrs}, each transition in the return must use its own consistently paired departure and arrival potentials. Recomputing an entire window under one current potential instead yields a different static-PBRS construction, whereas mixing incompatible potential versions within a transition or return window can break the intended telescoping identity.


\paragraph{Terminal states.} The condition $\Phi(s_{\mathrm{terminal}}) = 0$ must be imposed explicitly when the potential is learned \citep{grzes2017reward}. Assigning a non-zero potential to a terminal state can change the limiting objective and reintroduce opportunities for exploiting the shaped reward. This is precisely the type of behaviour that potential-based shaping is intended to prevent.

\paragraph{Normalisation.} Online normalisation of the combined reward can alter the effective relative scale of the task reward and shaping term. When shaping magnitude is a deliberate control variable, practitioners may monitor task and shaping components separately and, where appropriate, consider separate normalisation. This is an implementation choice rather than a condition for policy invariance.

\paragraph{Potential offset.} A constant added to $\Phi$ is often described as immaterial. The statement requires qualification, and separating five levels clarifies why. \emph{Policy invariance}: for a continuing discounted task under the assumptions of Theorem~\ref{thm:pbrs}, a uniform shift leaves the ordering of policies unchanged, and the claim is correct. \emph{Value offsets}: the shift does move $Q^{*}_{M'}$ by a constant, which is not observable in the greedy policy but is observable in any quantity compared against a fixed threshold. \emph{Terminal handling}: in episodic tasks a uniform shift is not available, because the terminal potential must remain zero; shifting non-terminal potentials alone is a genuine change to the shaping function rather than a re-parameterisation, and alters the effective incentive to terminate. \emph{Learning dynamics}: the shift changes the scale of early shaping rewards relative to the task reward, and therefore the gradient magnitudes seen before any task reward is observed. \emph{Initialisation}: by the equivalence of \citet{wiewiora2003potential}, the offset is an offset on the implied initial value function, and interacts with optimistic or pessimistic initialisation. \citet{muller2025improving} report that the offset measurably affects performance in deep RL and identify limitations of continuous potential functions that do not appear in tabular analysis. The offset and scale of a practical potential should therefore be treated as tunable implementation choices, and the episodic case should not be described as a uniform shift at all.

\subsection{Safety and robustness considerations}
\label{sec:safety}

Adaptive shaping can introduce additional failure modes because both the learner and the reward-related signal may change during training. Reward hacking and specification gaming arise when a policy exploits a proxy objective without satisfying the designer's intent \citep{amodei2016concrete,skalse2022defining}. In dynamic settings, the proxy also changes during training. A learned reward model may become inaccurate as the policy moves beyond its training distribution. An intrinsic bonus may favour stochastic noise, a world model may reward fictitious progress in latent space, and generated reward code may contain exploitable discontinuities. Reward-model over-optimisation in RLHF provides a particularly clear example of this general failure mode \citep{gao2023scaling,eisenstein2024helping,miao2024inform}.

Three notions of safety should be reported separately. \emph{Objective safety} concerns whether shaping preserves the intended optimum or equilibrium. \emph{Optimisation stability} concerns whether learning remains numerically and statistically stable under a moving target. \emph{Distributional robustness} concerns whether the information source used for shaping remains calibrated on the states and actions induced by the evolving policy. Potential-based structure addresses objective safety but does not establish optimisation stability or distributional robustness. Ensembling, uncertainty penalties, held-out evaluators, and periodic human audits may reduce exploitation of a learned reward model. None of these measures, however, replaces evaluation of the final policy under the original task criterion.

A safety-oriented evaluation protocol should include adversarial or deliberately misspecified shaping signals, held-out states and trajectories, tests for profitable cycles and terminal-state leakage, and comparisons between proxy return and an external reference criterion throughout training. For C1 this criterion is the task reward; for replacement settings it may instead be held-out human judgement or another independently specified evaluator. Ablations in which shaping updates are frozen should also be reported. For executable generated reward programs, automated structural verification and counterexample search may be appropriate before deployment. Dynamic shaping should be evaluated not only according to sample efficiency, but also according to its behaviour when the underlying information source is inaccurate.

\section{Applications}
\label{sec:applications}

The relevance of dynamic reward shaping depends less on the application label itself than on five properties of the learning problem: why a fixed signal becomes inadequate, what information can drive revision, which reference criterion remains available, which guarantee is meaningful, and which failure mode is most consequential. Table~\ref{tab:applications} compares the principal application settings along these dimensions. It therefore treats applications as distinct shaping problems rather than as examples of where a method has been deployed.

\begin{table}[!htbp]
\centering
\scriptsize
\caption{Application-level analysis of dynamic reward shaping and neighbouring adaptive mechanisms. ``Reference criterion'' is the quantity against which adaptation should ultimately be judged. The guarantee column identifies the most relevant notion of correctness, not a property already established for every method used in that domain.}
\label{tab:applications}
\begin{tabular}{@{}p{0.11\linewidth}p{0.15\linewidth}p{0.13\linewidth}p{0.13\linewidth}p{0.14\linewidth}p{0.16\linewidth}@{}}
\toprule
\textbf{Application setting} & \textbf{Why revision is useful} & \textbf{Typical information source} & \textbf{Reference criterion} & \textbf{Relevant guarantee} & \textbf{Dominant failure mode} \\
\midrule
Robotics & Intermediate behaviours are useful early but may become restrictive as competence increases & Designer objectives, success rates, demonstrations, learned models & Task success, safety constraints, and environment return & Policy preservation plus optimisation stability & Unsafe proxy exploitation, discontinuous generated rewards, or persistent over-shaping \\
Cooperative multi-agent systems & Team composition, policies, and contribution estimates co-adapt during training & Other agents, team return, counterfactual estimates, learned roles & Original stochastic game or team objective & Equilibrium consistency and valid individual credit assignment & Changing the game while appearing to accelerate learning \\
Formal-specification domains & The active subgoal changes as an automaton, plan, or temporal specification progresses & Symbolic plans, reward machines, temporal logic, measured success & Satisfaction probability or declared task objective & Policy/task-satisfaction preservation on the augmented process & Misspecified formal structure or confusion between state progression and rule revision \\
Partially observable planning & Beliefs and uncertainty change as evidence is gathered & Belief state, posterior uncertainty, predictive model & Expected return under the underlying POMDP & Preservation on an appropriate belief-augmented or history-based process & Overconfident priors and non-Markov information summaries \\
Language-model post-training & Sequence-level feedback is too delayed and the policy distribution shifts during optimisation & Preferences, critiques, reward models, language or vision-language models & Held-out human or external evaluation & Agreement with an independently specified evaluator rather than PBRS invariance & Reward-model over-optimisation and dense proxies detached from sequence quality \\
Games and sparse-control benchmarks & Novelty and value estimates evolve rapidly during hard exploration & Counts, prediction error, value estimates, learned intrinsic rewards & Original extrinsic return & Optimal-policy preservation where corrected intrinsic rewards are used & Noisy-TV behaviour, moving critic targets, and benchmark-specific overfitting \\
\bottomrule
\end{tabular}
\end{table}
\FloatBarrier

\paragraph{Robotics: competence-dependent guidance under safety constraints.} Robotics is a relevant use case when designers can specify useful intermediate behaviours without specifying an optimal controller. Scheduled auxiliary control in manipulation and teacher-driven reward weighting in navigation and physical off-road driving allow guidance to dominate early and recede or change as competence develops \citep{riedmiller2018learning,wang2025reward}. The analytical issue is not merely whether shaping improves sample efficiency, but whether the revised signal remains compatible with task success and physical safety. The same issue arises in industrial inspection and positioning tasks with prioritised multi-step targets, where staged intermediate objectives provide one way to express the required order \citep{bahrpeyma2023ur10}. A robotics evaluation should therefore compare dynamic shaping with the best static alternative, measure the original task return, and test deliberately misspecified guidance. Policy invariance is desirable when an additive potential construction is available, but it does not replace stability and safety analysis under function approximation.

\paragraph{Cooperative multi-agent systems: adaptation and objective identity.} Cooperative tasks combine sparse team rewards, long horizons, ambiguous credit assignment, and co-adaptation. Here the central question is whether an adaptive individual signal reveals contribution to the original team objective or silently defines a different game. Equilibrium consistency, rather than single-agent policy invariance, is the relevant structural criterion \citep{devlin2011theoretical,devlin2014potential}. Counterfactual advantages and value decomposition may improve attribution without modifying the reward consumed by the learner, whereas learned difference rewards or agent-specific shaping may modify it. These mechanisms can belong to the same application family while occupying different C1--C4 classes; they must therefore be evaluated separately for acceleration, credit quality, and changes to the equilibrium set. The benchmark difficulties motivating this analysis are documented across standard cooperative tasks \citep{papoudakis2021benchmarking}, and recur in deployed cooperative settings such as smart-factory scheduling and control, where the team objective is fixed by the production process while individual agents adapt around it \citep{bahrpeyma2022review}.

\paragraph{Formal specifications: interpretable progress with a classification boundary.} Temporal logic, plans, and reward machines provide an explicit representation of task progress \citep{icarte2022reward,camacho2019ltl,jiang2021temporal}. Their principal advantage is that intermediate guidance can be inspected against the declared specification. Their principal analytical ambiguity is whether the method revises a shaping rule or merely evaluates a fixed rule on a changing automaton or plan state. The latter is static shaping on an augmented representation under the definition in Section~\ref{sec:unified-framework}, even though the active subgoal changes during an episode. Truly adaptive reward design arises when the reward parameters or progress mapping are revised from performance or experience, as in adaptive LTL-progress rewards. Evaluation should therefore report both task-satisfaction performance and whether the specification itself, its reward encoding, or only its current state changed.

\paragraph{Partial observability: the state over which the guarantee is stated.} In finite-horizon online POMDP planning, belief-dependent potentials represent information about rewards beyond the planning horizon \citep{eck2016potential}. The practical benefit is an extension of the effective planning horizon, but the theoretical question is whether the belief or history is included in the process over which invariance is claimed. A fixed potential over beliefs is state-dependent static shaping; a potential re-estimated as the learner updates its model is genuinely revised and may require a larger learner-state or history-based formulation. The dominant risks are overconfident priors, insufficient belief summaries, and retrospective reward computation using information unavailable when the transition occurred.

\paragraph{Language-model post-training: evaluator agreement rather than shaping invariance.} Token-level reward redistribution can shorten extremely long credit-assignment horizons in language-model post-training \citep{chan2024dense}, while iterative reward modelling and preference optimisation adapt to the policy distribution. Most such methods are C2--C4 rather than additive shaping. Their relevant reference criterion is therefore held-out human judgement, a separately specified evaluator, or another external measure, not the PBRS preservation theorem. A central risk is that a dense or learned proxy can be optimised without improving the intended sequence-level outcome, as illustrated by reward-model over-optimisation \citep{gao2023scaling,eisenstein2024helping,miao2024inform}. Evaluation should track both proxy reward and external quality throughout training, include frozen-reward-model baselines, and test performance under distribution shift.

\paragraph{Games and sparse-control benchmarks: separating exploration benefit from benchmark fit.} Atari, sparse continuous-control suites, and hard-exploration games are common evaluation settings for the representative value-derived, uncertainty-aware, and intrinsic-reward methods reviewed here \citep{adamczyk2025bootstrapped,ma2025highly,hu2020learning}. Montezuma's Revenge is particularly useful for testing whether optimality-preserving corrections retain the exploration benefit of intrinsic motivation \citep{forbes2025action}. These domains offer controlled comparisons and an explicit extrinsic reference return, but they can overstate generality when shaping is tuned to a small benchmark set. Evaluation should therefore distinguish improvement due to adaptation from improvement due to the underlying bonus, test transfer across tasks, and report sensitivity to update rate, replay, and reward normalisation.

Across the six settings, the same selection principle recurs. Dynamic shaping is most defensible when the changing information source is necessary, the reference criterion remains independent of the adaptive signal, and the mechanism class determines the guarantee being claimed. Application evidence should therefore be organised around three comparisons: unshaped versus shaped learning, static versus dynamic shaping, and adaptive-signal return versus the original or external evaluation criterion. Without all three, an apparent application gain cannot be attributed specifically to dynamism or distinguished from objective change.

\section{Offline Reinforcement Learning}
\label{sec:offline}

Offline RL is treated separately because the fixed dataset changes the constraints under which a shaping signal can be estimated, evaluated, and corrected. Offline RL learns from a fixed dataset and cannot correct a misleading shaping rule by collecting targeted experience. Conservative Q-learning reduces overestimation of out-of-distribution actions by learning a conservative value function \citep{kumar2020conservative}. Implicit Q-learning avoids evaluating unseen actions during training \citep{kostrikov2022offline}, while TD3+BC regularises policy improvement toward the behaviour data \citep{fujimoto2021minimalist}. These methods are not dynamic reward-shaping algorithms, but they define the constraints under which shaping must operate in the offline setting.

Reward modification in offline RL can be implemented by fitting a reward model and imputing missing rewards over a fixed dataset \citep{romeo2024imputed}. If the reward model is subsequently revised through improved representations, uncertainty estimates, or limited online feedback, the same stored transitions can be associated with a sequence $R+F_k$. The latter case is included here as a dynamic extension of offline reward imputation rather than as an established algorithmic family. It creates an identification problem: without additional interaction, an increase in shaped return may reflect reward-model extrapolation rather than improved task performance. Conservative value estimation does not, by itself, establish that a learned or misspecified shaping function is safe.

Evaluation of offline shaping should therefore separate reward changes from policy regularisation, report performance under the original task criterion, test sensitivity to dataset coverage, and assess reward-model uncertainty on held-out trajectories. Where possible, reward relabelling should be constrained to potential-based form or verified against known terminal outcomes. Offline-to-online adaptation may help reveal reward-model errors that cannot be tested using a fixed dataset alone; this proposed benefit requires direct empirical evaluation.

\section{Evaluation Practice}
\label{sec:evaluation}

The preceding sections show that dynamic shaping can address several distinct limitations of reward information. They also expose different opportunities for a method to appear effective without improving its reference task criterion. Evaluation should therefore separate the benefit of shaping, the benefit of adaptation, and the validity of the final policy. The criteria set out below are proposed as a reporting standard for future work rather than applied here as an audit of the reviewed literature; a compliance audit would require re-examining the experimental sections of every included method, which is beyond the scope of this review and is identified in Section~\ref{sec:agenda} as a separate undertaking. Table~\ref{tab:protocol} summarises the criteria and states, for each, the inference that is blocked when it is omitted.

\begin{table}[!htbp]
\centering
\small
\caption{Proposed reporting criteria for dynamic-shaping studies. The final column states the inference that cannot be drawn when the criterion is omitted. The table is a reporting standard, not an assessment of compliance in the reviewed literature.}
\label{tab:protocol}
\begin{tabular}{@{}p{0.26\linewidth}p{0.28\linewidth}p{0.40\linewidth}@{}}
\toprule
\textbf{Criterion} & \textbf{What it isolates} & \textbf{Inference blocked if omitted} \\
\midrule
Unshaped baseline & Whether shaping contributes at all & That the reported gain is attributable to the shaping signal rather than to other differences in the training setup \\
\addlinespace
Best available static potential & Whether adaptation is required & That a time-varying signal is necessary, as opposed to a well-chosen fixed one \\
\addlinespace
Frozen-adaptation ablation & The contribution of dynamism itself & That the benefit arises from adaptation rather than from the content of the shaping signal \\
\addlinespace
Comparable tuning budgets & Sensitivity to hyperparameter search & That the comparison reflects the method rather than unequal search effort \\
\addlinespace
Deliberately misspecified signal & Robustness to inaccurate prior knowledge & Any claim that the method tolerates imperfect or misleading guidance \\
\addlinespace
Disclosure of implementation hazards (Section~\ref{sec:implementation}) & Whether the stated guarantee survives the implementation & That a proved invariance property applies to the code that produced the reported results \\
\addlinespace
Reference-criterion evaluation throughout training & Proxy against objective & That an improvement in the adaptive signal corresponds to an improvement under an independently stated task criterion \\
\addlinespace
Seed count and reported dispersion & Effect against run-to-run variation & That the reported difference exceeds random variation \\
\addlinespace
Natural against constructed sparsity & Scope of the result & That the result transfers to tasks that are sparse by design rather than by construction \\
\bottomrule
\end{tabular}
\end{table}
\FloatBarrier

\paragraph{Baselines.} The reporting standard proposed here recommends that studies making a dynamic-shaping claim report at least three comparisons: an unshaped baseline, the best available \emph{static} potential, and the proposed method with adaptation frozen. The third comparison isolates the effect of adaptation from the content of the shaping signal. Without it, showing that a learned potential outperforms an unshaped baseline establishes the benefit of shaping rather than the benefit of dynamic adaptation.

\paragraph{Tuning budgets.} Dynamic methods introduce adaptation rates, update frequencies, schedules, and outer-loop learning rates. A fair comparison should provide comparable tuning budgets for the proposed method and its baselines. This requirement is particularly important in deep RL, where implementation choices and random variation can materially affect comparative conclusions \citep{henderson2018deep}.

\paragraph{Robustness to misspecification.} Claims of tolerance to inaccurate prior knowledge should be tested directly by supplying a deliberately misleading signal and measuring whether performance degrades gracefully. This design is demonstrated by \citet{behboudian2022policy} and \citet{hu2020learning} and provides a suitable baseline protocol for other adaptive methods.

\paragraph{Constructed and natural sparsity.} When a sparse-reward task is created by removing dense feedback from an existing benchmark, the environment can retain smooth structure that a learned potential exploits. Results on constructed sparsity should therefore be distinguished from results on tasks that are sparse by design. Both settings are useful, but they support different claims about generalisation.

\paragraph{Guarantee claims are not audited.} The history of DPBA \citep{harutyunyan2015expressing,behboudian2020useful} demonstrates that an incorrect invariance claim can remain uncorrected while subsequent work adopts the method. Empirical improvement does not validate a theoretical guarantee. Conversely, the implementation hazards discussed in Section~\ref{sec:implementation} imply that a correct proof does not establish that a particular implementation preserves the theorem's assumptions.

\paragraph{A minimum protocol.} Reports of dynamic-shaping methods should include the three baselines defined above, per-method tuning budgets, and at least one experiment with a deliberately misspecified signal. The treatment of each implementation hazard in Section~\ref{sec:implementation} should be documented in the paper rather than only in source code. For additive shaping, performance should be evaluated throughout training under the task reward, with shaped return reported separately when needed. For replacement or adjacent mechanisms, the adaptive signal should be compared with an independently stated reference criterion. Finally, the number of random seeds and the uncertainty reported across them should be sufficient to distinguish the claimed improvement from run-to-run variation \citep{henderson2018deep}.

\section{Open Challenges and Research Agenda}
\label{sec:agenda}

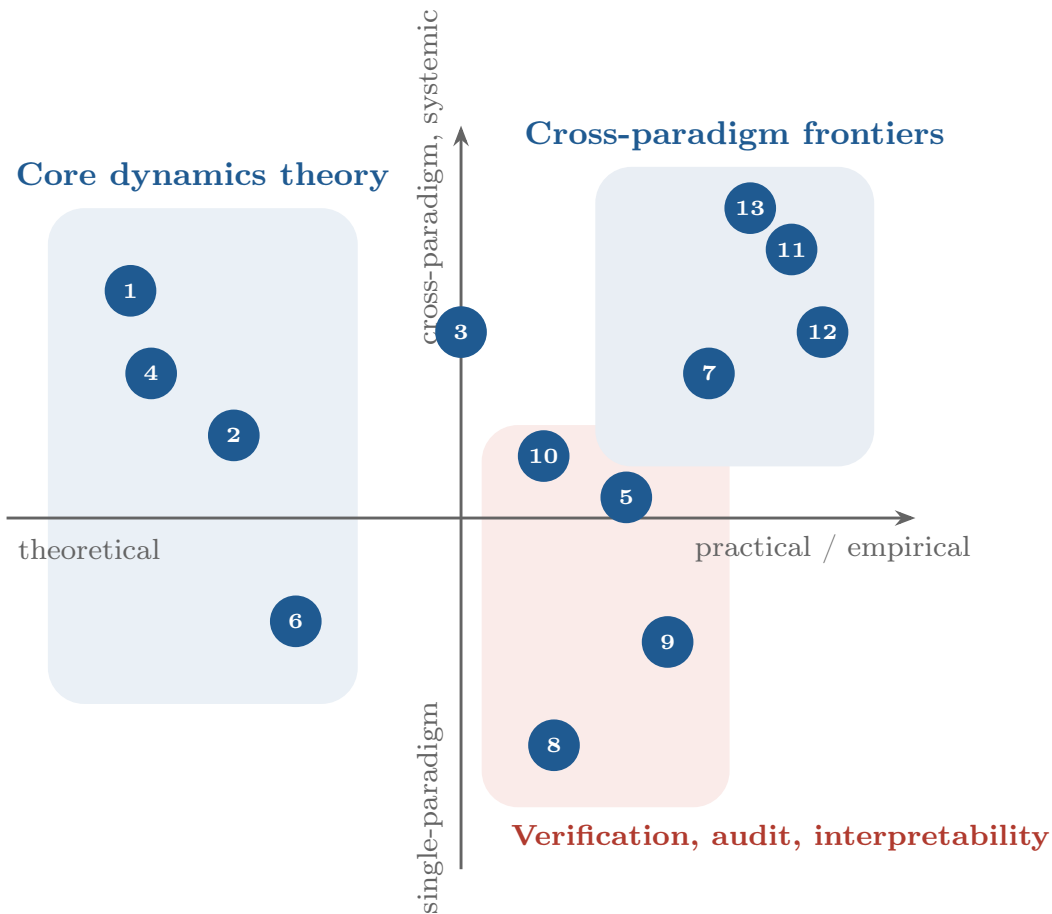
\begin{figure}[!htbp]
\centering
\resizebox{0.85\textwidth}{!}{%
\begin{tikzpicture}
  \tikzset{
    numdot/.style = {circle, fill=drsblue, text=white, font=\tiny\bfseries,
                      minimum size=5mm, inner sep=0pt}
  }
  \draw[rounded corners=10pt, fill=drsfill, draw=none] (-4.0,-1.8) rectangle (-1.0,3.0);
  \draw[rounded corners=10pt, fill=drswfill, draw=none] (0.2,-2.8) rectangle (2.6,0.9);
  \draw[rounded corners=10pt, fill=drsblue!10, draw=none] (1.3,0.5) rectangle (4.0,3.4);

  \node[font=\footnotesize\bfseries, drsblue] at (-2.5,3.3) {Core dynamics theory};
  \node[font=\scriptsize\bfseries, drswarn, anchor=west] at (0.35,-3.1) {Verification, audit, interpretability};
  \node[font=\footnotesize\bfseries, drsblue] at (2.65,3.7) {Cross-paradigm frontiers};

  \draw[drsgrey, thick, -{Stealth[length=2mm]}] (-4.4,0) -- (4.4,0);
  \draw[drsgrey, thick, -{Stealth[length=2mm]}] (0,-3.4) -- (0,3.8);
  \node[font=\scriptsize, drsgrey] at (-3.6,-0.3) {theoretical};
  \node[font=\scriptsize, drsgrey] at (3.6,-0.3) {practical / empirical};
  \node[font=\scriptsize, drsgrey, rotate=90] at (-0.3,3.3) {cross-paradigm, systemic};
  \node[font=\scriptsize, drsgrey, rotate=90] at (-0.3,-2.8) {single-paradigm};

  \node[numdot] (n1) at (-3.2,2.2) {1};
  \node[numdot] (n2) at (-2.2,0.8) {2};
  \node[numdot] (n4) at (-3.0,1.4) {4};
  \node[numdot] (n6) at (-1.6,-1.0) {6};
  \node[numdot] (n3) at (0.0,1.8) {3};
  \node[numdot] (n5) at (1.6,0.2) {5};
  \node[numdot] (n8) at (0.9,-2.2) {8};
  \node[numdot] (n9) at (2.0,-1.2) {9};
  \node[numdot] (n10) at (0.8,0.6) {10};
  \node[numdot] (n7) at (2.4,1.4) {7};
  \node[numdot] (n11) at (3.2,2.6) {11};
  \node[numdot] (n12) at (3.5,1.8) {12};
  \node[numdot] (n13) at (2.8,3.0) {13};
\end{tikzpicture}}

\vspace{6pt}
\begin{minipage}{0.94\textwidth}
\centering
\scriptsize
\textbf{1} rate of change \quad \textbf{2} self-referential stability \quad \textbf{3} learning potentials, not rewards \quad \textbf{4} sequences of shaped problems \quad \textbf{5} verification of generated rewards \quad \textbf{6} shaping signals as advantages \quad \textbf{7} partial observability \quad \textbf{8} dedicated benchmark \quad \textbf{9} compliance audit \quad \textbf{10} interpretability of learned potentials \quad \textbf{11} reward learning and RLHF \quad \textbf{12} foundation and world models \quad \textbf{13} offline and multi-agent RL
\end{minipage}
\caption{The thirteen open directions below, placed by how theoretical versus practical the question is (horizontal) and how confined to the standard single-agent setting versus how systemic and cross-paradigm it is (vertical). Placement is qualitative and intended to reveal grouping, not to support numerical comparison between items. Item 3 sits outside every shaded cluster because it is the connective pattern between them rather than a member of any one group.}
\label{fig:agenda-map}
\end{figure}

The reviewed candidate set suggests a gap between the flexibility of adaptive shaping mechanisms and the theory available to analyse their learning behaviour. The following research directions are ordered around the central need to connect dynamic information sources, learning behaviour, and verifiable task preservation. Read individually, the thirteen items below are a list; Figure~\ref{fig:agenda-map} places them on two axes, how theoretical versus practical the open question is, and how confined to a single paradigm versus how systemic and cross-cutting it is, so that the clustering itself becomes part of the argument. Three groups emerge. A cluster of foundational theory questions (items 1, 2, 4, 6) concerns the standard single-agent setting and is largely self-contained; a cluster of verification and tooling questions (items 5, 8, 9, 10) is practical but still bounded within existing method families; and a cluster of cross-paradigm frontiers (items 7, 11, 12, 13) sits where dynamic shaping theory meets partial observability, preference learning, foundation models, and multi-agent or offline RL, none of which the classical theory of Section~\ref{sec:foundations} was built to cover. Item 3, learning potentials rather than rewards, sits outside all three clusters by design: it is the bridging pattern, identified independently in Section~\ref{sec:taxonomy-observations} and Section~\ref{sec:trajectory}, that lets a result proved in the foundational cluster be carried into the frontier cluster without a new invariance proof.

\begin{enumerate}[leftmargin=*]
  \item \textbf{A theory of the rate of change.} Time-varying potentials can preserve optimal policies when the dynamic potential terms are paired according to the cited construction \citep{devlin2012dynamic}, and shaping can improve sample complexity in restricted settings \citep{laud2003influence,gupta2022unpacking}. The formal results reviewed here do not provide a general rule for selecting a shaping signal's update rate. Annealing rates, update frequencies, and freezing points are selected through empirical search. A result relating the rate of change of a potential to learning stability and speed, even in tabular settings, would convert the central design variable of dynamic shaping from a tuning choice into an analysable quantity.
  \item \textbf{Stability of self-referential shaping under function approximation.} Source I3 is the most frequent source among the methods included here, as shown in Section~\ref{sec:taxonomy-observations}. Its characteristic feedback loop has been analysed for individual methods and primarily in tabular settings; the works reviewed here do not provide general stability conditions covering function approximation, bootstrapping, and replay.
  \item \textbf{Learning potentials rather than rewards.} The approach exemplified by \citet{muller2026automating} restricts the object that is learned rather than the information used to learn it. This design pattern offers a direct route to combining expressive information sources with exact guarantees and warrants broader investigation.
  \item \textbf{Guarantees for sequences of shaped problems.} Curricula, staged objectives, and regenerated reward programs define sequences of shaped MDPs rather than a single shaped problem. The properties preserved across such sequences remain largely uncharacterised, despite their prevalence in current practice.
  \item \textbf{Verification of generated reward functions.} Foundation models can generate reward code faster than it can be audited manually \citep{ma2024eureka,xie2024text2reward}. Automated checks for potential-based structure, terminal-state correctness, and profitable cycles are promising verification problems for generated reward functions. Such tools could connect the formal discipline of reward shaping with the productivity of generative methods.
  \item \textbf{Shaping signals as advantages.} \citet{macglashan2017interactive} showed that feedback defined relative to current competence behaves as an advantage rather than a reward. Many contemporary shaping signals have the same dependence on the current learner. Whether a reward-advantage mismatch explains instabilities in value-derived and intrinsic-motivation methods is an open and testable question.
  \item \textbf{Partial observability.} Beyond online planning \citep{eck2016potential} and Bayesian formulations \citep{marom2018belief}, theory remains limited for agents that observe only partial information about the state. Standard invariance arguments do not transfer directly to this setting.
  \item \textbf{A dedicated benchmark.} No benchmark suite with all four properties was identified in the literature reviewed for this study: naturally sparse rewards, controllable prior quality, non-stationary targets, and long compositional horizons. Developing such a benchmark would support more consistent application of the evaluation protocol in Section~\ref{sec:evaluation}.
  \item \textbf{A compliance audit of the existing literature.} The criteria in Table~\ref{tab:protocol} are proposed here as a reporting standard; they have not been applied retrospectively to the methods reviewed. A broader structured audit, recording for each published dynamic-shaping method which criteria its experiments satisfy, would establish how much of the reported evidence supports the benefit of \emph{adaptation} as distinct from the benefit of shaping. Such an audit requires re-examination of the experimental sections and, where available, the released code of each method, and is therefore a separate undertaking from the present review.
  \item \textbf{Interpretability of learned potentials.} A learned potential represents which states the system considers worth approaching. This object is particularly important to inspect in safety-relevant applications, yet it is generally implemented as an opaque function approximator. The reward-misspecification literature \citep{amodei2016concrete,pan2022effects,skalse2022defining} illustrates the consequences of an incorrect representation.

  \item \textbf{Reward learning and RLHF.} RLHF, preference optimisation, constitutional AI, and iterative reward refinement blur the distinction between reward learning and reward shaping \citep{ouyang2022training,bai2022constitutional,rafailov2023direct}. Definition~\ref{def:drs-general} provides a common information-conditioned language. Further theory is needed for coupled policy and reward-model updates, distribution shift, and the boundary between additive shaping and direct objective optimisation.
  \item \textbf{Foundation models and world models.} Language and vision-language models generate reward programs or semantic scores, while latent world models provide uncertainty, novelty, and imagined-progress signals \citep{ma2024eureka,rocamonde2024vision,hafner2025mastering}. The central challenge is to constrain these expressive sources to verifiable shaping classes without discarding their semantic or predictive advantages.
  \item \textbf{Offline and multi-agent reinforcement learning.} Offline RL and cooperative MARL present complementary identification problems. Fixed datasets make reward-model errors difficult to falsify, while shared returns obscure individual contributions \citep{kumar2020conservative,foerster2018counterfactual,wang2021qplex}. Theory should distinguish acceleration from objective change and provide guarantees for conservative relabelling, learned difference rewards, changing team composition, and potential-game formulations.

\end{enumerate}

\section{Conclusion}
\label{sec:conclusion}

The classical account of reward shaping uses a potential function to preserve the optimal policy. This account describes a static regime and does not encompass the dynamic methods considered in this review. Contemporary shaping signals include decaying novelty bonuses, potentials derived from evolving value functions, Bayesian priors that diminish with evidence, annealed or meta-optimised weights, automaton-dependent subgoal rewards, feedback defined relative to current competence, and reward programs revised by foundation models. Their dependence on training progress and acquired information is a defining design feature rather than an incidental implementation detail.

This review has organised the literature within a common framework. Policy invariance extends to time-varying potentials under the assumptions of the cited construction, optimality-preserving constructions now include classes of intrinsic motivation beyond potential-based form, and the Bayes-adaptive formulation represents shaping and exploration bonuses within a shared object. Classifying methods by mechanism class, temporal signature, information source, and guarantee class, while separating parametric revision from mere state dependence, reveals recurring designs across otherwise disconnected subfields. The same analysis identifies specific requirements for reliable use. Interactions among time-varying rewards, replay buffers, bootstrapped critics, multi-step returns, and reward normalisation should be handled explicitly. The evaluation protocol proposed here recommends a frozen-adaptation baseline to isolate the contribution of dynamism, and that theoretical guarantees be audited against both their formal assumptions and their implementation.

The central open question is consequently narrower than the breadth of the literature suggests. A central design variable of dynamic shaping is the rate at which the signal changes, yet no general result relates that rate to the stability or sample complexity of the learner. Establishing this relationship would connect the safety theory of reward shaping to its intended learning benefit. Together with implementation-level audits of invariance claims and benchmarks based on naturally sparse rewards, such a theory would provide the foundation required for reliable progress in dynamic reward shaping.

\appendix
\section{Search Protocol and Supplementary Candidate Table}
\label{app:candidates}

\subsection{What the search protocol did and did not do}
\label{app:protocol}

Section~\ref{sec:review-approach} describes the review as theory-led and citation-traced rather than systematic. This appendix states explicitly which elements of a systematic-review protocol were followed and which were not, so that the completeness of Table~\ref{tab:classification} can be assessed on its actual basis rather than assumed.

\emph{What was done.} A seed set of foundational results was identified from established familiarity with the reward-shaping literature \citep{ng1999policy,wiewiora2003potential,devlin2011theoretical,devlin2012dynamic}. Forward citations (papers citing the seed set) and backward citations (papers cited by the seed set and by each subsequently added paper) were traced through publisher pages, Semantic Scholar, and Google Scholar, in rounds repeated as new method families were identified, with the last round conducted in August 2026. A candidate paper was added to Table~\ref{tab:classification} when it met at least one of the inclusion criteria stated in Section~\ref{sec:review-approach}: it establishes a formal property, introduces a distinct adaptation mechanism, corrects an earlier claim, provides a widely used baseline, or sharpens a boundary in the mechanism classification of Section~\ref{sec:unified-framework}. A candidate was left out of the table when it adapts a learning signal without touching the reward and does not sharpen a class boundary, or when a later, more general result from the same research group superseded it and both could not be discussed without redundancy.

\emph{What was not done.} No database (e.g., Scopus, Web of Science, ACM Digital Library) was queried with a fixed search string, no search-string log was kept, no fixed cut-off date range per database was applied, no independent second screener verified inclusion or exclusion decisions, no count of papers screened but excluded was retained, and no formal duplicate-record deduplication procedure was needed because candidates were added individually through citation tracing rather than merged from parallel database exports. Consequently this review cannot report a PRISMA-style flow diagram or an exclusion count, and Table~\ref{tab:classification} should be read as a curated, theory-motivated candidate pool rather than a census of the field. The taxonomic counts of Section~\ref{sec:taxonomy-observations} and Figure~\ref{fig:landscape} describe this pool and not a claim about the full literature.

\subsection{Candidate table with publication status}
\label{app:status-table}

Table~\ref{tab:candidates} lists every C1--C4 entry of Table~\ref{tab:classification}, grouped by the same method families, together with its venue, year, and peer-review status at the time of writing (August 2026). \emph{Peer-reviewed} denotes a venue with an independent review process: a conference proceedings, journal, or workshop with reviewed submissions. \emph{Preprint} denotes a manuscript that has not, to our knowledge, completed peer review at any venue, including arXiv-only technical reports from industrial laboratories. Where a paper exists in both forms, the peer-reviewed venue is the primary citation and the preprint identifier is retained as a note. Two entries in this table, PIES and VLM semantic rewards, combine a peer-reviewed citation with a preprint or workshop-only citation for the same method; each is marked \emph{Mixed} and annotated rather than assigned a single status.

{\small
\begin{longtable}{@{}p{0.27\linewidth}p{0.24\linewidth}p{0.15\linewidth}p{0.28\linewidth}@{}}
\caption{Supplementary candidate table. For each C1--C4 entry of Table~\ref{tab:classification}, the venue and year of the primary citation, its peer-review status as of August 2026, and a note on preprint identifiers or mixed status. Ordered by the method families of Section~\ref{sec:families}.}
\label{tab:candidates}\\
\toprule
\textbf{Method (Table~\ref{tab:classification})} & \textbf{Venue and year} & \textbf{Status} & \textbf{Note} \\
\midrule
\endfirsthead
\multicolumn{4}{c}{\tablename\ \thetable\ (continued)}\\
\toprule
\textbf{Method (Table~\ref{tab:classification})} & \textbf{Venue and year} & \textbf{Status} & \textbf{Note} \\
\midrule
\endhead
\midrule
\multicolumn{4}{r}{Continued on next page}\\
\endfoot
\bottomrule
\endlastfoot
\multicolumn{4}{@{}l}{\textit{Reference point}}\\
Static PBRS \citep{ng1999policy} & ICML, 1999 & Peer-reviewed & \\
\midrule
\multicolumn{4}{@{}l}{\textit{\S\ref{sec:fam-schedule} Schedule-based}}\\
Scheduled auxiliary control \citep{riedmiller2018learning} & ICML, 2018 & Peer-reviewed & \\
Heuristic-guided RL \citep{cheng2021heuristic} & NeurIPS, 2021 & Peer-reviewed & \\
Reward Training Wheels \citep{wang2025reward} & IROS, 2025 & Peer-reviewed & Preprint: arXiv:2503.15724 \\
\midrule
\multicolumn{4}{@{}l}{\textit{\S\ref{sec:fam-value} Value-derived potentials}}\\
Online shaping-reward learning \citep{grzes2010online} & Neural Networks (journal), 2010 & Peer-reviewed & \\
Bootstrapped shaping \citep{adamczyk2025bootstrapped} & AAAI, 2025 & Peer-reviewed & \\
Exploration-guided shaping \citep{devidze2022exploration} & NeurIPS, 2022 & Peer-reviewed & \\
\midrule
\multicolumn{4}{@{}l}{\textit{\S\ref{sec:fam-bayes} Uncertainty-aware}}\\
Belief reward shaping \citep{marom2018belief} & AAAI, 2018 & Peer-reviewed & \\
Self-adaptive shaping \citep{ma2025highly} & ICLR, 2025 & Peer-reviewed & \\
POMDP potential shaping \citep{eck2016potential} & Autonomous Agents and Multi-Agent Systems (journal), 2016 & Peer-reviewed & \\
\midrule
\multicolumn{4}{@{}l}{\textit{\S\ref{sec:fam-advice} Advice- and demonstration-derived}}\\
Plan-based shaping \citep{grzes2008plan} & IEEE Intelligent Systems Conf., 2008 & Peer-reviewed & \\
Dynamic potential-based advice \citep{harutyunyan2015expressing} & AAAI, 2015 & Peer-reviewed & Invariance claim subsequently refuted; see \S\ref{sec:fam-advice} \\
PIES \citep{behboudian2020useful,behboudian2022policy} & ALA Workshop at AAMAS, 2020; Neural Computing and Applications (journal), 2022 & Mixed & 2020 is a workshop paper; 2022 journal article is the primary, reviewed citation \\
Subgoal-based shaping \citep{okudo2021subgoal} & IEEE Access (journal), 2021 & Peer-reviewed & Preprint: arXiv:2104.06411 \\
Self-supervised online shaping \citep{memarian2021self} & IROS, 2021 & Peer-reviewed & \\
\midrule
\multicolumn{4}{@{}l}{\textit{\S\ref{sec:fam-human} Interactive human shaping}}\\
TAMER \citep{knox2009interactively} & K-CAP, 2009 & Peer-reviewed & \\
Policy shaping \citep{griffith2013policy} & NeurIPS, 2013 & Peer-reviewed & \\
COACH \citep{macglashan2017interactive} & ICML, 2017 & Peer-reviewed & \\
Deep TAMER \citep{warnell2018deep} & AAAI, 2018 & Peer-reviewed & \\
\midrule
\multicolumn{4}{@{}l}{\textit{\S\ref{sec:fam-im} Intrinsic motivation}}\\
Count-based / ICM / RND \citep{bellemare2016unifying,pathak2017curiosity,burda2019exploration} & NeurIPS 2016; ICML 2017; ICLR 2019 & Peer-reviewed & \\
PBIM \citep{forbes2024pbim} & AAMAS, 2024 & Peer-reviewed & G1 in Table~\ref{tab:classification} \\
GRM \citep{forbes2024generalized} & arXiv, 2024 & Preprint & arXiv:2410.12197; had not appeared at a reviewed venue at time of writing; G1$^{\ddagger}$ in Table~\ref{tab:classification}, excluded from Figure~\ref{fig:landscape} \\
ADOPS \citep{forbes2025action} & ICML, 2025 (PMLR vol.\ 267, pp.\ 17437--17451) & Peer-reviewed & Preprint: arXiv:2505.12611 \\
BAMDP shaping \citep{lidayan2025bamdp} & ICLR, 2025 & Peer-reviewed & \\
\midrule
\multicolumn{4}{@{}l}{\textit{\S\ref{sec:fam-bilevel} Bi-level and meta-optimised}}\\
Online reward design \citep{sorg2010reward} & NeurIPS, 2010 & Peer-reviewed & \\
Learned intrinsic rewards \citep{zheng2018learning} & NeurIPS, 2018 & Peer-reviewed & \\
Shaping-weight bi-level \citep{hu2020learning} & NeurIPS, 2020 & Peer-reviewed & \\
Behaviour alignment \citep{gupta2023behavior} & NeurIPS, 2023 & Peer-reviewed & \\
ROSA \citep{mguni2023learning} & AAAI, 2023 & Peer-reviewed & \\
Assistant reward agent \citep{ma2024assistant} & ICML, 2024 (PMLR vol.\ 235, pp.\ 33925--33939) & Peer-reviewed & \\
\midrule
\multicolumn{4}{@{}l}{\textit{\S\ref{sec:fam-structure} Structure-driven}}\\
Reward machines \citep{icarte2022reward} & Journal of Artificial Intelligence Research, 2022 & Peer-reviewed & \\
LTL-based shaping \citep{camacho2019ltl,jiang2021temporal} & IJCAI 2019; AAAI 2021 & Peer-reviewed & \\
Adaptive LTL-progress rewards \citep{kwon2025adaptive} & UAI, 2025 (PMLR vol.\ 286, pp.\ 2472--2485) & Peer-reviewed & Preprint: arXiv:2412.10917 \\
Curriculum-coupled shaping \citep{narvekar2020curriculum} & Journal of Machine Learning Research, 2020 & Peer-reviewed & \\
\midrule
\multicolumn{4}{@{}l}{\textit{\S\ref{sec:fam-preference} Preference-based and RLHF}}\\
Iterative RLHF \citep{ouyang2022training} & NeurIPS, 2022 & Peer-reviewed & \\
Constitutional AI \citep{bai2022constitutional} & arXiv, 2022 & Preprint & Industrial technical report; no independent peer review identified at time of writing \\
DPO \citep{rafailov2023direct} & NeurIPS, 2023 & Peer-reviewed & \\
\midrule
\multicolumn{4}{@{}l}{\textit{\S\ref{sec:fam-world} World-model and latent-prediction}}\\
Predictive uncertainty / disagreement \citep{pathak2019disagreement,sekar2020planning} & ICML 2019; ICML 2020 & Peer-reviewed & \\
Latent novelty / imagined reachability \citep{fu2023gobi} & ICML, 2023 & Peer-reviewed & \\
\midrule
\multicolumn{4}{@{}l}{\textit{\S\ref{sec:fam-foundation} Foundation-model-driven}}\\
Eureka \citep{ma2024eureka} & ICLR, 2024 & Peer-reviewed & \\
Text2Reward \citep{xie2024text2reward} & ICLR, 2024 & Peer-reviewed & \\
LLM heuristics \citep{bhambri2024extracting} & arXiv, 2024 & Preprint & arXiv:2405.15194; no reviewed venue identified at time of writing \\
VLM semantic rewards \citep{rocamonde2024vision,baumli2024vision} & ICLR 2024 (Rocamonde et al.); arXiv, 2024 (Baumli et al.) & Mixed & Baumli et al.\ is an industrial technical report (arXiv:2312.09187) \\
VLM-guided potentials \citep{muller2026automating} & arXiv, 2026 & Preprint & arXiv:2606.27180 (June 2026); presented here alongside peer-reviewed conference papers only for conceptual comparison \\
Dense reward for free \citep{chan2024dense} & ICML, 2024 & Peer-reviewed & \\
\midrule
\multicolumn{4}{@{}l}{\textit{\S\ref{sec:fam-marl} Multi-agent}}\\
Potential-based difference rewards \citep{devlin2014potential} & AAMAS, 2014 & Peer-reviewed & \\
Counterfactual advantages \citep{foerster2018counterfactual} & AAAI, 2018 & Peer-reviewed & \\
Value decomposition \citep{rashid2018qmix,wang2021qplex} & ICML 2018; ICLR 2021 & Peer-reviewed & \\
\end{longtable}
}

Of the entries above, three are preprint-only at the time of writing (GRM, LLM heuristics, and VLM-guided potentials) and one is an industrial technical report without independent peer review (Constitutional AI); Baumli et al.\ is a further industrial report. GRM and VLM-guided potentials both claim an optimality-preserving construction, potential-based or reward-matching in form, of the kind reviewed in Section~\ref{sec:beyond-pbrs}; this review reports rather than independently verifies that claim for either, marks both G1$^{\ddagger}$ in Table~\ref{tab:classification}, and excludes both from the counts in Figure~\ref{fig:landscape} for that reason. The remaining three preprint or industrial-report entries are discussed only as illustrations of an adaptation mechanism, without a G1--G3 claim attached, consistent with the qualification already placed on \citet{muller2026automating} in Sections~\ref{sec:taxonomy-observations} and \ref{sec:agenda}.

\bibliographystyle{plainnat}
\bibliography{references}

\end{document}